%% file: conference.tex
\documentclass{article} 
\usepackage{conference,times}

\usepackage{placeins}
\usepackage{graphicx}
\usepackage{amsmath}
\usepackage{algorithm}
\usepackage{algorithmic}
\usepackage{multirow}
\usepackage{hyperref}
\usepackage{url}
\usepackage{xcolor}
\usepackage{multirow}
\usepackage{amssymb}
\usepackage{enumitem}
\usepackage{subcaption}
\usepackage{bm}
\usepackage{booktabs}
\usepackage[table]{xcolor}
\usepackage{svg}

\title{PFedMA: Who Should Join Next as Federated Multi-Agent Orchestration Evolves?}

\author{Shuai Li\textsuperscript{1}  \quad Shengyuan Liu\textsuperscript{1}  \quad  Fan Qi\textsuperscript{1,2}  \quad Xiaoshan Yang\textsuperscript{3} \\
\textsuperscript{1}Tianjin University of Technology  \quad \textsuperscript{2}Inner Mongolia University \\ \textsuperscript{3}Institute of automation, Chinese academy of science
}

\iclrfinalcopy

\begin{document}

\maketitle

\fancyhead{}
\renewcommand{\headrulewidth}{0pt}

\begingroup
\renewcommand{\thefootnote}{}
\footnotetext{
Correspondence to: Fan Qi\textsuperscript{1,2}
  \texttt{<fanqi@imu.edu.cn; fanqi@email.tjut.edu.cn>}.
  Additional author contact information:
  Shuai Li\textsuperscript{1} \texttt{<lshuai@stud.tjut.edu.cn>} and Xiaoshan Yang\textsuperscript{3} \texttt{<xiaoshan.yang@nlpr.ia.ac.cn>}.
}
\endgroup

\input{context/change/abstract_change}
\input{context/change/introduction_change}
\input{context/related_work}

\input{context/change/method_change}
\input{context/change/experiments_change}

\input{context/conclusion}

\newpage

\bibliography{main}
\bibliographystyle{conference}
\newpage
\input{context/change/appendix_change}

\end{document}

%% file: context/change/abstract_change.tex
\begin{abstract}

Federated Multi-Agent Systems (FedMAS) offer a promising paradigm for collaborative reasoning among LLM-based agents while keeping raw client data decentralized. However, reliable multi-round collaboration remains challenging. 
Agent selection must take into consideration evolving task requirements and the agent’s historical reliability, while privacy rules must adapt to changing collaboration contexts and meet the client’s personalized needs.
To address these challenges, we propose \textbf{PFedMA}, a federated multi-agent framework for feedback driven multi-round collaboration under heterogeneous privacy requirements. Across rounds, PFedMA maintains historical capability and reliability states to guide agent selection, evaluates task responses before aggregation, and refines subsequent queries according to accumulated collaboration feedback. PFedMA enables bidirectional local to global and global to local private rule evolution across heterogeneous clients. Clients generalize locally updated privacy rules and selectively incorporate globally maintained rules, thereby facilitating cross-client privacy rule evolution while preserving personalized constraints. Extensive experiments across reasoning, privacy sensitive, and adversarial settings show that PFedMA improves task performance, reduces measured privacy leakage, and improves system robustness. 
\end{abstract}

%% file: context/change/introduction_change.tex
\section{Introduction}
\label{section:introduction}

LLM based multi-agent systems (MAS) coordinate agents with complementary capabilities to address complex tasks~\citep{guo2024large,wang2024mixture}. 
In privacy sensitive settings, however, the knowledge required for collaboration is distributed across multiple data holders with distinct privacy requirements, data ownership constraints, and access control policies.
To complete the collaborative task, agents must integrate distributed knowledge and intermediate reasoning information, while raw data cannot be centralized. This challenge motivates federated multi-agent systems (FedMAS), a distributed collaboration paradigm in which agents reason over locally held information and exchange task relevant information without directly centralizing raw client data~\citep{shi2025privacy}.  However, existing FedMAS does not fully address how to select reliable agents or adapt information sharing rules as reasoning unfolds across rounds.

As collaboration progresses, both \emph{who should participate} and \emph{what information should be shared} may change with the evolving reasoning state. Early rounds may require broad domain expertise, whereas later rounds may call for specialized agents to resolve uncertainties exposed by intermediate responses. 
Meanwhile, as reasoning progresses and collaboration demands evolve, newly required information may introduce sharing scenarios that are not covered by existing privacy rules. 
Consequently, agent orchestration and privacy control become closely interdependent across reasoning rounds. \emph{Agent orchestration determines whose knowledge enters the collaborative reasoning process}, while \emph{privacy rules determine what information can be shared}. The resulting responses and reasoning state, in turn, affect both subsequent agent selection and the privacy constraints relevant to later rounds.
FedMAS therefore needs to use feedback from intermediate responses to update both agent selection and privacy rules across rounds.

These two interdependent decisions give rise to two challenges. First, task response reliability must inform agent selection across rounds. Previous studies have explored iterative interactions between clients in federated collaboration~\citep{wang2025federated}, as well as multi-round routing based on the current interaction context~\citep{zhang2025router}.
However, selecting agents based only on the current query context may overlook agents' historical reliability in previous rounds.
Because agents may produce incomplete, erroneous, or adversarial responses~\citep{deng2025belief,zhang2026searching}, selecting agents without accounting for their past reliability can propagate errors across rounds.

Second, privacy rules must adapt to evolving collaboration while preserving heterogeneous client requirements. Existing agent-level approaches protect sensitive information through context filtering and rule-based protection~\citep{shi2025privacy,xiang2025guardagent}. 
Separately, applying DP to agent responses~\citep{panda2023differentially} may introduce perturbations that propagate across rounds, increasing the risk of semantic drift and inconsistent reasoning. 
These approaches typically rely on predefined privacy constraints or fixed protection strategies, which may fail to accommodate changing information needs across reasoning rounds, either suppressing information necessary for subsequent reasoning or failing to capture newly emerging privacy risks~\citep{truong2021privacy,shi2025privacy}. Moreover updating rules only locally limits cross client reuse, while applying shared rules uniformly can overlook differences in client requirements, data characteristics, and task contexts~\citep{kossek2024survey}. 
The challenge is to reuse newly identified privacy rules across clients without overriding their distinct local requirements.

To address these challenges, we propose \textbf{PFedMA}, a federated multi-agent framework for feedback driven multi-round collaboration under heterogeneous privacy requirements.
At each round, PFedMA evaluates task responses and updates historical capability and reliability estimates to guide agent selection and query refinement in the next round.
Response reliability also determines which clients contribute new privacy rule candidates.
These candidates are generalized for cross-client reuse, while each client selectively adopts rules according to its local requirements; the adopted rules then govern subsequent queries and responses.

Our main contributions are summarized as follows:

\begin{itemize}[leftmargin=*, align=left]

\item We propose PFedMA, a federated multi-agent system with feedback driven orchestration that leverages cross round capability and reliability feedback to adapt agent selection, thereby mitigating repeated selection of unreliable agents and error accumulation across rounds.

\item We introduce a federated privacy rule co-evolution mechanism in PFedMA that enables privacy rules to adapt to evolving collaboration while preserving heterogeneous client requirements, allowing local privacy rules to benefit other clients without overriding personalized privacy rules.

\item Experiments across diverse benchmarks show that PFedMA improves average task performance and reduces measured privacy leakage compared with Fed-ICL, while achieving a lower attack success rate than EPEAgents under the evaluated prompt injection attacks.
\end{itemize}

%% file: context/related_work.tex
\section{Related Work}

\paragraph{Multi-Agent System.}
Multi-agent systems (MAS) predominantly rely on centralized architectures, structuring collaboration via domain personas  \citep{tang2024medagents}, external retrieval  \citep{xiao2025moodangels}, or predefined execution pipelines  \citep{chen2023agentverse, wang2024mixture}. 
To resolve conflicts and reach consensus, systems employ diverse alignment strategies, including textual debates and voting  \citep{choi2025debate, sun2025cortexdebate}, adaptive termination  \citep{hu2025multi}, and belief state broadcasting  \citep{deng2025belief, keshmirian2025many}. 
Recent advances in task routing further optimize this orchestration using dynamic paths  \citep{yue2025masrouter} and reinforcement learning  \citep{zhang2025router}. 
However, these centralized paradigms mandate transparent raw context sharing, inherently violating data privacy. In contrast, our framework achieves dynamic multi-round dispatch while operating under explicit local data boundaries.

\paragraph{Federated Multi-Agent System.} 
To reduce communication overhead, federated paradigms increasingly leverage text driven prompting~\citep{qiu2023text}, textual gradients ~\citep{chen2025can}, and in context learning~\citep{wu2024federated}. 
Similarly, cooperative frameworks iteratively aggregate plain text responses from local exemplars~\citep{wang2025federated}. 
However, sharing raw text introduces severe prompt extraction risks~\citep{duan2024privacy}, making secure communication a critical challenge~\citep{kossek2024survey}.  

Existing privacy enhancing approaches, including trusted environments~\citep{nagar2021privacy}, node decomposition~\citep{wang2021privacy}, and differential privacy~\citep{panda2023differentially}, reduce information exposure through different mechanisms. However, depending on the mechanism, they may require additional trusted infrastructure or restrict the information available during collaboration, making it difficult to preserve semantically useful context across multiple reasoning rounds~\citep{shi2025privacy}.
To mitigate this risk, we restrict server bound communication to privacy processed responses, generalized privacy rule candidates and compact rule level feedback, thereby preserving the stated client local data boundary.

\begin{figure}
    \centering
    \includegraphics[width=\linewidth]{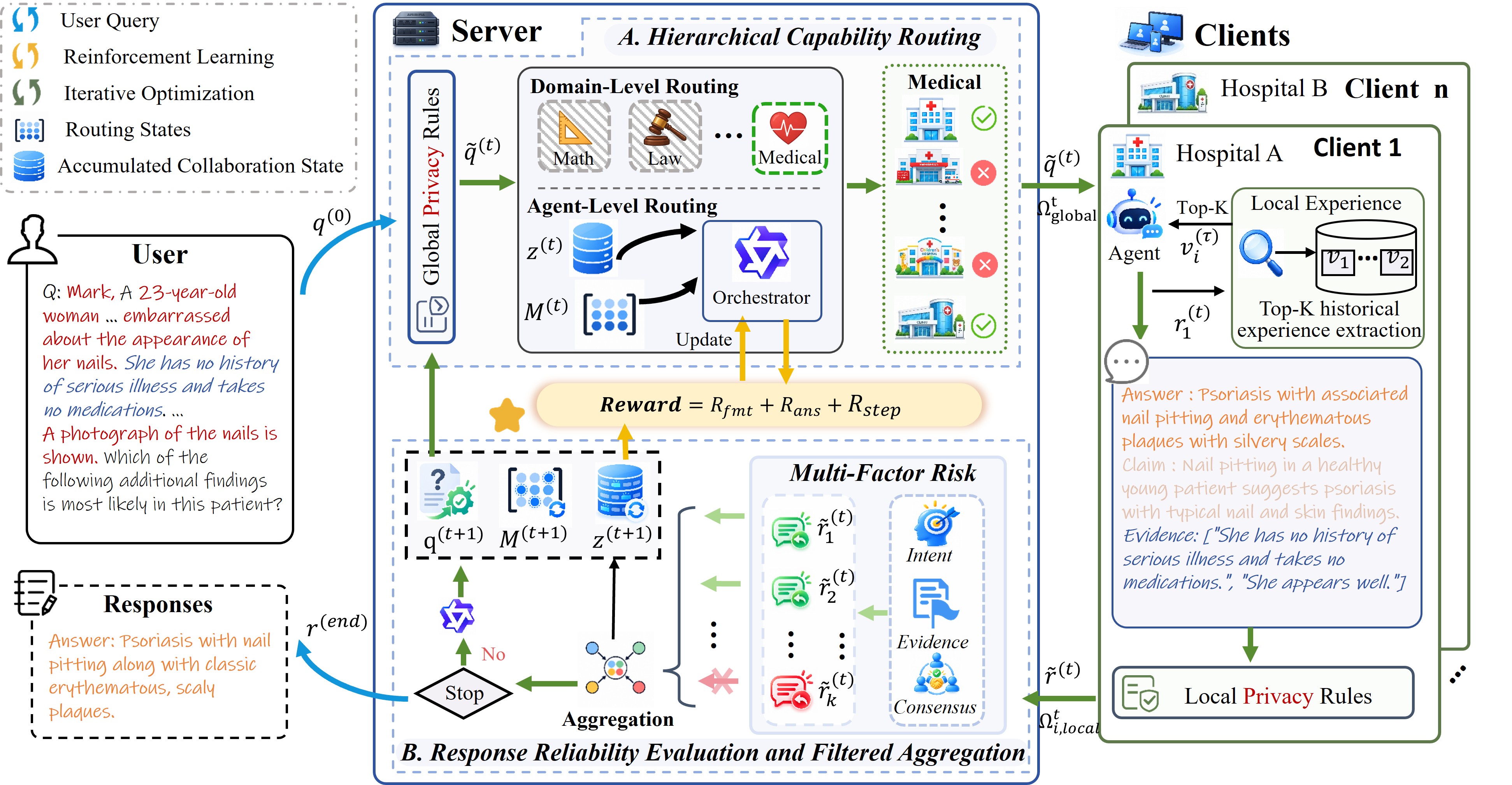}
    \caption{Overview of PFedMA. Task response feedback guides query refinement and client routing, while privacy rules evolve through cross client sharing and personalized adaptation. Together, they jointly determine the participants and the content to be shared, thereby laying the foundation for continued collaboration. The algorithm is provided in Appendix~\ref{alg:pfedma}.}

    \label{fig:kuangjiatu}
\end{figure}

%% file: context/change/method_change.tex
\section{Method}
\label{headings}

\subsection{Preliminaries}
\label{section:preliminaries}

We consider a federated multi-agent system with an honest but curious server and $N$ heterogeneous clients $\mathcal{A}=\{A_1,\ldots,A_N\}$. Each client $A_i$ maintains a private dataset $\mathcal{D}_i$ and a local experience buffer $\mathcal{X}_i$, both of which remain on device. Although the server cannot directly access clients' raw local data or unprocessed local responses, it may attempt to infer private information from uploaded privacy processed responses $\tilde{r}_i^t$. Meanwhile, some clients may provide unreliable or malicious task responses. 

To address these concerns, we introduce historical capability and reliability states, along with global and personalized privacy rule repositories for query and response privacy processing, respectively. At round $t$, the server maintains the accumulated collaboration state $z^{t-1}$ and processes the query $q^t$ using its global privacy rule repository $\Omega_{\mathrm{global}}^t$ to obtain $\tilde{q}^t$. The orchestrator $\pi_\theta$ selects a subset of participating clients based on their routing state $M_i^t$ and sends $\tilde{q}^t$ to the selected clients. Each selected client generates a local response $r_i^t=A_i(\tilde{q}^t,\mathcal{D}_i,\mathcal{X}_i)$ and processes it using its personalized privacy rule repository $\Omega_{i,\mathrm{local}}^t$ before uploading $\tilde{r}_i^t$. Both $r_i^t$ and $\Omega_{i,\mathrm{local}}^t$ remain on device. The server evaluates the uploaded responses, updates the routing states to $M_i^{t+1}$, and aggregates reliable responses to obtain $z^t$. Meanwhile, global and personalized privacy rules co-evolve through generalized local rule sharing and selective global to local adaptation. If the task remains unresolved, the server refines the query using the updated routing states and applies the updated global privacy rules before dispatching it to selected clients in the next round. The process ends when no further clients are needed or the round budget is exhausted.

\subsection{Iterative Orchestration for Federated MAS}
Building on the overview in Section~\ref{section:preliminaries}, this section details client selection, response evaluation and aggregation, and the reward used to train the orchestrator.

\subsubsection{Agent Selection and Routing State Updates}
PFedMA first identifies the expertise domain $\rho_d$ most relevant to $\tilde q^t$ as $d_\star^t=\arg\max_{d\in\mathcal Domain}\mathrm{Cos}(\tilde q^t,\rho_d)$, and then retrieves the corresponding candidate set $\mathcal{A}_{d_\star^t}$. The orchestrator then selects participating agents as 
\begin{equation}
    \mathcal{E}^{t} = \pi_{\theta}\Big( \mathcal{A}_{d_\star^t}\mid\tilde q^{t},z^{t-1},\{M_i^{t}\}_{A_i\in\mathcal{A}_{d_\star^t}}\Big).
\end{equation}
Where $M_i^t=[\mathbf{C}_i^t\Vert S_i^t]$ maintains the historical capability  and reliability states of client $A_i$. After the returned response of $A_i$ is evaluated, its routing state is updated using the newly observed historical capability and reliability feedback
\begin{equation}
    \mathbf{C}_i^{t+1}
=
\mathbf{C}_i^{t}
+
f_i^{t}\bar{\mathbf{v}}_i^{t},
\qquad
S_i^{t+1}
=
\frac{
n_i^t S_i^t+
[\mathrm{Int}_i^t,\mathrm{Evd}_i^t,\mathrm{Con}_i^t]^\top
}{
n_i^t+1
}, 
\qquad
A_i \in \widehat{\mathcal E}^t
\end{equation}
where $\widehat{\mathcal E}^t$ denotes the set of responses retained by the reliability filter, $n_i^t$ is the number of previously evaluated responses from $A_i$, and $\mathrm{Int}_i^t$, $\mathrm{Evd}_i^t$, and $\mathrm{Con}_i^t$ denote its anomaly scores.
We define $\mathbf C_i^t$ as accumulated historical capability state of client $A_i$, where $f_i^t$ indicates whether the current retained response improves upon the most recent retained response of the same client, and $\bar{\mathbf{v}}_i^{t}$ identifies the capability dimensions in which the current response improves upon that reference. In contrast, $S_i^t$ summarizes the agent's historical risk as the mean of its observed anomaly scores.

The updated routing state is $M_i^{t+1}=\left[\mathbf{C}_i^{t+1}\Vert S_i^{t+1}\right].$ Thus, response evaluations from the current round are retained in $M_i^{t+1}$ and directly influence agent selection in subsequent rounds. Detailed state update definitions are provided in Appendix~\ref{section:agent_routing_state_update}.

\subsubsection{Response Reliability Evaluation and Filtered Aggregation}
\label{section:robust_aggregation}

Similar to the recently proposed decision centered model Jev~\citep {almeida2026jev}, PFedMA explicitly presents task level results rather than burying them in free form responses. While Jev focuses on decision making, PFedMA focuses on conclusions and records the claims and evidence corresponding to each conclusion.

At round $t$, the privacy processed response of client $A_i$ is represented as
$\label{eq:response_format}
\tilde r_i^t=\left\langle a_i^t,\Gamma_i^t,\mathcal{L}_i^t \right\rangle,$
where $a_i^t$ denotes the final conclusion, $\Gamma_i^t$ denotes the collection of claims supporting the conclusion, and $\mathcal{L}_i^t$ denotes the corresponding collection of evidence associated with these claims. The details are provided in Appendix~\ref{section:Structured_Response_Format}.

PFedMA evaluates each response from the complementary perspectives of intent, evidence, and cross agent consensus. To aggregate fine grained anomaly signals, we use
\begin{equation}
\label{eq:anomaly_pool}
\operatorname{Pool}_{\lambda_{pool}}
(\{x_j\}_{j=1}^{J})
=
\lambda_{pool}\max_j x_j
+
(1-\lambda_{pool})\frac{1}{J}\sum_{j=1}^{J}x_j, 
\end{equation}
where $J$ denotes the number of fine grained anomaly scores being aggregated and $x_j$ is the $j$-th score. The operator balances the most severe anomaly with the overall anomaly level. This score captures abnormal intents such as instruction overriding, false authority claims, aggregator manipulation, and task goal substitution. The evidence anomaly $\mathrm{Evd}_i^t$ assesses claim level evidence quality from support, verifiability, and coherence, while the consensus anomaly $\mathrm{Con}_i^t$ measures semantic deviation from peer conclusions. Intent and evidence are evaluated by a frozen judge model, whereas consensus is computed deterministically; details are provided in Appendix~\ref{section:anomaly_tails} and Appendix~\ref{section:evaluator_implementation}.

The three anomaly dimensions are combined into an overall anomaly score, and task responses exceeding the tolerance threshold are excluded:

\begin{equation}
\label{eq:overall_anomaly_filtering}
\mathcal Risk_i^t
=
\mathrm{Int}_i^t
+
\mathrm{Evd}_i^t
+
\alpha_{\mathrm{con}}\mathrm{Con}_i^t,
\qquad
\widehat{\mathcal{E}}^t
=
\left\{
A_i \in \mathcal{E}^t
\mid
\mathcal Risk_i^t \leq \tau_{\mathrm{risk}}
\right\}.
\end{equation}

where $\alpha_{\mathrm{con}}$ controls the contribution of consensus anomaly, and $\tau_{\mathrm{risk}}$ denotes the anomaly tolerance threshold. A relatively small $\alpha_{\mathrm{con}}$ is used because disagreement alone does not necessarily indicate an unreliable response. The retained set $\widehat{\mathcal{E}}^t$ is then used for aggregation.

The retained privacy processed responses are aggregated to update the accumulated collaboration state:
\begin{equation}
\label{eq:robust_fusion}
z^t
=
\begin{cases}
\operatorname{Fusion}
\left(
\{\tilde r_i^t\}_{A_i\in\widehat{\mathcal{E}}^t},
z^{t-1}
\right),
& \widehat{\mathcal{E}}^t\neq\emptyset,\\[4pt]
z^{t-1},
& \widehat{\mathcal{E}}^t=\emptyset.
\end{cases}
\end{equation}
In our implementation, $\operatorname{Fusion}$ is realized by the server side orchestrator $\pi_\theta$, which synthesizes the retained privacy processed responses and the previous global reasoning state into $z^t$.

\subsubsection{Composite Reward for Multi-Round Orchestration}
\label{section:reward}

During RL training, PFedMA uses a composite reward that balances final answer correctness, answer quality improvements across rounds, response reliability, and agent invocation cost. For a collaboration trajectory $\tau$ containing $T$ routing rounds, the total reward is
\begin{equation}
\label{eq:rl_reward}
R(\tau)
=
R_{\mathrm{fmt}}(\tau)
+
R_{\mathrm{ans}}(\tau)
+
\sum_{t=1}^{T}R_{\mathrm{step}}^t,
\end{equation}
where $R_{\mathrm{fmt}}$ evaluates format compliance and
$R_{\mathrm{ans}}$ evaluates the final answer. Details for these two formulas are provided in Appendix~\ref{sec:Format and final answer rewards}. Meanwhile, $R_{\mathrm{step}}^t$ delivers intermediate feedback for routing, with the step wise reward defined as
\begin{equation}
\label{eq:trajectory_reward}
R_{\mathrm{step}}^t
=
\Delta Q_t
-
\overline{\mathcal Risk}_t
-
\frac{|\mathcal E^t|}{K_{\max}},
\qquad
\overline{\mathcal Risk}_t
=
\frac{1}{|\mathcal E^t|}
\sum_{A_i\in\mathcal E^t}
\frac{\mathcal Risk_i^t}{2+\alpha_{\mathrm{con}}},
\qquad
|\mathcal E^t|>1.
\end{equation}
Here, $\Delta Q_t=Q_t-Q_{t-1}$ records the change in answer quality from $z^{t-1}$ to $z^t$, while $\overline{\mathcal Risk}_t$ measures the average normalized response risk of the agents invoked at round $t$. $K_{\max}$ denotes the maximum number of agents that can be invoked in a single round, and $|\mathcal E^t|/K_{\max}$ represents the fraction of the invocation budget used by the current routing decision. Detailed definitions and RL settings are provided in Appendix~\ref{sec:Cross-round reasoning progress}.

\subsection{Federated Privacy Rule Co-Evolution Mechanism}
\label{section:privacy}

Existing privacy rule based data minimization approaches for Federated MAS mainly regulate information disclosure according to agent roles and access requirements ~\citep{shi2025privacy}, but do not address how client specific privacy knowledge can be reused across the federation without centralizing personalized policies. To bridge this gap, PFedMA introduces a \emph{federated privacy rule co-evolution} mechanism, in which shareable privacy knowledge is propagated globally while complete client specific privacy rule repositories remain local.

\textbf{Global Privacy Rule Repository $\Omega^{t}_{\mathrm{global}}$}. The global privacy rule repository is maintained by the server to preserve privacy knowledge that is shareable across clients. At round $t$, the server receives generalized rule candidates $\bar{\omega}_i^t$ from clients and evaluates their cross client semantic support. When multiple candidates are available, candidates achieving the highest positive support proportion are selected to form the global candidate set:
\begin{equation}
\mathcal G^t
=
\left\{
\bar{\omega}_i^t\in\mathcal B^t
\;\middle|\;
\zeta_i^t>0,\;
\zeta_i^t=
\displaystyle
\max_{\bar{\omega}_j^t\in\mathcal B^t}\zeta_j^t
\right\},
\qquad
|\mathcal B^t|>1.
\end{equation}
Here, $\mathcal B^t$ is the candidate pool, and $\zeta_i^t=\operatorname{Supp}^t(\bar{\omega}_i^t)/(|\mathcal B^t|-1)$ is the support proportion of candidate $\bar{\omega}_i^t$. The set $\mathcal G^t$ retains all candidates with the highest positive support proportion. Details are provided in Appendix~\ref{section:privacy_details}. The server also collects compact rule level feedback $\mathcal F^t$ from previous rule executions and uses it together with the promoted candidates to update the global repository:
\begin{equation}
\Omega_{\mathrm{global}}^{t+1}
=
\operatorname{RuleUpdate}
\left(
\Omega_{\mathrm{global}}^t,
\mathcal G^t,
\mathcal F^t
\right).
\end{equation}
This update consolidates shareable rules and resolves conflicting or ineffective global rules according to accumulated client feedback. The updated repository is then used by the server to process the next query before dispatch:
$
\tilde q^{t+1}
=
P_{\mathrm{in}}
\left(
q^{t+1},
\Omega_{\mathrm{global}}^{t+1}
\right).
$
The resulting privacy rule processed query $\tilde q^{t+1}$ is then dispatched to the selected clients for local inference, thereby transferring the updated global privacy constraints to the next round of collaboration. Detailed definitions and update procedures are provided in Appendix~\ref{section:privacy_details}.

\textbf{Personalized Privacy Rule Repository $\Omega^{t}_{i,\mathrm{local}}$}. Upon receiving the privacy processed query $\tilde q^{t+1}$, client $A_i$ performs local inference using its private data and accumulated experience $r_i^{t+1}=A_i\left(\tilde q^{t+1},\mathcal D_i,\mathcal X_i\right)$. $\Phi_i^t(\omega)$ denotes the local utility score of privacy rule $\omega$ for client $A_i$ at communication round $t$, and is defined as follows:

\begin{equation}
\begin{aligned}
\Phi_i^t(\omega)
=\;&
\operatorname{Rel}
\left(
\omega,\mathcal X_i
\right)
+
\frac{n_{i,\mathrm{valid}}^t(\omega)}
{\max\left(1,n_{i,\mathrm{trigger}}^t(\omega)\right)}
-
\frac{n_{i,\mathrm{conflict}}^t(\omega)}
{\max\left(1,n_{i,\mathrm{trigger}}^t(\omega)\right)}
\end{aligned}
\end{equation}

Here, the term $\operatorname{Rel}(\omega,\mathcal{X}_i)$ measures the semantic relevance of rule $\omega$ to the local experience buffer $\mathcal{X}_i$. Furthermore, $n_{i,\mathrm{trigger}}^t(\omega)$, $n_{i,\mathrm{valid}}^t(\omega)$, and $n_{i,\mathrm{conflict}}^t(\omega)$ denote the numbers of times that rule $\omega$ is triggered, successfully enforced, and found to conflict with a local privacy requirement, respectively. The details are provided in Appendix~\ref{sec:Local Utility Evaluation and Rule Selection.}.

Meanwhile, the updated global rules are redistributed to the client for personalized adaptation. Rather than directly replacing its local rules, client $A_i$ compares each redistributed global rule $\omega_g^{t+1}$ with its semantically corresponding local rule $\omega_i^t$ according to their local utility and selects

\begin{equation}
\begin{aligned}
\omega_i^\star
&=
\begin{cases}
\omega_i^t,
& \Phi_i^t(\omega_i^t)
\geq
\Phi_i^t(\omega_g^{t+1}),\\
\omega_g^{t+1},
& \text{otherwise},
\end{cases}
\qquad
\Omega_{i,\mathrm{local}}^{t+1}
=
\Omega_{i,\mathrm{local}}^t
\cup_{merge}
\{\omega_i^\star\}.
\end{aligned}
\label{eq:local_privacy_update}
\end{equation}
Here, $\omega_g^{t+1} \in \Omega_{\mathrm{global}}^{t+1}$, with a detailed definition provided in Appendix~\ref{section:privacy_details}. This selective adaptation allows global privacy knowledge to refine the personalized repository without overriding client specific alternatives. The personalized repository is applied to each locally generated response before transmission 
$
\tilde{r}_i^{t+1}
=
P_{\mathrm{resp},i}
\left(
r_i^{t+1},
\Omega_{i,\mathrm{local}}^{t+1}
\right).
\label{eq:local_response_processing}
$ If a previously uncovered privacy constraint is detected, the corresponding rule $\omega_i^{t+1}\in\Omega_{i,\mathrm{local}}^{t+1}$ remains on device, and only its generalized representation is shared 
$
\bar{\omega}_i^{t+1}
=
\mathcal{P}_i^{\mathrm{gen}}
\left(
\omega_i^{t+1}
\right).
\label{eq:rule_generalization}
$
Here, $\mathcal{P}_i^{\mathrm{gen}}$ denotes the client side private rules generalization operator, and $\bar{\omega}_i^{t+1}$ is the shareable rule representation. The generalized rule is sent to the server for the subsequent global privacy rule updates; see Appendix~\ref{section:privacy_details}.

%% file: context/change/experiments_change.tex
\section{Experiments}
\label{section:Experimental setting}
In the experiments, we seek to answer the following research questions:

    \textbf{RQ1:} Does cross round response feedback improve multi-round federated reasoning over feedback free orchestration?

    \textbf{RQ2:} Can federated privacy rule co-evolution enable cross client privacy rule coordination and personalized adaptation while maintaining rule enforcement effectiveness and task utility?

    \textbf{RQ3:} Can multi-factor reliability evaluation improve robustness against unreliable task responses while preserving benign task utility?

\subsection{Experimental Setup}

\textbf{Models.}
Following KABB~\citep{zhang2025kabb}, we construct 20 agent profiles across 12 knowledge domains and distribute them over $N$ federated clients. Each client hosts a local agent with private data, an experience buffer, and personalized privacy rules, while multiple clients may share the same profile. Agent profiles use domain specific prompts with one of seven LLM backbones: MiniMax2.5~\citep{chen2026minimax}, DeepSeek-V3~\citep{liu2024deepseek}, Qwen2-72B-Instruct~\citep{bai2023qwen}, GPT-4o~\citep{hurst2024gpt}, DeepSeek-R1~\citep{guo2025deepseek}, LLaMA-3-70B-Instruct~\citep{adams2024llama}, or Kimi-K2.5~\citep{team2026kimi}.

\textbf{Dataset and Evaluation Metrics.}
We evaluate PFedMA on mathematical reasoning benchmarks, including GSM8K~\citep{cobbe2021training}, MATH~\citep{hendrycks2021measuring}, and AIME24~\citep{muennighoff2025s1}; clinical QA benchmarks, including MedQA and MedMCQA~\citep{tang2024medagents}; instruction following with AlpacaEval 2.0~\citep{zhang2025kabb}; and privacy sensitive web agent tasks with SeeAct on Mind2Web-SC~\citep{xiang2025guardagent}.
We report ASR, RR, PNA, and NRP following ASB~\citep{zhang2024agent}, together with BP to evaluate benign task utility with defense mechanisms enabled, together with LPA, LPP, LPR, EA, FRA, and LR for privacy evaluation~\citep{xiang2025guardagent,shao2024privacylens}.
Metric definitions are provided in Appendix~\ref{sec:evaluation_metrics_details}.

\textbf{Baselines.}
We conduct a comparative analysis of the PFedMA algorithm against other MAS methods. For the single agent baseline, the primary goal is to demonstrate that our method’s performance is not due to the server LLM model, but rather to the effectiveness of the federated MAS. For federated MAS, we compare with Fed-ICL~\citep{wang2025federated} and EPEAgents~\citep{shi2025privacy}. For centralized MAS, we compare with Router-R1~\citep{zhang2025router}, KABB~\citep{zhang2025kabb}, and OPTAGENT~\citep{bi2025optagent}. All multi-agent methods use the same pool of 20 experts and the same number of iteration rounds to isolate the effect of orchestration and aggregation.

\subsection{Main Results and Analysis}

\begin{table}[htbp]
    \centering
    \caption{Performance and privacy comparison of different frameworks. Best results are shown in \textbf{bold}, and second best results are \underline{underlined}. For Avg calculation, lower is better metrics are converted using $100-score$ before averaging.}
    \label{tab:framework_results}
    \footnotesize
    \setlength{\tabcolsep}{2pt}
    \renewcommand{\arraystretch}{1.05}

    \resizebox{0.98\linewidth}{!}{
    \begin{tabular}{lccccccc}
    \toprule
    \rowcolor{gray!15}
     \multicolumn{8}{c}{\textbf{RQ1: Performance comparison across multiple benchmarks}}  \\
       
         Method & Math & GSM8K & AIME24 & MedQA & AlpacaEval 2.0 & MedMCQA & Avg \\
        \midrule
    Single Agent & $42.63_{\pm3.51}$ & $79.11_{\pm1.39}$ & $6.67_{\pm0.73}$ & $63.35_{\pm4.31}$ & $58.51_{\pm0.18}$ & $57.82_{\pm0.21}$ & $51.35_{\pm1.72}$\\

    Router-R1 & $76.06_{\pm2.73}$ & $84.02_{\pm2.91}$ & $73.40_{\pm4.34}$ & $75.61_{\pm1.71}$ & $\underline{75.43_{\pm3.94}}$ & $69.57_{\pm1.07}$ & $75.68_{\pm2.78}$\\
    KABB  & $77.51_{\pm1.12}$ & $85.66_{\pm0.60}$ & $71.93_{\pm3.00}$ & $\bm{78.15_{\pm0.73}}$ & $72.38_{\pm4.30}$ & $\bm{72.05_{\pm1.70}}$ & $76.28_{\pm1.91}$\\
    OPTAGENT & $75.02_{\pm1.27}$ & $\underline{86.26_{\pm4.31}}$ & $62.43_{\pm2.43}$ & $66.27_{\pm1.21}$ & $69.01_{\pm1.78}$ & $64.82_{\pm0.91}$ & $70.64_{\pm{1.99}}$\\

    EPEAgents & $70.03_{\pm0.39}$ & $80.10_{\pm1.76}$ & $68.57_{\pm3.16}$ & $77.20_{\pm1.99}$ & $70.88_{\pm1.36}$ & $63.54_{\pm0.81}$ & $71.72_{\pm1.58}$ \\
    Fed-ICL   & $\underline{79.33_{\pm0.83}}$ & $85.04_{\pm1.04}$ & $\underline{74.78_{\pm1.51}}$ & $76.03_{\pm3.00}$ & $73.51_{\pm0.29}$ & $69.34_{\pm0.93}$ & $76.34_{\pm1.27}$\\
    \cellcolor{gray!15}PFedMA   
    & \cellcolor{gray!15}$\bm{84.53_{\pm2.51}}$ &\cellcolor{gray!15} $\bm{86.60_{\pm0.81}}$ &\cellcolor{gray!15} $\bm{76.67_{\pm1.83}}$ &\cellcolor{gray!15} $\underline{77.91_{\pm0.84}}$ 
    &\cellcolor{gray!15} $\bm{77.65_{\pm0.18}}$ &\cellcolor{gray!15} $\underline{69.80_{\pm0.53}}$ &\cellcolor{gray!15} $\bm{78.86_{\pm1.12}}$\\
    \bottomrule

    \toprule
    \rowcolor{gray!15}
     \multicolumn{8}{c}{\textbf{RQ2: Evaluation of privacy processed and performance}}  \\
        Method 
        & LR $\downarrow$ 
        & LPR $\uparrow$ 
        & LPA $\uparrow$ 
        & LPP $\uparrow$ 
        & EA  $\uparrow$ 
        & FRA $\uparrow$ 
        & Avg $\uparrow$ \\
        \midrule
        Single Agent  & $43.40_{\pm0.62}$ & $69.00_{\pm1.14}$ & $83.50_{\pm0.87}$ & $98.60_{\pm0.35}$ & $68.00_{\pm1.26}$ & $62.50_{\pm0.94}$ & $73.03_{\pm0.86}$ \\
        Router-R1   & $39.08_{\pm0.78}$ & $73.73_{\pm1.31}$ & $85.70_{\pm0.69}$ & $96.90_{\pm0.52}$ & $73.70_{\pm1.08}$ & $73.05_{\pm0.83}$ & $77.33_{\pm0.87}$ \\
        KABB     & $35.20_{\pm0.91}$ & $81.64_{\pm0.74}$ & $90.00_{\pm0.58}$ & $98.10_{\pm0.41}$ & $76.50_{\pm1.17}$ & $69.50_{\pm0.96}$ & $80.09_{\pm0.80}$ \\
        OPTAGENT    & $37.30_{\pm0.85}$ & $80.04_{\pm1.09}$ & $90.00_{\pm0.63}$ & $95.32_{\pm0.22}$ & $\bm{80.00_{\pm0.48}}$ & $72.10_{\pm0.77}$ & $80.03_{\pm0.67}$ \\
        EPEAgents   & $33.30_{\pm0.66}$ & $\underline{86.50_{\pm0.82}}$ & $\underline{93.37_{\pm0.54}}$ & $\underline{99.01_{\pm0.19}}$ & $70.50_{\pm1.21}$ & $\underline{75.63_{\pm0.73}}$ & $\underline{81.95_{\pm0.69}}$ \\ 
        Fed-ICL     & $\underline{32.60_{\pm0.71}}$ & $84.00_{\pm0.95}$ & $92.00_{\pm0.61}$ & $97.24_{\pm0.17}$ & $75.60_{\pm1.04}$ & $73.30_{\pm0.88}$ & $81.59_{\pm0.73}$ \\

        \cellcolor{gray!15}
        PFedMA  & \cellcolor{gray!15}$\bm{26.70_{\pm0.43}}$ &\cellcolor{gray!15} $\bm{89.30_{\pm0.69}}$ & \cellcolor{gray!15}$\bm{94.70_{\pm0.38}}$ &\cellcolor{gray!15} $\bm{99.86_{\pm0.12}}$ &\cellcolor{gray!15} $\underline{77.00_{\pm0.97}}$ & \cellcolor{gray!15}$\bm{79.71_{\pm0.55}}$ & \cellcolor{gray!15}$\bm{85.65_{\pm0.52}}$ \\
        \bottomrule

    \end{tabular}
    }
\end{table}
\vspace{-10pt}
\begin{table}[htbp]
    \centering
    \caption{Robustness of different frameworks under adversarial attacks}
    \label{tab:security_result}
    \footnotesize
    \setlength{\tabcolsep}{2pt}
    \renewcommand{\arraystretch}{1.05}

    \resizebox{0.98\linewidth}{!}{
    \begin{tabular}{lcccccc}
    \toprule
    \rowcolor{gray!15}
     \multicolumn{7}{c}{\textbf{RQ3: Evaluation of system security and robustness}}  \\
      Method & ASR $\downarrow$ & RR $\uparrow$ & BP $\uparrow$ & PNA $\uparrow$ & NRP $\uparrow$ & Avg $\uparrow$ \\
\midrule
Single Agent & $54.82_{\pm1.23}$ & $5.34_{\pm2.41}$ & $\underline{77.75_{\pm3.84}}$ & $66.57_{\pm2.03}$ & $30.08_{\pm3.20}$ & $44.98_{\pm2.54}$ \\

Router-R1  
& $52.07_{\pm0.81}$  
& $9.10_{\pm0.36}$  
& $76.73_{\pm0.74}$  
& $76.70_{\pm0.65}$  
& $36.76_{\pm0.88}$ 
& $49.44_{\pm0.69}$ \\

KABB      
& $36.20_{\pm0.69}$  
& $18.50_{\pm0.54}$  
& $75.29_{\pm0.83}$  
& $\bm{88.00_{\pm0.47}}$  
& $56.14_{\pm0.76}$ 
& $60.35_{\pm0.66}$ \\

OPTAGENT  
& $50.07_{\pm0.77}$  
& $11.50_{\pm2.49}$  
& $69.40_{\pm0.91}$  
& $73.54_{\pm0.58}$  
& $36.72_{\pm1.02}$ 
& $48.22_{\pm1.15}$ \\

EPEAgents 
& $\underline{32.01_{\pm0.52}}$  
& $\underline{22.13_{\pm0.61}}$  
& $76.57_{\pm1.72}$  
& $82.84_{\pm0.55}$  
& $\underline{56.32_{\pm0.69}}$ 
& $\underline{61.17_{\pm0.82}}$ \\

Fed-ICL   
& $49.30_{\pm0.84}$  
& $13.00_{\pm0.43}$  
& $74.75_{\pm3.79}$  
& $74.71_{\pm1.66}$  
& $37.88_{\pm1.73}$ 
& $50.21_{\pm1.69}$ \\

\cellcolor{gray!15}PFedMA    
& \cellcolor{gray!15}$\bm{30.80_{\pm0.45}}$  
& \cellcolor{gray!15}$\bm{24.60_{\pm0.58}}$  
& \cellcolor{gray!15}$\bm{78.80_{\pm0.63}}$  
& \cellcolor{gray!15}$\underline{85.50_{\pm0.49}}$  
& \cellcolor{gray!15}$\bm{59.17_{\pm0.71}}$ 
& \cellcolor{gray!15}$\bm{63.45_{\pm0.57}}$ \\
\bottomrule
    \end{tabular}
    }
\end{table}

\textbf{Cross Round Feedback and Iterative Orchestration (RQ1).}

\begin{figure}[htbp]
    \begin{subfigure}[t]{0.23\linewidth}
        \centering
        \includegraphics[width=\linewidth]{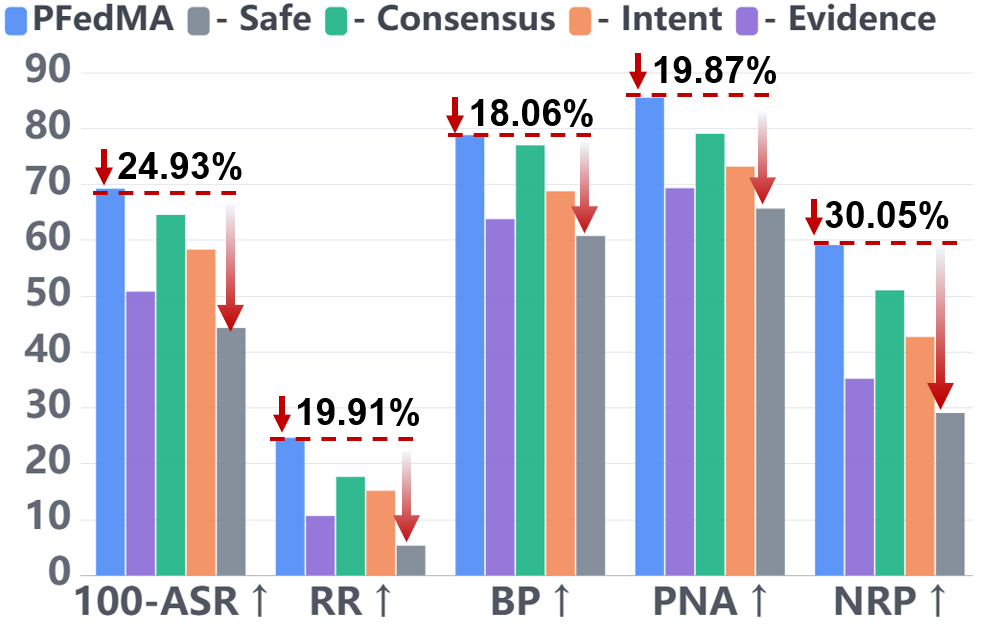}
        \caption{Ablation of structured reliability evaluation}
        \label{fig:ablation_safe}
    \end{subfigure}
    \hfill
    \begin{subfigure}[t]{0.23\linewidth}
        \centering
        \includegraphics[width=\linewidth]{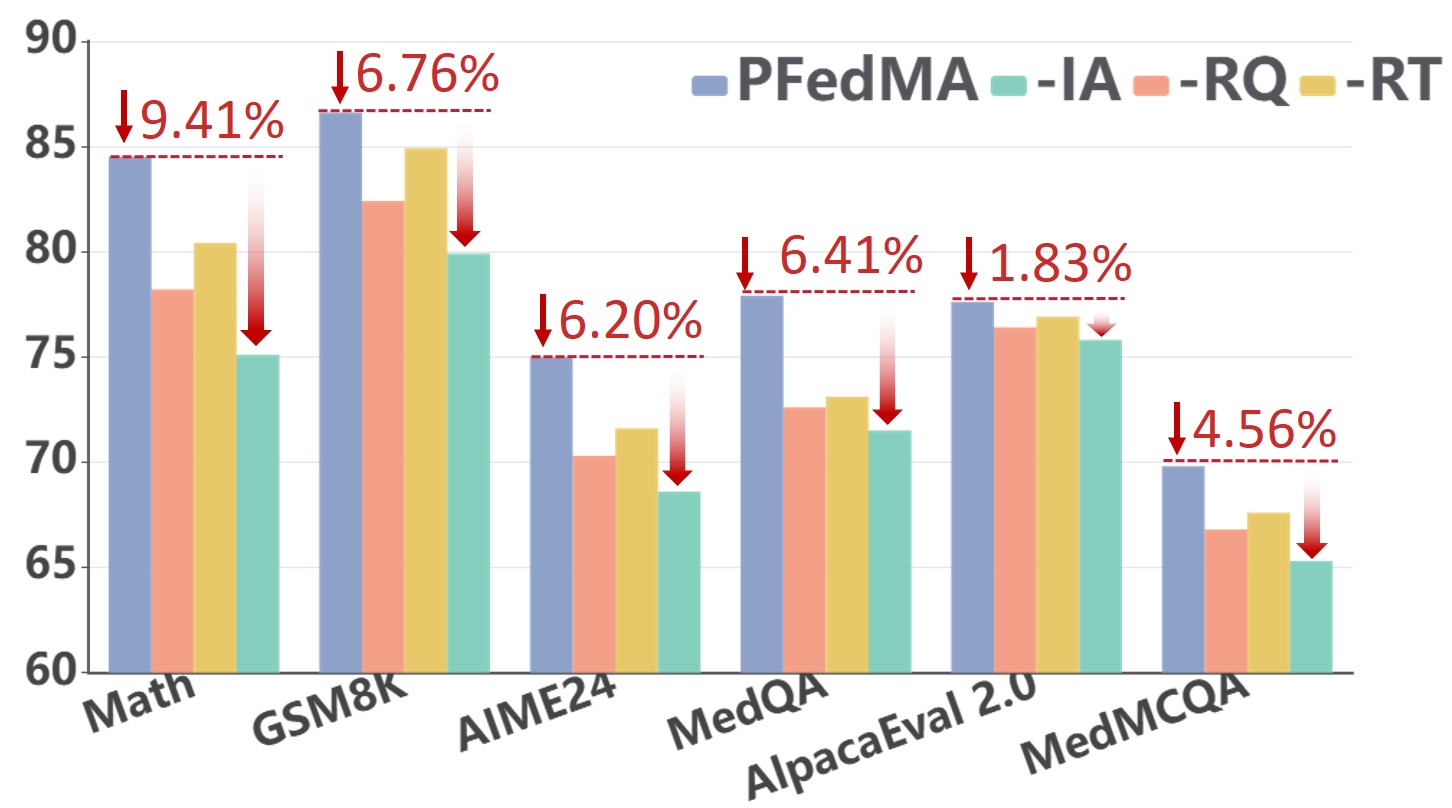}
        \caption{Feedback Driven Orchestration Ablation}
        \label{fig:benchmark_results}
    \end{subfigure}
    \hfill
    \begin{subfigure}[t]{0.23\linewidth}
        \centering
        \includegraphics[width=0.7\linewidth]{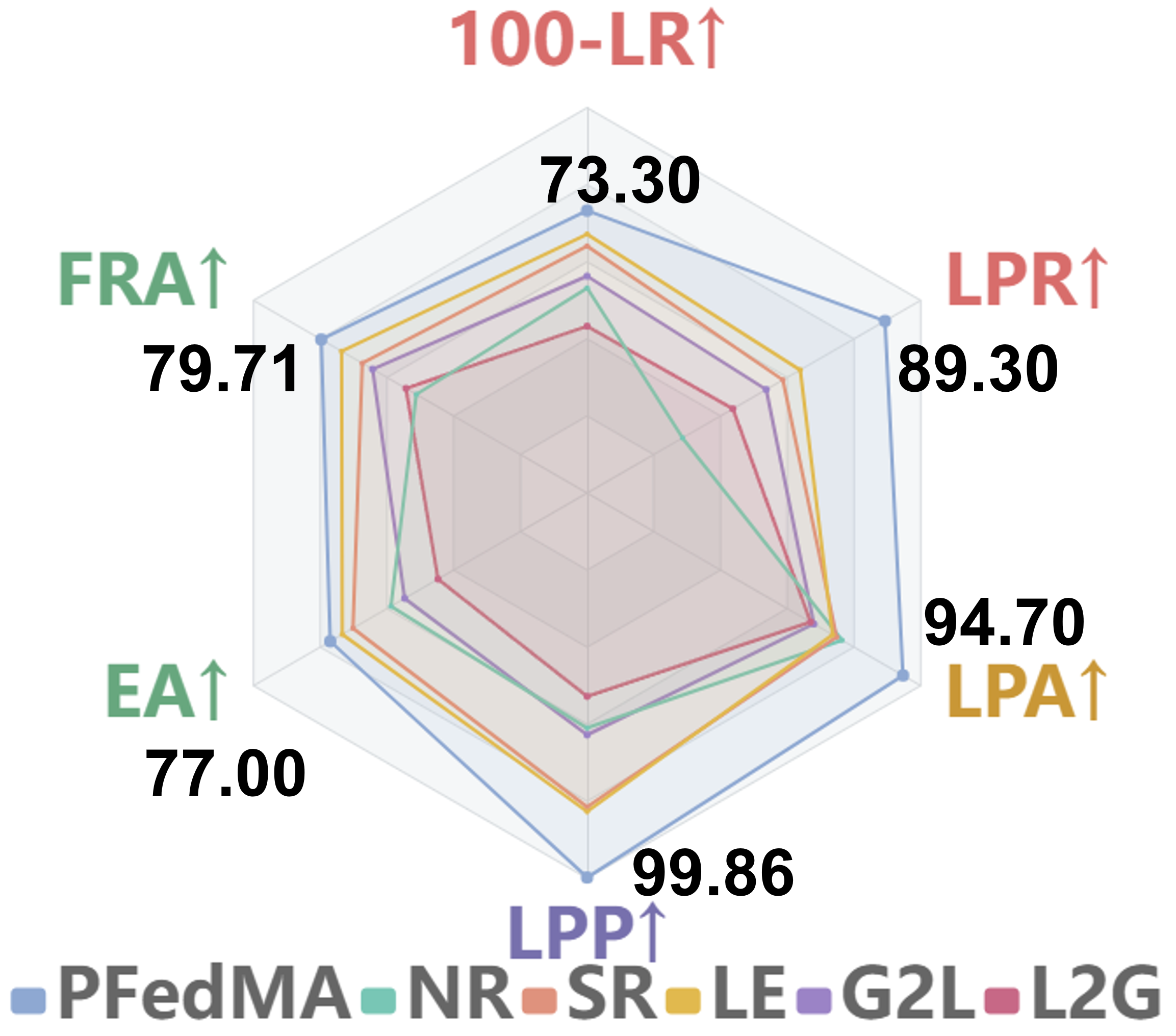}
            \caption{Federated Rule Co-Evolution Ablation}
            \label{fig:ablation_private}
    \end{subfigure}
    \hfill
    \begin{subfigure}[t]{0.23\textwidth}
        \centering
        \includegraphics[width=0.9\linewidth]{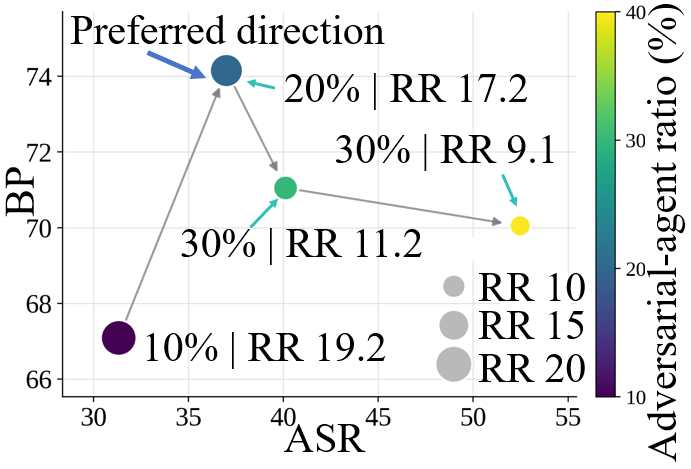}
        \caption{Adversarial Robustness Utility Trade off}
        \label{fig:robustness-utility}
    \end{subfigure}
    \caption{
        Ablation studies of the reliability module, iterative orchestration loop, federated privacy rule co-evolution loop and robustness utility trade off across adversarial ratios.
    }
    \label{fig:blation analysis.}
    
\end{figure}

\begin{figure}[htbp]
    
    \centering

    \begin{subfigure}[b]{0.23\textwidth}
        \centering
        \includegraphics[width=\linewidth]{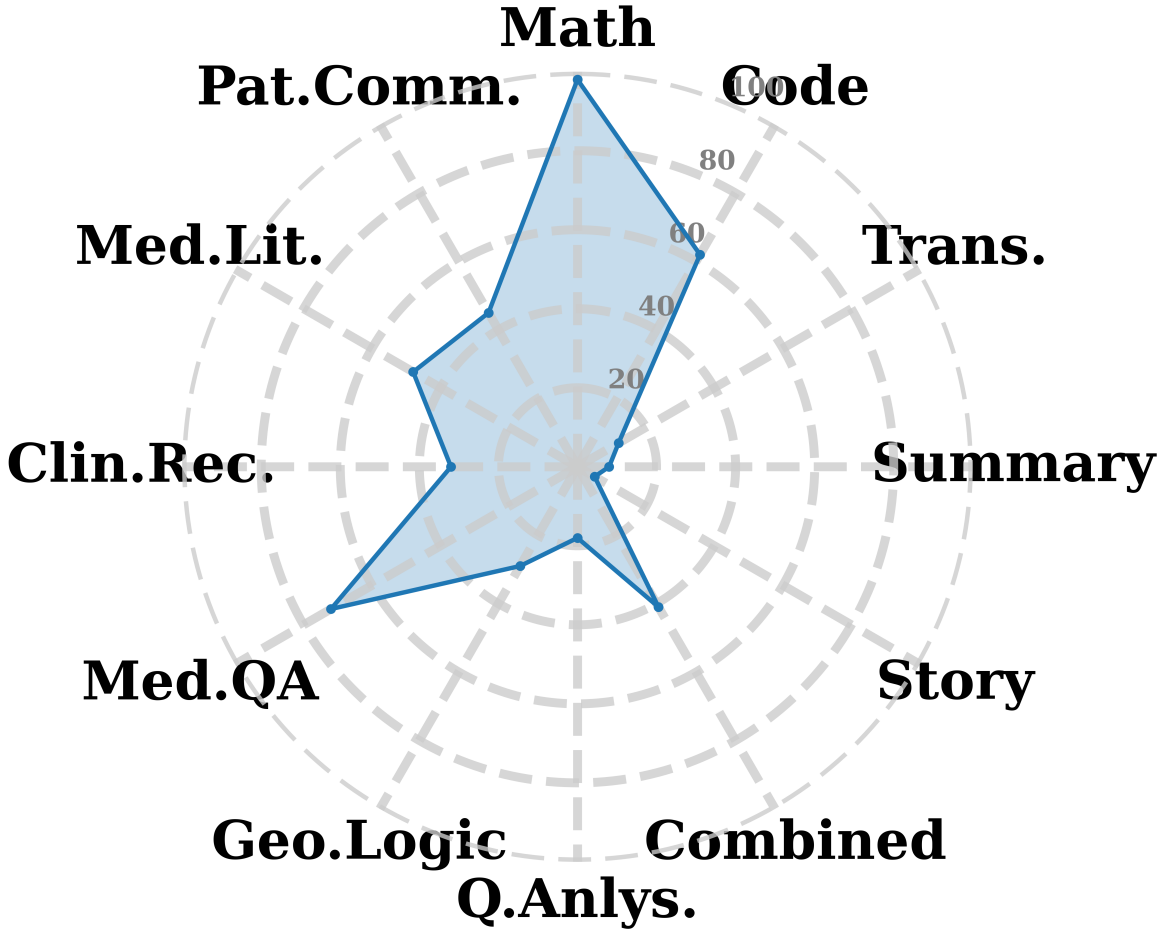}
        \caption{Abilities of Agent 1} 
        \label{fig:radar_expert1}
    \end{subfigure}
    \hfill
    \begin{subfigure}[b]{0.23\textwidth}
        \centering
        \includegraphics[width=\linewidth]{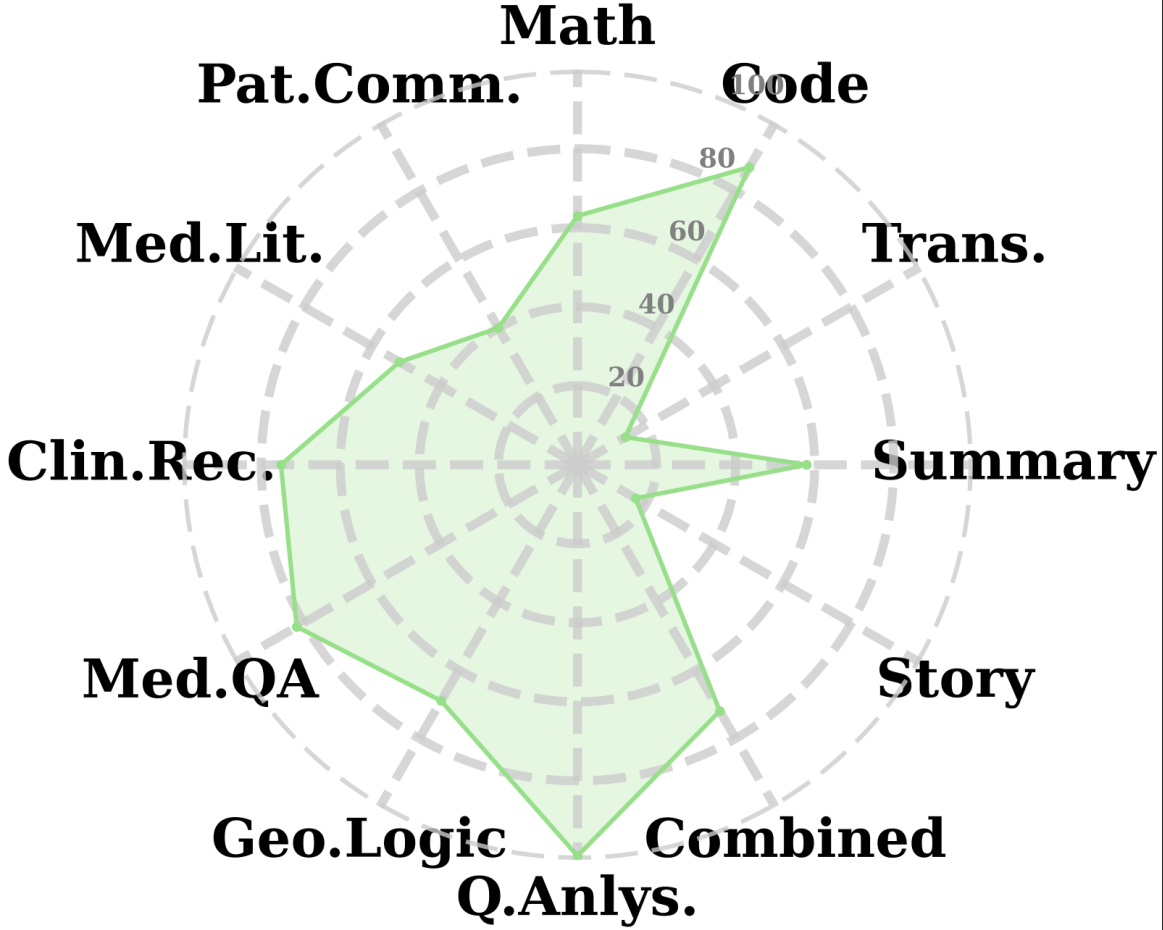}
        \caption{Abilities of Agent 6} 
        \label{fig:radar_expert6}
    \end{subfigure}
    \hfill
    \begin{subfigure}[b]{0.5\textwidth}
        \centering
        \includegraphics[width=\linewidth]{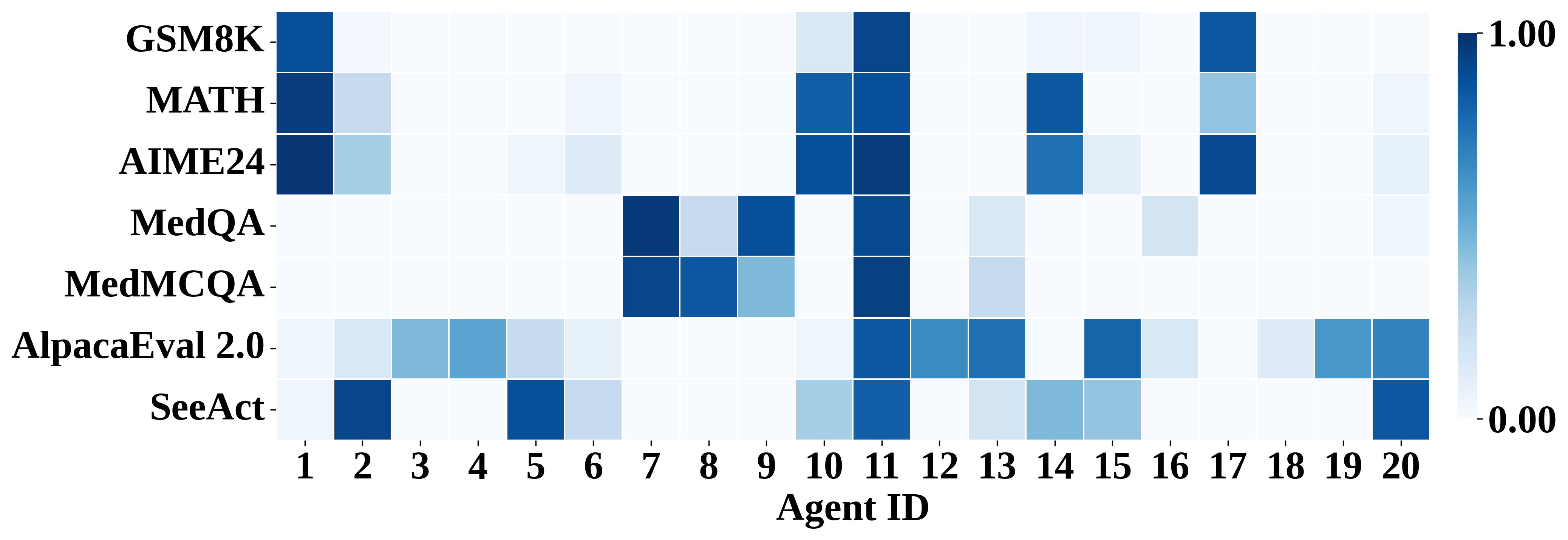} 
        \caption{Agent Selection Heatmap} 
        \label{fig:heatmap_selection}
    \end{subfigure}
    \caption{Agent Abilities and Agent Selection. Full results are given in Appendix \ref{section:Expert Ability}.}
    \label{fig:expert_ability}
    \vspace{-10pt}
\end{figure}

Table~\ref{tab:framework_results} compares PFedMA with representative single-agent, centralized MAS, and federated MAS baselines under the same expert pool and collaboration-round budget. PFedMA achieves the highest average score of $78.86$, outperforming the strongest baseline, Fed-ICL, by $2.52$ points. It ranks first on MATH, GSM8K, AIME24, and AlpacaEval 2.0, and second on MedQA and MedMCQA, indicating strong performance across the evaluated tasks. Figure~\ref{fig:itera_acc} shows that accuracy generally increases across collaboration rounds, suggesting that useful reasoning information is progressively accumulated. The ablation results in Figure~\ref{fig:benchmark_results} further
show that removing query optimization or either historical routing state consistently degrades performance. These results indicate that cross round state evolution, query adaptation, and learned orchestration each contribute to the overall improvement. Figure~\ref{fig:heatmap_selection} provides complementary evidence that PFedMA adaptively selects agents according to task requirements: mathematical agents are selected more frequently for mathematical benchmarks, while medical agents are favored for medical tasks. Overall, these results support the effectiveness of feedback driven iterative orchestration in progressively refining the reasoning state and adapting subsequent collaboration decisions.

\vspace{-15pt}
\begin{figure}[htbp]
    \centering
    \begin{subfigure}[t]{0.48\textwidth}
        \centering
        \includegraphics[width=\linewidth]{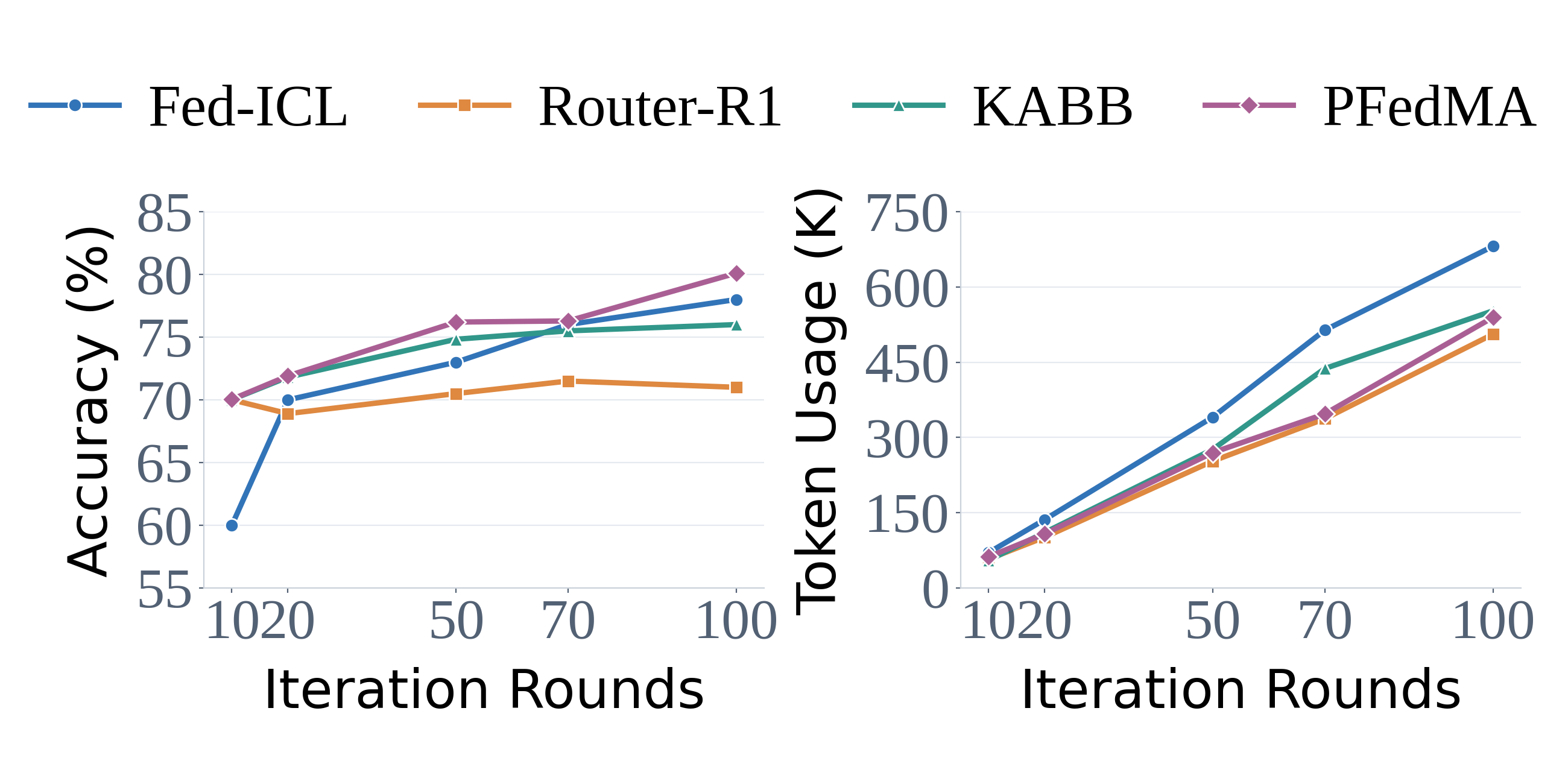}
        \caption{Accuracy and Token Usage by Round}
        \label{fig:training_comparison_reference}
    \end{subfigure}
    \hfill
    \begin{subfigure}[t]{0.48\textwidth}
        \centering
        \includegraphics[width=\linewidth]{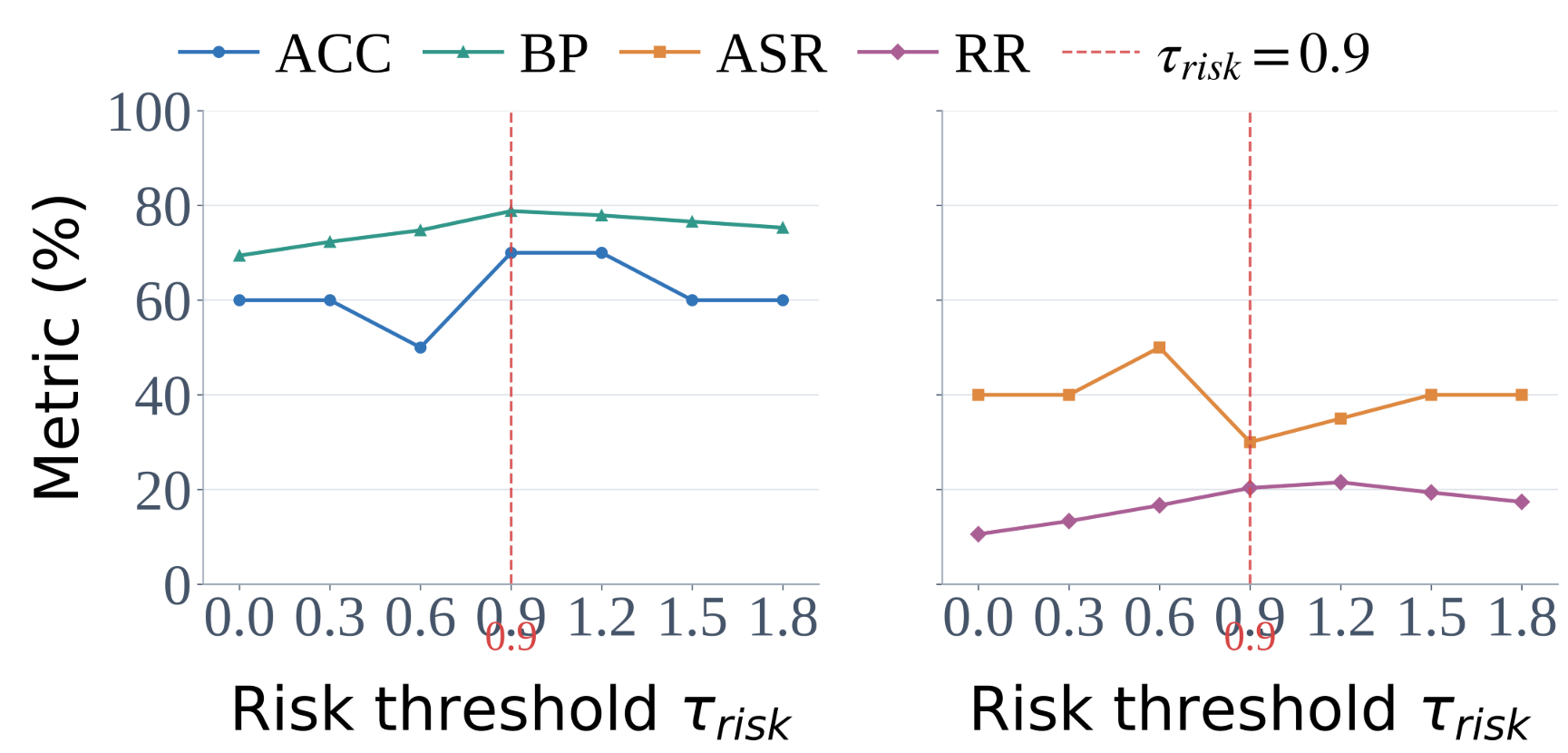}
        \caption{Sensitivity Analysis of the Risk Threshold $\tau$}
        \label{fig:tau_risk_sensitivity}
    \end{subfigure}
    \caption{
        Performance Evaluation under Iterative Optimization and Risk Threshold Variation.}
    \label{fig:performance-robustness}
\end{figure}

\begin{figure}[htbp]
    \centering
    
    \begin{minipage}[b]{0.5\textwidth}
        \centering
        \begin{subfigure}[b]{0.49\linewidth}
            \centering
            \includegraphics[width=\linewidth]{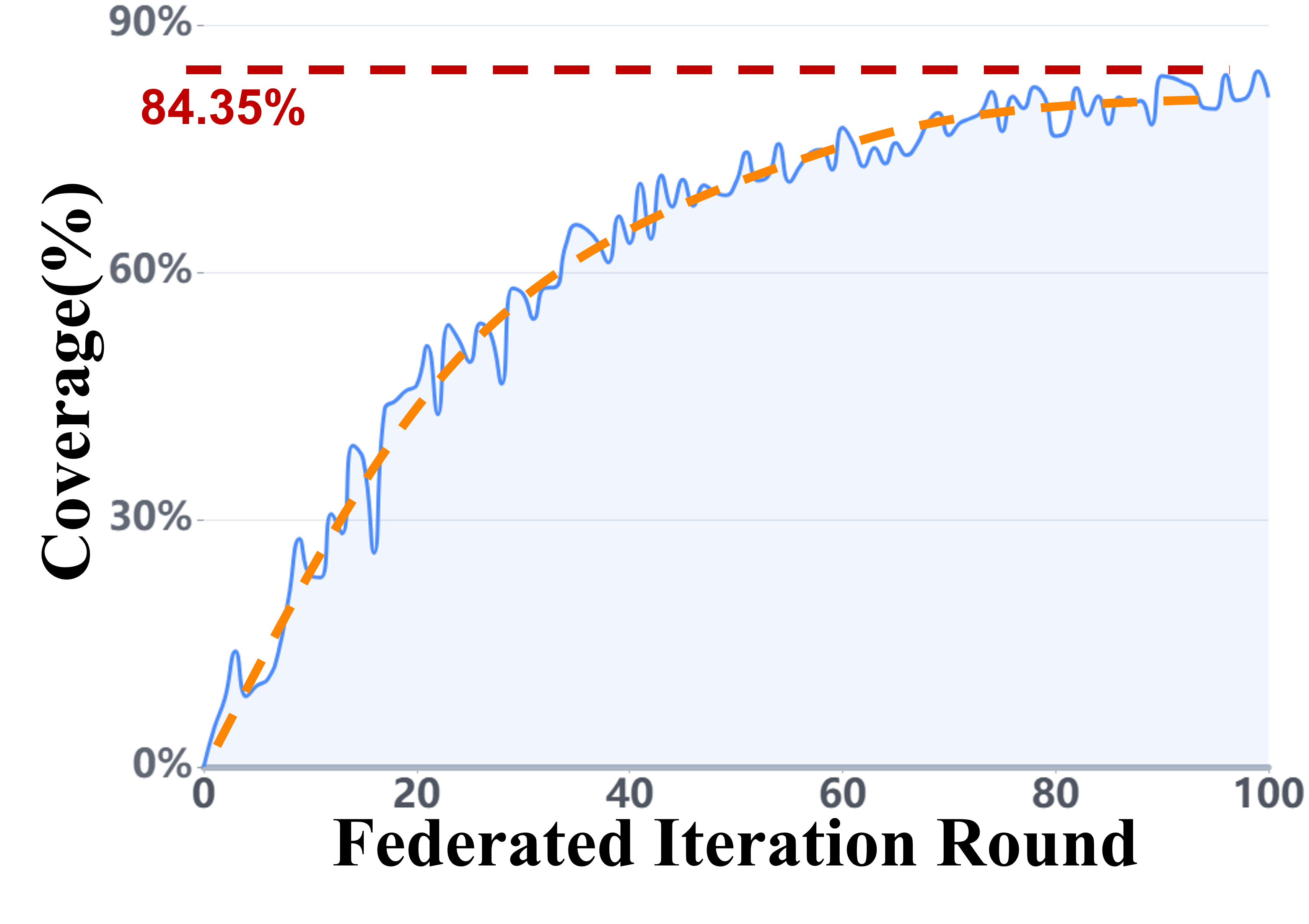}
        \caption{Privacy Coverage Rate}
        \label{fig:privacy_coverage}
        \end{subfigure}
        \hfill
        \begin{subfigure}[b]{0.47\linewidth}
            \centering
            \includegraphics[width=\linewidth]{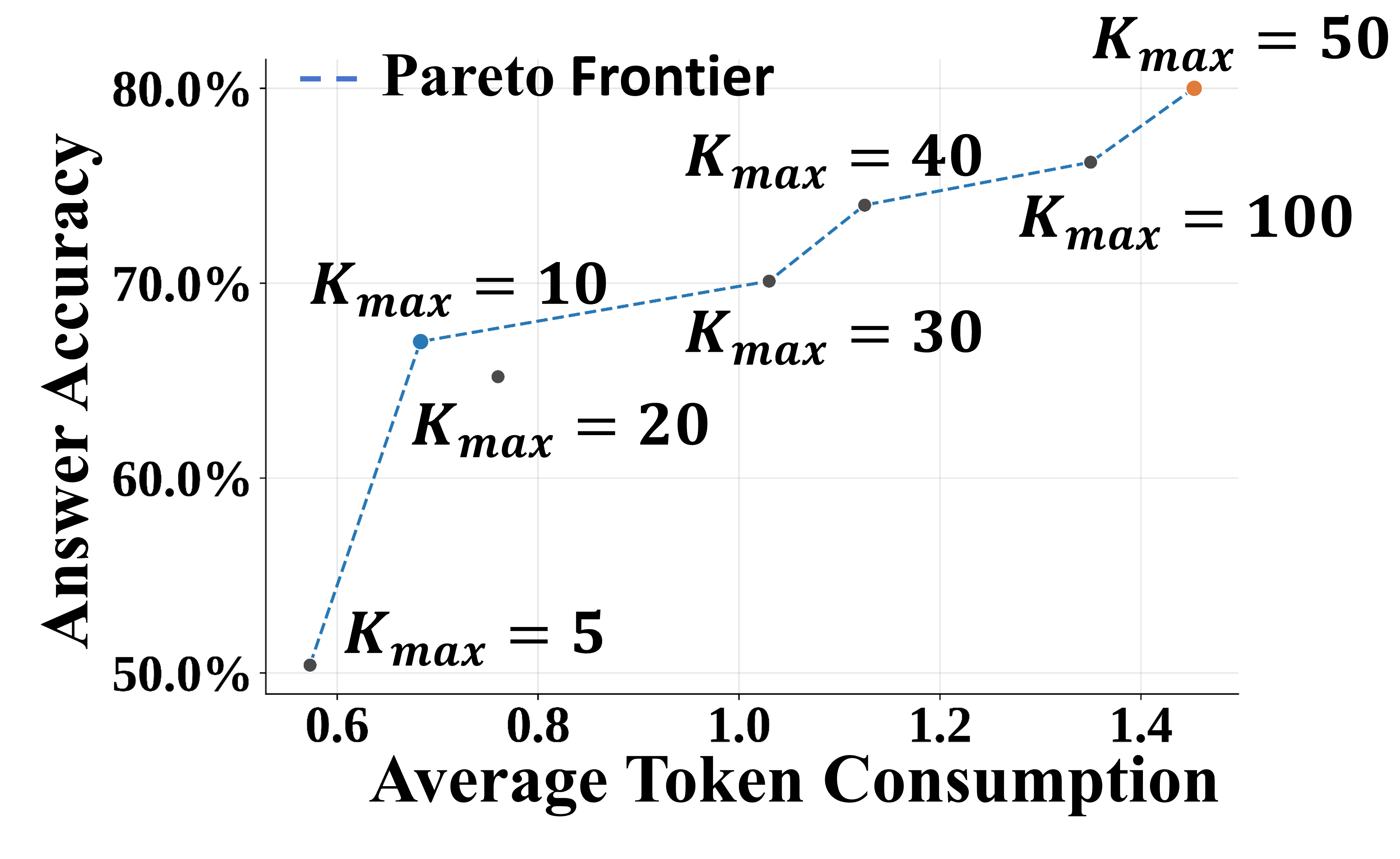}
        \caption{Accuracy Efficiency}
        \label{fig:kmax_pareto}
        \end{subfigure}
        \caption{(a) Privacy rule coverage across rounds and (b) accuracy efficiency trade off across routing budgets.}
        \label{fig:safety_privacy_pareto}
    \end{minipage}
    \hfill 
    \begin{minipage}[b]{0.48\textwidth}
        \centering
        \begin{subfigure}[b]{0.48\linewidth}
            \centering
            \includegraphics[width=\linewidth]{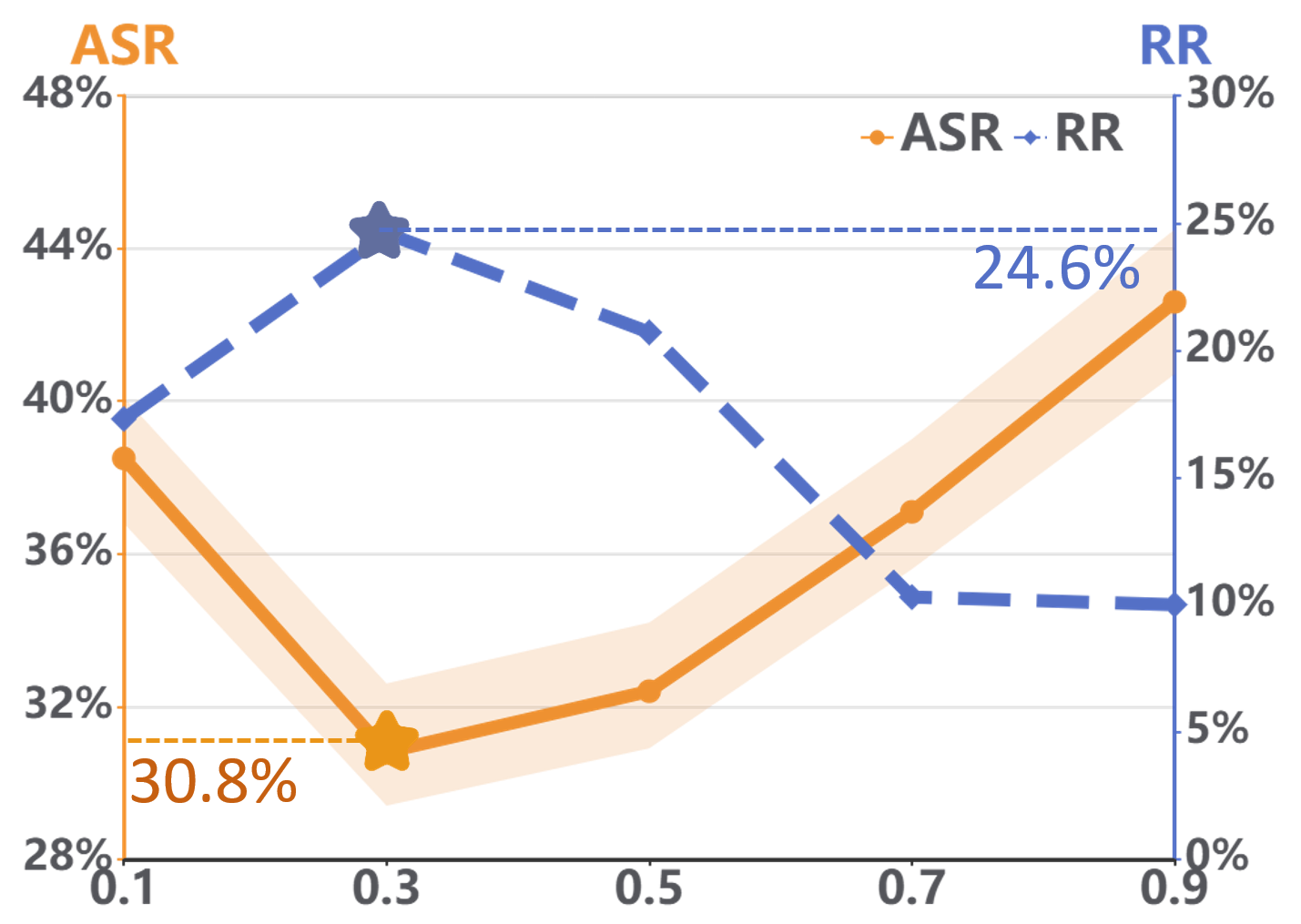}
            \caption{Hyperparameter $\lambda_{pool}$}
            \label{fig:asr_vs_hyperparameter_pool}
        \end{subfigure}
        \hfill
        \begin{subfigure}[b]{0.48\linewidth}
            \centering
            \includegraphics[width=\linewidth]{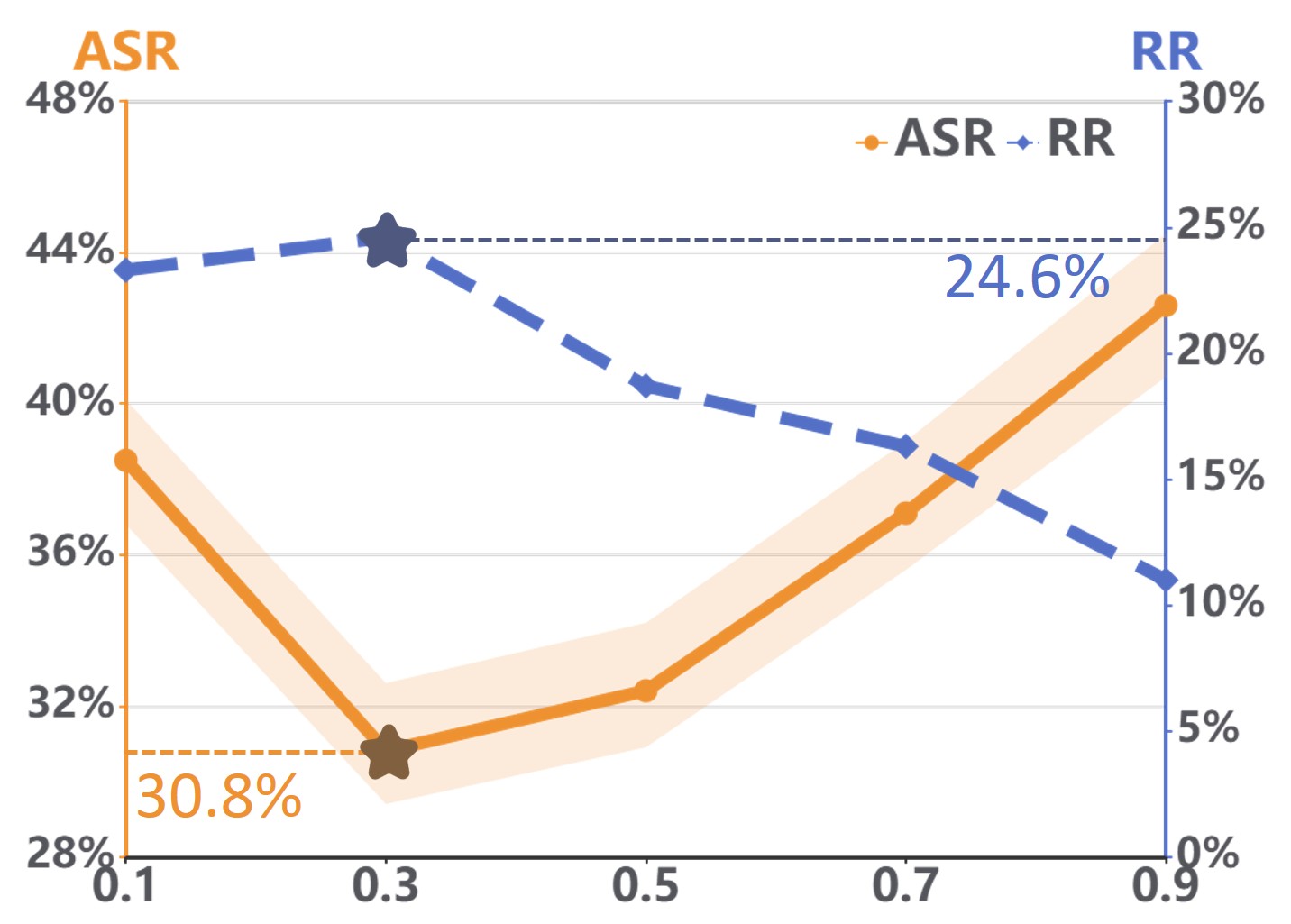}
            \caption{Consensus $\alpha_{con}$}
            \label{fig:asr_vs_hyperparameter_line_chart}
        \end{subfigure}
        \caption{Hyperparameters $\lambda_{pool}$ and $\alpha_{con}$. The x-axis represents the parameter values, and the y-axis shows the corresponding ASR and RR.}
        \label{fig:Hyperparameter}
    \end{minipage}
    
\end{figure}

\textbf{Cross Client Privacy Rule Coordination and Personalized Adaptation (RQ2).}
Table~\ref{tab:framework_results} shows that PFedMA achieves the lowest leakage rate of $26.70$, which is $5.90$ points lower than the second best result.  It also obtains the best LPR, LPA, LPP, and FRA scores  while achieving the second best EA, indicating that the improved privacy protection does not come at the cost of substantially degraded task utility. The lower leakage rate, together with the higher FRA, indicates that PFedMA strengthens measured privacy rule enforcement while preserving task utility under the evaluated setting. Figure~\ref{fig:privacy_coverage} further shows that privacy rule coverage increases progressively across collaboration rounds. 
This trend is consistent with the intended role of federated privacy rule co-evolution, through which reusable rules identified locally are consolidated globally and subsequently adapted to client specific requirements. Together, these results provide empirical support for cross client privacy rule coordination and personalized adaptation while maintaining a favorable privacy utility balance.

\textbf{Reliability Evaluation and Adversarial Robustness (RQ3).}
Table~\ref{tab:security_result} compares PFedMA with representative baselines under prompt injection attacks. PFedMA achieves the lowest ASR of $30.80$, reducing the attack success rate by $1.21$ points compared with the second best method, EPEAgents. Meanwhile, PFedMA obtains the highest RR, BP, and NRP, and the second highest PNA, resulting in the best overall score of $63.45$. These results indicate that the robustness improvement is not achieved by simply rejecting more responses. PFedMA remains effective on benign inputs while reducing the influence of adversarial responses. Overall, the results demonstrate a favorable robustness utility trade off under adversarial federated collaboration.

\subsection{Ablation Study}

\label{sec:ablation}

\textbf{Effectiveness of Feedback Driven Iterative Orchestration.}
Figure~\ref{fig:benchmark_results} evaluates the contributions of query optimization and the historical routing states. Specifically, the -RQ variant removes feedback guided query optimization, the -IA variant removes the historical capability state $C_i$, and the -RT variant removes the historical reliability state $S_i$. Each ablation breaks a distinct pathway through which outcomes from the current round influence subsequent collaboration decisions. The consistent degradation of these variants demonstrates that PFedMA benefits from maintaining a closed cross round feedback loop, rather than merely repeating independent multi-agent interactions.

\textbf{Necessity of Federated Privacy Rule Co-Evolution.}
Figure~\ref{fig:ablation_private} compares PFedMA with five variants, where NR removes privacy rules, SR keeps rules static, LE updates rules only locally, G2L retains only global to local rule adaptation, and L2G retains only local to global rule aggregation. PFedMA achieves the privacy utility trade off, confirming the importance of combining personalized local adaptation with cross client rule synchronization. Figure~\ref{fig:privacy_evolution_curve} further shows how privacy rule effectiveness evolves across federated communication rounds.

PFedMA consistently achieves the highest FRA, outperforming local only evolution, static rules, and the variant without privacy rules. Its persistent advantage across rounds indicates that local adaptation alone is insufficient to fully exploit reusable privacy knowledge. Together, these results demonstrate that bidirectional local to global rule consolidation and global to local personalized adaptation are both important for improving privacy rule effectiveness while preserving task utility.

\textbf{Effectiveness of Multi Factor Reliability Evaluation.}
Figure~\ref{fig:ablation_safe} compares PFedMA with variants that remove the complete reliability evaluator or individual evaluation dimensions, including intent, evidence, and consensus. PFedMA achieves the best robustness utility profile, confirming the importance of reliability aware response assessment before aggregation. Removing the complete evaluator causes the largest overall degradation, while removing any individual dimension also weakens performance. These results indicate that relying on a single reliability signal is insufficient to consistently identify unreliable responses.

%% file: context/conclusion.tex
\section{Conclusion and Limitations}
\label{section:Limitations}

We propose PFedMA, a federated multi-agent framework integrating feedback-driven orchestration with federated privacy rule co-evolution. The orchestration loop adapts agent selection based on evolving task requirements and historical reliability, while the privacy loop enables bidirectional rule evolution while preserving personalized constraints. Experiments across reasoning, privacy-sensitive, and adversarial tasks demonstrate improvements in task performance, privacy rule effectiveness, and robustness. PFedMA focuses on semantic privacy rule coordination under federated data isolation without providing formal privacy guarantees.

%% file: context/change/appendix_change.tex
\appendix

\section{Algorithm}
\label{alg:pfedma}

In this section, we mainly introduce the core algorithm in our paper.
\begin{algorithm}[htbp]
\caption{Execution Workflow of PFedMA}

\begin{algorithmic}[1]
\STATE \textbf{Input:}
initial task $q^1$,
clients set $\mathcal{A}=\{A_1,\dots,A_N\}$,
domain set $\mathcal Domain=\{d_1,\dots,d_L\}$,
domain prototypes $\{\rho_d\}_{d\in\mathcal Domain}$,
maximum number of rounds $T_{\max}$,
maximum number of clients per round $K_{\max}$,
risk threshold $\tau_{\mathrm{risk}}$,
consensus weight $\alpha_{\mathrm{con}}$,
global privacy rule repository $\Omega_{\mathrm{global}}^1$,
and personalized privacy rule repositories
$\{\Omega_{i,\mathrm{local}}^1\}_{i=1}^N$
\STATE Initialize
$t\gets1$
$C^1_i\gets\mathbf{0}$,
$S^1_i\gets\mathbf{0}$,
$z^0\gets\emptyset$,
$n_i^1\gets0$
for all $A_i\in\mathcal A$
\STATE Each agent $A_i$ maintains local experience
$\mathcal X_i$ and routing state
$\mathbf M_i^1=
[\mathbf C_i^1\Vert\mathbf S_i^1]$
\WHILE{$t\leq T_{\max}$}
    \STATE
    Apply the global privacy rule repository to the current query:
    $
    \tilde q^t
    \gets
    P_{\mathrm{in}}
    (q^t,\Omega_{\mathrm{global}}^t)
    $
    \STATE
    Identify the most relevant domain:
    $
    d_\star^t
    \gets
    \arg\max_{d\in\mathcal Domain}
    \mathrm{Cos}(\tilde q^t,\rho_d)
    $
    \STATE
    Retrieve the corresponding candidate agent set
    $\mathcal A_{d_\star^t}$
    \STATE
    Select participating agents:
    $
    \mathcal E^t
    \gets
    \pi_\theta
    \left(
    \mathcal A_{d_\star^t}
    \mid
    \tilde q^t,
    z^{t-1},
    \{\mathbf M_i^t\}_{A_i\in\mathcal A_{d_\star^t}}\right),
    1\leq|\mathcal E^t|\leq K_{\max}
    $
    \IF{$\pi_\theta$ selects the termination action}
        \STATE \textbf{break}
    \ENDIF
    \FORALL{$A_i\in\mathcal E^t$}
        \STATE
        Generate the local response:
        $
        r_i^t
        \gets
        A_i(\tilde q^t,D_i,\mathcal X_i)
        $
        \STATE
        Apply personalized privacy processing:
        $
        \tilde r_i^t
        \gets
        P_{\mathrm{resp},i}
        (r_i^t,\Omega_{i,\mathrm{local}}^t)
        $
    \ENDFOR
    \FORALL{$A_i\in\mathcal E^t$}
        \STATE
        $\mathrm{Int}_i^t
        \gets
        \mathrm{IntentAnomaly}(\tilde r_i^t)$
        \STATE
        $\mathrm{Evd}_i^t
        \gets
        \mathrm{EvidenceAnomaly}(\tilde r_i^t)$
        \STATE
        $\mathrm{Con}_i^t
        \gets
        \mathrm{ConsensusAnomaly}(\tilde r_i^t)$
        \STATE
        $\mathrm{Risk}_i^t
        \gets
        \mathrm{Int}_i^t
        +
        \mathrm{Evd}_i^t
        +
        \alpha_{\mathrm{con}}
        \mathrm{Con}_i^t$
    \ENDFOR
    \STATE
    Retain responses satisfying the reliability threshold:
    $
    \hat{\mathcal E}^t
    \gets
    \left\{
    A_i\in\mathcal E^t
    \mid
    \mathrm{Risk}_i^t
    \leq
    \tau_{\mathrm{risk}}
    \right\}
    $
    \IF{$\hat{\mathcal E}^t\neq\emptyset$}
        \STATE
        Update the global reasoning state:
        $
        z^t
        \gets
        \mathrm{Fusion}
        \left(
        \{\tilde r_i^t\}_{A_i\in\hat{\mathcal E}^t},
        z^{t-1}
        \right)
        $
    \ELSE
        \STATE $z^t\gets z^{t-1}$
    \ENDIF
    \STATE
    Update the routing states
    $\{\mathbf M_i^{t+1}=\left[\mathbf{C}_i^{t+1}\Vert S_i^{t+1}\right]\}_{A_i\in\mathcal E^t}$ using the observed capability and reliability feedback
    \STATE
    Construct generalized local rule candidates from newly identified personalized privacy constraints
    \STATE
    Build the candidate pool $\mathcal B^t$ and obtain the supported global candidate set $\mathcal G^t$
    \STATE
    Collect rule-level feedback $F^t$ and update the global repository:
    $
    \Omega_{\mathrm{global}}^{t+1}
    \gets
    \mathrm{RuleUpdate}
    \left(
    \Omega_{\mathrm{global}}^t,
    \mathcal G^t,
    F^t
    \right)
    $
    \STATE
    Redistribute updated global rules and selectively update each personalized repository
    $\Omega_{i,\mathrm{local}}^{t+1}$
    \STATE
    Refine the query for the next round:
    $
    q^{t+1}
    \gets
    \mathrm{Refine}(q^t,z^t)
    $
    \STATE $t\gets t+1$
\ENDWHILE

\STATE
\textbf{Output:}
Final answer derived from the latest global reasoning state

\end{algorithmic}
\end{algorithm}

\section{Comparison with Related Frameworks}
\label{sec:method_comparison}

\begin{table}[htbp]
\centering
\caption{Comparison of PFedMA with representative multi-agent and federated multi-agent frameworks. PFedMA differs from prior methods by jointly coupling iterative reasoning, agent reliability, and heterogeneous privacy constraints within a closed loop orchestration process.}
\label{tab:method_comparison}
\setlength{\tabcolsep}{2.5pt}
\renewcommand{\arraystretch}{1.05}
\small
\resizebox{1\linewidth}{!}{
\begin{tabular}{lcccccc}
\toprule
\textbf{Method} 
& \textbf{Setting} 
& \textbf{Expert Coordination} 
& \textbf{Iterative Refinement} 
& \textbf{Reliability Feedback} 
& \textbf{Privacy Personalization} 
& \textbf{Privacy Rules Co-evolution} \\
\midrule

Router-R1 
& Centralized MAS
& RL-based dynamic routing
& \checkmark
& --
& --
& -- \\

KABB 
& Centralized MAS
& Knowledge aware expert selection
& Partial
& --
& --
& -- \\

Fed-ICL 
& Federated MAS
& Federated response aggregation
& \checkmark
& --
& --
& -- \\

\textbf{PFedMA} 
& \textbf{Federated MAS}
& \textbf{Reliability aware dynamic routing}
& \checkmark
& \checkmark
& \checkmark
& \checkmark \\

\bottomrule
\end{tabular}
}
\end{table}
As summarized in Table~\ref{tab:method_comparison}, existing methods primarily optimize
expert selection, sequential routing, or federated response refinement in isolation. In contrast,
PFedMA establishes a closed loop orchestration process in which reasoning progress, response
reliability, and heterogeneous privacy constraints jointly influence subsequent collaboration decisions.

\section{Structured Response Evaluation Details}
\label{section:anomaly_details}

\subsection{Structured Response Format and Example}
\label{section:Structured_Response_Format}
\label{section:Response Format}

The agent's response follows a structured format consisting of three components: final\_answer, claim, and evidence.
At round $t$, each selected client $A_i$ returns a privacy rule based information minimization response
using the following structured format:
\begin{minipage}{\linewidth}
\begin{verbatim}
{
  "final_answer": "...",
  "claims": [
    {
      "claim": "...",
      "evidence": ["...", "..."]
    },{
      "claim": "...",
      "evidence": ["...", "..."]
    }
  ],
  "domain_fit": 0.0,
  "self_confidence": 0.0
}
\end{verbatim}
\end{minipage}

The corresponding mathematical representation is
\begin{equation}
\tilde r_i^t
=
\left\langle
a_i^t,\Gamma_i^t,\mathcal{L}_i^t
\right\rangle,
\end{equation}
The structured response separates the final conclusion from the claim level information used for subsequent reliability evaluation. Here, $a_i^t$ denotes the final conclusion returned by agent $A_i$ at round $t$. 
\begin{equation}
\Gamma_i^t={\gamma_{i,n}^t}_{n=1}^{|\Gamma_i^t|},
\end{equation}
where $\gamma_{i,n}^t$ denotes the $n$-th explicit claim in the response, and $|\Gamma_i^t|$ denotes the number of claims.

Each claim is paired with a corresponding evidence set, such that
\begin{equation}
\mathcal{L}_i^t={L_{i,n}^t}_{n=1}^{|\Gamma_i^t|},
\end{equation}
where $L_{i,n}^t$ contains the evidence provided specifically for
claim $\gamma_{i,n}^t$. In particular,
\begin{equation}
L_{i,n}^t
=
\left\{
\epsilon_{i,n,k}^t
\right\}_{k=1}^{m_{i,n}^t},
\end{equation}
where $\epsilon_{i,n,k}^t$ denotes the $k$-th evidence item
associated with the $n$-th claim, and $m_{i,n}^t$ denotes the number of evidence items provided for that claim. Thus, the structured representation establishes an explicit claim evidence correspondence: each $\gamma_{i,n}^t$ is evaluated together with its own evidence set $L_{i,n}^t$, while different claims may be supported by different numbers of evidence items.

The fields \texttt{domain\_fit} and \texttt{self\_confidence} are
self reported auxiliary metadata in $[0,1]$. Because these values may
be manipulated by unreliable clients, they are not directly treated
as reliability scores.

\begin{figure}[H]
    
    \centering
    \includegraphics[width=\linewidth]{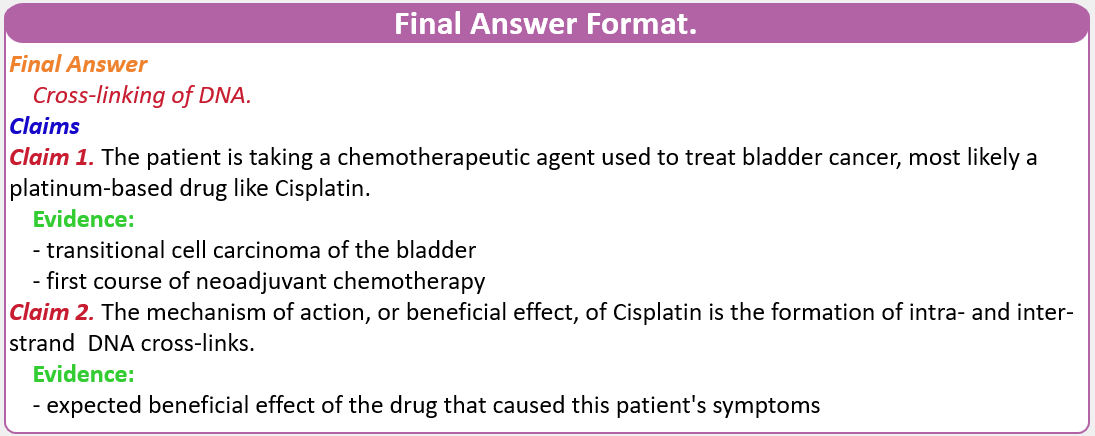}
    \caption{Format of the client's response.}
    \label{format2}
\end{figure}

\paragraph{Example.}
\textbf{[AGE]} year old \textbf{[GENDER]} with transitional cell carcinoma of the bladder comes to the physician because of a \textbf{[TIME]} day history of ringing sensation in his ear. He received this first course of neoadjuvant chemotherapy \textbf{[TIME]}. Pure tone audiometry shows a sensorineural hearing loss of \textbf{[NUMBER]} dB. The expected beneficial effect of the drug that caused this patient's symptoms is most likely due to which of the following actions? An example of a client response is shown in the Figure \ref{format2}.

\subsection{Evaluator Implementation and Parameter Isolation}
\label{section:evaluator_implementation}

The response evaluators are separated from the trainable orchestrator $\pi_\theta$(Qwen2.5-3B-Instruct). Specifically, the intent and evidence anomaly scores are produced by a frozen judge model using fixed evaluation prompts and predefined scoring criteria. The judge does not share trainable parameters with $\pi_\theta$, and its parameters remain unchanged during both reinforcement learning training and inference. The intent and evidence evaluators use the same frozen judge model (Qwen2.5-3B-Instruct) backbone but different task specific instructions.

The consensus anomaly does not invoke the judge model. It is computed deterministically from the semantic similarity between the conclusions of the participating agents. Therefore, the three anomaly dimensions do not introduce additional trainable components into PFedMA.

The judge receives only the response fields required by the corresponding evaluator. The intent evaluator receives the response sentences and the predefined intent risk categories, whereas the evidence evaluator receives each claim and its associated evidence. It does not observe the ground truth answer, the routing reward, or the orchestrator's subsequent decision. The resulting anomaly scores are used by the server for response filtering and agent state updates, but no evaluator gradient is propagated to the orchestrator.
\begin{table}[t]
\centering
\caption{Implementation and training status of the response evaluators.}
\label{tab:evaluator_implementation}
\begin{tabular}{lccc}
\toprule
Component & Implementation & Shares $\pi_\theta$ & Trainable \\
\midrule
Intent evaluator   & Frozen LLM judge       & No & No \\
Evidence evaluator & Frozen LLM judge       & No & No \\
Consensus evaluator& Semantic similarity    & No & No \\
Orchestrator        & RL fine tuned LLM      & -- & Yes \\
\bottomrule
\end{tabular}
\end{table}

\subsection{Anomaly Definitions and Boundary Cases}
\label{section:anomaly_tails}
\paragraph{Pooling Operator}

For any non empty set of scores $\{x_j\}_{j=1}^{J}$, we define
\begin{equation}
\operatorname{Pool}_{\lambda_{pool}}
\left(
\{x_j\}_{j=1}^{J}
\right)
=
\lambda_{pool}
\max_{1\leq j\leq J}x_j
+
(1-\lambda_{pool})
\frac{1}{J}
\sum_{j=1}^{J}x_j,
\end{equation}
where $J$ denotes the number of fine-grained anomaly scores being aggregated, $x_j$ denotes the $j$-th anomaly score, and $\lambda_{\mathrm{pool}}\in[0,1]$ controls the trade-off between the most severe anomaly and the average anomaly level.

\paragraph{Intent Anomaly.}

Let
$
\Xi_i^t
=
\left\{
\xi_{i,j}^t
\right\}_{j=1}^{J_i^t}
$
denote the sentence level units extracted from
$\tilde r_i^t$. The predefined intent risk set is
\begin{equation}
\nu_{\mathrm{int}}
=
\{
\text{instruction overriding},
\text{false authority},
\text{aggregator manipulation},
\text{goal substitution}
\}.
\end{equation}

For each response unit, the evaluator produces an intent risk score
\begin{equation}
u_{i,j}^t
=
\operatorname{Eval}_{\mathrm{int}}
\left(
\xi_{i,j}^t,
\nu_{\mathrm{int}}
\right)
\in[0,1].
\end{equation}

The response level intent anomaly is
\begin{equation}
Int_i^t
=
\operatorname{Pool}_{\lambda_{\mathrm{pool}}}
\left(
\{u_{i,j}^t\}_{j=1}^{J_i^t}
\right).
\end{equation}

\paragraph{Evidence Anomaly.}

Privacy rule based information minimization adjustment or abstracts sensitive details while aiming to preserve the evidence supporting each claim. The evaluator assesses the support provided by the privacy processed evidence; omitted sensitive details are not themselves treated as anomalies when the remaining information sufficiently supports the claim. Verifiability refers to the judge's assessment based on the supplied text, rather than independent verification of external sources. For claim $\gamma_{i,n}^t$, let
\begin{equation}
\mathcal{L}_{i,n}^t
=
\{\varepsilon_{i,n,k}^t\}_{k=1}^{m_{i,n}^t},
\end{equation}
where $\varepsilon_{i,n,k}^t$ denotes the $k$-th evidence item associated with the $n$-th claim, and $m_{i,n}^t$ denotes the number of evidence items provided for that claim. For $m_{i,n}^t>0$, let
$\operatorname{Sup}_{i,n,k}^t\in[0,1]$ and
$\operatorname{Ver}_{i,n,k}^t\in[0,1]$ denote the support strength and verifiability of evidence item $\varepsilon_{i,n,k}^t$,
respectively. We define
\begin{align}
\overline{\operatorname{Sup}}_{i,n}^t
&=
\frac{1}{m_{i,n}^t}
\sum_{k=1}^{m_{i,n}^t}
\operatorname{Sup}_{i,n,k}^t,
\\
\operatorname{Ver}_{i,n}^t
&=
\max_{1\leq k\leq m_{i,n}^t}
\operatorname{Ver}_{i,n,k}^t.
\end{align}

The internal coherence of the evidence set is
\begin{equation}
\operatorname{Coh}_{i,n}^{t}
=
\begin{cases}
\displaystyle
\frac{2}{m_{i,n}^{t}\left(m_{i,n}^{t}-1\right)}
\sum_{1 \leq k_1 < k_2 \leq m_{i,n}^{t}}
\operatorname{Cons}
\left(
\varepsilon_{i,n,k_1}^{t},
\varepsilon_{i,n,k_2}^{t}
\right),
& m_{i,n}^{t} > 1, \\[6pt]
1,
& m_{i,n}^{t} = 1, \\[2pt]
0,
& m_{i,n}^{t} = 0.
\end{cases}
\end{equation}

The claim level evidence quality is
\begin{equation}
\operatorname{Qual}^{\mathrm{evd}}_{i,n}{}^t
=
\begin{cases}
\displaystyle
\frac{
\overline{\operatorname{Sup}}_{i,n}^t
+
\operatorname{Ver}_{i,n}^t
+
\operatorname{Coh}_{i,n}^t
}{3},
& m_{i,n}^t>0,\\[9pt]
0,
& m_{i,n}^t=0.
\end{cases}
\end{equation}

Accordingly, the claim level evidence anomaly is
\begin{equation}
\operatorname{Anom}^{\mathrm{evd}}_{i,n}{}^t=1-\operatorname{Qual}^{\mathrm{evd}}_{i,n}{}^t.
\end{equation}

The response level evidence anomaly is
\begin{equation}
\label{eq:response_evidence_anomaly}
\mathrm{Evd}_i^t
=
\begin{cases}
\displaystyle
\mathrm{Pool}_{\lambda_{\mathrm{pool}}}
\left(
\left\{
\mathrm{Anom}_{i,n}^{\mathrm{evd},t}
\right\}_{n=1}^{|\Gamma_i^t|}
\right),
& |\Gamma_i^t| > 0,\\[8pt]
1,
& |\Gamma_i^t| = 0.
\end{cases}
\end{equation}
When no explicit claim is provided, i.e., $|\Gamma_i^t|=0$, the evidence anomaly is set to its maximum value of $1$. This boundary condition prevents an empty claim set from bypassing evidence-quality evaluation and is consistent with the structured response format, in which claims and their associated evidence provide the basis for evidence reliability assessment.

\paragraph{Consensus Anomaly.}

The consensus anomaly measures whether the conclusion of client $A_i$
is supported by the conclusions of the other selected clients:
\begin{equation}
Con_i^t
=
\begin{cases}
\displaystyle
1-
\frac{1}{|\mathcal{E}^t|-1}
\sum_{\substack{A_{i'}\in\mathcal{E}^t\\ i'\neq i}}
\operatorname{Sim}
\left(
a_i^t,
a_{i'}^t
\right),
& |\mathcal{E}^t|>1,\\[10pt]
0,
& |\mathcal{E}^t|=1.
\end{cases}
\end{equation}
where $\operatorname{Sim}(\cdot,\cdot)\in[0,1]$ denotes normalized semantic similarity. A larger value therefore indicates stronger deviation from peer conclusions. The singleton case is assigned zero because consensus cannot be assessed from only one response.

\subsection{Overall Anomaly Score and Robust Aggregation}

The three anomaly dimensions are combined into an overall response
anomaly score:
\begin{equation}
\label{eq:appendix_overall_anomaly}
\mathcal Risk_i^t
=
Int_i^t
+
Evd_i^t
+
\alpha_{\mathrm{con}}Con_i^t.
\end{equation}

Responses whose anomaly scores exceed the tolerance threshold are
excluded from aggregation:
\begin{equation}
\label{eq:appendix_response_filtering}
\widehat{\mathcal{E}}^t
=
\left\{
A_i\in\mathcal{E}^t
\mid
\mathcal Risk_i^t\leq\tau_{\mathrm{risk}}
\right\}.
\end{equation}

The orchestrator then aggregates the retained privacy rule based information minimization responses to update the accumulated collaboration state:
\begin{equation}
z^t =
\begin{cases}
\operatorname{Fusion}\!\left(
\{\tilde r_i^t\}_{A_i\in\widehat{\mathcal E}^{t}},
z^{t-1}
\right),
& \widehat{\mathcal E}^{t}\neq\varnothing, \\[4pt]
z^{t-1},
& \widehat{\mathcal E}^{t}=\varnothing.
\end{cases}
\end{equation}

\section{Agent Routing State Update}
\label{section:agent_routing_state_update}

After each orchestration round, PFedMA updates the routing metadata of the participating agents based on their observed historical capability and historical response reliability.

For agent $A_i$ with at least one previously retained response, define:
\begin{equation}
\label{eq:last_participation}
\tau_i(t)
=
\max\{s<t \mid A_i \in \widehat{\mathcal E}^s\}
\end{equation}
Thus, $\tau_i(t)$ indexes the most recent round before $t$ in which the response of client $A_i$ was retained by the reliability filter. Denote the most recent round before $t$ in which the response of agent $A_i$ was retained by the reliability filter. Here, $s$ indexes previous collaboration rounds. This definition allows the historical capability update to compare the current response with the most recent reliable response of the same agent while naturally accommodating intermittent agent participation across rounds.

If the current response passes the reliability filter and the agent has a previously retained response, the orchestrator compares the current response with that agent’s most recent retained response from an earlier round to obtain a response level improvement signal and an agent specific capability increment vector.
\begin{equation}
\label{eq:capability_feedback}
f_i^t
=
\pi_{\theta}^{\mathrm{fb}}\!\left(
\tilde{r}_i^t
\mid
\tilde{r}_i^{\tau_i(t)}
\right),
\qquad
\bar{\mathbf v}_i^t
=
\pi_{\theta}^{\mathrm{cap}}\!\left(
\tilde{r}_i^t
\mid
\tilde{r}_i^{\tau_i(t)}
\right).
\end{equation}

Here, $\pi_{\theta}^{\mathrm{fb}}$ and $\pi_{\theta}^{\mathrm{cap}}$ denote two comparative evaluations performed by the same orchestrator $\pi_\theta$ under different fixed evaluation instructions, rather than conditional likelihoods. The response level feedback signal $f_i^t \in \{0,1\}$ indicates whether the current retained response improves upon the most recent retained response of the same client. The capability vector $\bar{\mathbf v}_i^t$ identifies the capability dimensions in which the current response improves upon that reference. Before collaboration begins, the historical capability state of each agent is initialized as a zero vector: $\mathbf C_i^1 = [0,\ldots,0]^\top. $

For the first retained response of an agent, no previously retained response is available for comparison. This response therefore establishes the initial comparison reference, and no capability  progress increment is assigned in that round.

For subsequent responses, the historical capability state is updated as
\begin{equation}
\label{eq:capability_update}
\mathbf C_i^{t+1}
=
\begin{cases}
\mathbf C_i^t
+
f_i^t \bar{\mathbf v}_i^t,
& A_i \in \widehat{\mathcal E}^t,\\
\mathbf C_i^t,
& A_i \notin \widehat{\mathcal E}^t,
\end{cases}
\end{equation}
where $\widehat{\mathcal E}^t$ denotes the set of responses retained by the reliability filter. Therefore, responses excluded from aggregation do not receive positive progress credit.

The historical reliability state records the cumulative mean of the anomaly scores observed from the responses of agent $A_i$. Let $n_i^t$ denote the number of responses from agent $A_i$ that have been evaluated prior to round $t$. Before collaboration begins, the observation count and historical reliability state are initialized as
\begin{equation}
\label{eq:safety_initialization}
n_i^1 = 0,
\qquad
\mathbf S_i^1
=
\begin{bmatrix}
0 & 0 & 0
\end{bmatrix}^{\top}.
\end{equation}

Whenever a new response from agent $A_i$ is evaluated, the observation count is updated as
\begin{equation}
\label{eq:safety_observation_count}
n_i^{t+1}
=
n_i^t + 1,
\end{equation}
and the historical reliability state is updated as
\begin{equation}
\label{eq:safety_state_update}
\mathbf S_i^{t+1}
=
\frac{
n_i^t \mathbf S_i^t
+
\left[
\mathrm{Int}_i^t,
\mathrm{Evd}_i^t,
\mathrm{Con}_i^t
\right]^{\top}
}{
n_i^t + 1
}.
\end{equation}

Here, $\mathrm{Int}_i^t$, $\mathrm{Evd}_i^t$, and $\mathrm{Con}_i^t$ denote the intent, evidence, and consensus anomaly scores of the current response, respectively. Unlike the historical capability update, every evaluated response contributes to the historical reliability state, including responses excluded from aggregation by the reliability filter. This allows unreliable responses to remain reflected in the agent's historical risk profile.

If agent $A_i$ does not return a newly evaluated response at round $t$, its historical capability state, reliability states, and observation count remain unchanged. Finally, the updated routing state is defined as
\begin{equation}
\label{eq:metadata_update}
M_i^{t+1}
=
\left[
\mathbf C_i^{t+1}
\,\Vert\,
\mathbf S_i^{t+1}
\right].
\end{equation}

Thus, the routing state jointly captures accumulated evidence of agent-specific historical capability and historical response reliability, allowing subsequent routing decisions to account for both observed collaboration progress and historical reliability.

\section{Reinforcement Learning Details}
\label{section:reward_details}

\subsection{Reward Definitions}
\label{section:reward_definitions}

Let $\tau$ denote a collaboration trajectory containing $T$ executed routing rounds, where $T_{\max}$ is the maximum routing round budget, $T\leq T_{\max}$, and $t\in\{1,\ldots,T\}$ indexes the executed rounds. As defined in Equation~\ref{eq:rl_reward}, the trajectory reward consists of a format reward, a terminal answer reward, and step-wise rewards that provide intermediate supervision for multi-round orchestration.

\paragraph{Format and final conclusion rewards.}
\label{sec:Format and final answer rewards}
The format reward evaluates whether the orchestrator generated outputs satisfy the required structured format:
\begin{equation}
R_{\mathrm{fmt}}(\tau)
=
\begin{cases}
0,  & \text{if the required output format is satisfied},\\[4pt]
-1, & \text{otherwise}.
\end{cases}
\end{equation}
The terminal answer reward evaluates the correctness of the final
answer using the corresponding benchmark evaluator:
\begin{equation}
R_{\mathrm{ans}}(\tau)
=
\begin{cases}
1,  & \text{if the terminal answer is judged correct},\\[4pt]
-1, & \text{otherwise}.
\end{cases}
\end{equation}

\paragraph{Cross-round reasoning progress.}
\label{sec:Cross-round reasoning progress}
For a training instance with ground truth answer $y^\star$, let
$\hat y^t$ denote the answer extracted from the aggregated reasoning
state $z^t$ using the corresponding benchmark's answer extraction
procedure. Exact Match (EM) is defined as
\begin{equation}
\mathrm{EM}_t
=
\mathbb{I}\!\left[\hat y^t=y^\star\right],
\end{equation}
where $\mathbb{I}[\cdot]$ is the indicator function. To measure partial
correctness, let $\mathcal{T}_{\mathrm{pred}}^t$ and
$\mathcal{T}_{\mathrm{gt}}$ denote the token multisets obtained after
applying the same benchmark specific normalization procedure to
$\hat y^t$ and $y^\star$, respectively. Token level precision is
\begin{equation}
\mathrm{Prec}_t
=
\begin{cases}
\displaystyle
\frac{
\left|\mathcal{T}_{\mathrm{pred}}^t
\cap\mathcal{T}_{\mathrm{gt}}\right|
}{
\left|\mathcal{T}_{\mathrm{pred}}^t\right|
},
& \left|\mathcal{T}_{\mathrm{pred}}^t\right|>0,\\[10pt]
0, & \text{otherwise}.
\end{cases}
\end{equation}
Token level recall is
\begin{equation}
\mathrm{Rec}_t
=
\begin{cases}
\displaystyle
\frac{
\left|\mathcal{T}_{\mathrm{pred}}^t
\cap\mathcal{T}_{\mathrm{gt}}\right|
}{
\left|\mathcal{T}_{\mathrm{gt}}\right|
},
& \left|\mathcal{T}_{\mathrm{gt}}\right|>0,\\[10pt]
0, & \text{otherwise}.
\end{cases}
\end{equation}
Here, $\cap$ denotes multiset intersection, $|\cdot|$ denotes the
number of tokens counted with multiplicity, precision is the proportion
of predicted tokens that match the ground truth answer, and recall is
the proportion of ground truth tokens covered by the prediction. The
token level F1 score is
\begin{equation}
\mathrm{F1}_t
=
\begin{cases}
\displaystyle
\frac{2\,\mathrm{Prec}_t\mathrm{Rec}_t}
{\mathrm{Prec}_t+\mathrm{Rec}_t},
& \mathrm{Prec}_t+\mathrm{Rec}_t>0,\\[8pt]
0, & \text{otherwise}.
\end{cases}
\end{equation}
Since $\mathrm{Prec}_t$ and $\mathrm{Rec}_t$ are nonnegative, their sum
is zero only when both scores are zero. The second case therefore avoids
division by zero.
Following~\citep{zhang2025router}, the intermediate answer quality score
is
\begin{equation}
\label{eq:appendix_quality_score}
Q_t
=
\mathrm{EM}_t
+
\left(1-\mathrm{EM}_t\right)\mathrm{F1}_t.
\end{equation}
Thus, $Q_t=1$ for an exact match and $Q_t=\mathrm{F1}_t$ otherwise.
We set $Q_0=0$ and define the quality gain at round $t$ as
\begin{equation}
\label{eq:appendix_quality_progress}
\Delta Q_t=Q_t-Q_{t-1}.
\end{equation}

\paragraph{Round level response risk.}
For each response $\tilde r_i^t$, we reuse the overall anomaly score $\mathcal Risk_i^t$. Since its three components are bounded in $[0,1]$, its range is
\begin{equation}
B_{\mathrm{risk}}^t =
\begin{cases}
2, & |\mathcal{E}^t| = 1, \\[3pt]
2 + \alpha_{\mathrm{con}}, & |\mathcal{E}^t| > 1.
\end{cases}
\end{equation}
Each response anomaly score is normalized by $B_{\mathrm{risk}}^t$, and the normalized scores are averaged over the agents invoked at round $t$. The round-level response risk is
\begin{equation}
\label{eq:round_response_risk}
\overline{\mathcal Risk}_t
=
\frac{1}{|\mathcal{E}^t|}
\sum_{A_i \in \mathcal{E}^t}
\frac{\mathrm{Risk}_i^t}{B_{\mathrm{risk}}^t}.
\end{equation}
where $\mathcal E^t$ is the set of agents invoked at round $t$. Here, $\alpha_{\mathrm{con}}\geq0$ is the consensus anomaly weight and $|\mathcal E^t|$ is the number of agents invoked at round $t$. Consequently, $\overline{\mathcal Risk}_t\in[0,1]$ and a larger value indicates a less reliable set of responses.

\paragraph{Agent invocation cost.}
The normalized invocation cost is
\begin{equation}
\label{eq:invocation_cost}
\frac{|\mathcal E^t|}{K_{\max}},
\end{equation}
where $K_{\max}$ is the maximum number of agents that may be invoked in one routing round. 

Combining the three components gives the step wise reward in
Equation~\ref{eq:trajectory_reward}:
\begin{equation}
R_{\mathrm{step}}^t
=
\Delta Q_t-\overline{\mathcal Risk}_t-\frac{|\mathcal E^t|}{K_{\max}}.
\end{equation}
A routing step satisfies $1\leq|\mathcal E^t|\leq K_{\max}$. If the orchestrator selects the termination action, no additional agent is invoked and no further step wise reward is generated.

The ground truth dependent quantities $R_{\mathrm{ans}}$, $Q_t$, and $\Delta Q_t$ are used only during RL training. They are not available to the orchestrator during inference.

\subsection{Policy Optimization}
\label{section:policy_optimization}

We optimize the server side orchestrator $\pi_\theta$, where $\theta$ denotes its trainable parameters, using Proximal Policy Optimization (PPO) implemented with the veRL framework. At round $t$, the policy observes the privacy processed query $\tilde q^t$, the accumulated collaboration state $z^{t-1}$, and the capability and safety metadata $M_i^t$ of the candidate agents. It then generates a structured routing action that either selects a subset $\mathcal E^t$ satisfying $1\leq|\mathcal E^t|\leq K_{\max}$ or terminates the trajectory when no additional agent is required.

After the selected agents return Privacy rule based information minimization responses $\{\tilde r_i^t\}_{A_i\in\mathcal E^t}$, the server evaluates response reliability, aggregates the retained responses, and computes the training reward. PPO updates are performed centrally using the orchestrator visible trajectory. Raw client datasets $\mathcal D_i$, local experience buffers $\mathcal X_i$, complete personalized privacy rules $\Omega_{i,\mathrm{local}}^t$, and unsanitized responses $r_i^t$ remain client local and are not used in policy optimization. This training protocol preserves the federated data isolation boundary, but does not by itself provide a cryptographic or differential privacy guarantee.

\subsection{Training Configuration}
\label{sec:appendix_training_configuration}

We train the Orchestrator $\pi_{\theta}$ using the veRL framework for reinforcement learning in LLMs, with Proximal Policy Optimization (PPO) as the default algorithm. Specifically, we use the training splits of MATH and MedQA for RL optimization of the Orchestrator, while all remaining benchmarks are evaluated in a zero shot manner without further RL fine tuning. During the RL training phase, the batch size is set to 64, with a maximum of 220 training steps. The anomaly evaluation parameters are set to $\lambda_{\mathrm{pool}}=0.3$ and $\alpha_{\mathrm{con}}=0.3$; their sensitivity is analyzed in Section~\ref{sec:appendix_hyperparameter_sensitivity}. Training this model requires 4 NVIDIA A800 GPUs, while inference requires 8 NVIDIA 3090 GPUs. Below are the hyperparameter settings for reinforcement learning in our work. We refer to~\citep{zhang2025router} to set the hyperparameters required for the experiments.

\begin{table}[htbp]
    \centering
    \caption{Hyperparameter Settings.}
    \label{tab:hyperparameter_settings}
    \setlength{\tabcolsep}{2pt}
    \renewcommand{\arraystretch}{1.05}

    \begin{tabular}{@{}lclc@{}}
        \toprule
        \textbf{Hyperparameter} & \textbf{Value} 
        & \textbf{Hyperparameter} & \textbf{Value} \\
        \midrule

        Learning Rate (Actor) & 1e-5
        & Learning Rate (Critic) & 1e-5 \\

        Total Batch Size & 64
        & Mini-batch Size & 64 \\

        Micro-batch Size & 8
        & Max Training Steps & 220 \\

        Max Routing Steps & 100
        & Max Sequence Length & 4096 \\

        Max Response Length & 2048
        & Max Length for LLM API Response & 1024 \\

        Tensor Parallel Size & 1
        & GPU Utilization Ratio & 0.8 \\

        Rollout Sampling Temperature (Train) & 1.0
        & Rollout Sampling Temperature (Eval) & 1.0 \\

        \bottomrule
    \end{tabular}
\end{table}

\section{Federated Privacy Rule Co-Evolution}
\label{section:privacy_details}

\paragraph{Candidate Pool Construction.}
Only clients retained by the response reliability evaluation contribute new rule candidates. To prevent one client from dominating a semantic cluster, each client contributes at most one candidate to that cluster in a collaboration round. After this distinct client contribution check, explicit client identifiers are removed from the stored representations, and the candidates are randomly permuted:
\begin{equation}
\mathcal B^t
=
\operatorname{Shuffle}
\left(
\left\{
\bar{\omega}_i^t
\;\middle|\;
A_i\in\widehat{\mathcal E}^t,
\Delta\Omega_{i,\mathrm{local}}^t\neq\emptyset
\right\}
\right),
\label{eq:privacy_candidate_pool}
\end{equation}
where $\widehat{\mathcal E}^t$ is the retained agent set defined in Equation~\ref{eq:appendix_response_filtering}, and $\mathcal B^t$ is the resulting unordered candidate pool. A client that identifies no new rule at round $t$ contributes no candidate. Removing explicit identifiers and permuting candidates reduces direct identifier exposure in the shared representations but does not provide an anonymity guarantee.

\paragraph{Semantic Support Filtering.}
\label{sec:Semantic-Support Filtering}
For each candidate $\bar{\omega}_i^t\in\mathcal B^t$,
we compute semantic support by counting the other uploaded
candidates whose cosine similarity to it is at least $0.9$:
\begin{equation}
\operatorname{Supp}^t\left(\bar{\omega}_i^t\right)
=
\sum_{\substack{
\bar{\omega}_j^t\in\mathcal B^t\\
j\neq i
}}
\mathbb I
\left[
\operatorname{Cos}
\left(
e\left(\bar{\omega}_i^t\right),
e\left(\bar{\omega}_j^t\right)
\right)
\geq 0.9
\right],
\label{eq:privacy_rule_support}
\end{equation}
where $e(\cdot)$ maps each generalized privacy rule to a semantic embedding, $\operatorname{Cos}(\cdot,\cdot)$ computes cosine similarity, and $\mathbb I[\cdot]$ is the indicator function. The condition $j\neq i$ excludes self support. Under the distinct client contribution check used in candidate pool construction, repeated contributions from the same client to a semantic cluster do not increase its support.

We then normalize the support count to obtain the proportion of other candidates that are semantically similar to $\bar{\omega}_i^t$:
\begin{equation}
\zeta_i^t
=
\frac{
\operatorname{Supp}^t\left(\bar{\omega}_i^t\right)
}{
\max\left\{1,|\mathcal B^t|-1\right\}
},
\label{eq:privacy_rule_support_proportion}
\end{equation}
where $\zeta_i^t\in[0,1]$. The denominator prevents division by zero for a singleton pool.

For a pool containing multiple candidates, we select those with the highest positive support proportion, without introducing an additional support count or support proportion threshold. For a singleton pool, the sole candidate is passed directly to the global update.
The resulting global candidate set is
\begin{equation}
\mathcal G^t
=
\begin{cases}
\varnothing,
& |\mathcal B^t|=0,\\[4pt]
\mathcal B^t,
& |\mathcal B^t|=1,\\[4pt]
\left\{\bar{\omega}_i^t\in\mathcal B^t
\;\middle|\;
\zeta_i^t>0,\;
\zeta_i^t=
\displaystyle
\max_{\bar{\omega}_j^t\in\mathcal B^t}\zeta_j^t
\right\},
& |\mathcal B^t|>1.
\end{cases}
\label{eq:supported_privacy_candidates}
\end{equation}
All candidates tied for the highest positive support proportion are retained. If multiple candidates are present but none receives support, then $\mathcal G^t=\varnothing$. The singleton case explicitly permits a candidate to enter the global update without cross client corroboration. For larger pools, the selection favors candidates with the broadest support and excludes those with lower support proportions.
Semantically equivalent accepted candidates are subsequently
consolidated into a representative global rule, and conflicting
alternatives are handled through feedback based conflict resolution.

\paragraph{Local Utility Evaluation and Rule Selection.}
\label{sec:Local Utility Evaluation and Rule Selection.}
After the global update, reusable rules are redistributed to the clients and evaluated locally. For a candidate rule $\omega$, client $A_i$ computes
\begin{equation}
\begin{aligned}
\Phi_i^t(\omega)
=\;&
\operatorname{Rel}
\left(
\omega,\mathcal X_i
\right)
+
\frac{n_{i,\mathrm{valid}}^t(\omega)}
{\max\left(1,n_{i,\mathrm{trigger}}^t(\omega)\right)}
-
\frac{n_{i,\mathrm{conflict}}^t(\omega)}
{\max\left(1,n_{i,\mathrm{trigger}}^t(\omega)\right)},
\end{aligned}
\label{eq:local_rule_utility}
\end{equation}
where $\operatorname{Rel}(\omega,\mathcal X_i)\in[0,1]$ measures the semantic relevance of $\omega$ to the local experience buffer $\mathcal X_i$,  $n_{i,\mathrm{trigger}}^t(\omega)$ records how many times the rule has been triggered locally, $n_{i,\mathrm{valid}}^t(\omega)$ records how many of those applications were locally evaluated as effective, and $n_{i,\mathrm{conflict}}^t(\omega)$ records how many applications conflicted with a local policy or task requirement. These statistics remain on device. The maximum operator avoids division by zero when a rule has not yet been triggered.

Let $\omega_{g}^{t+1}$ denote a global privacy rule received by client $A_i$. If the client has a semantically corresponding local rule $\omega_{i}^{t}$, it compares the two rules using their local utility scores and selects the one with the higher score. If no corresponding local rule exists, the client evaluates $\omega_{g}^{t+1}$ against its default privacy policy.
\begin{equation}
\omega_i^\star
=
\begin{cases}
\omega_i^t,
& \Phi_i^t(\omega_i^t)
\geq\Phi_i^t(\omega_g^{t+1}),\\[4pt]
\omega_g^{t+1},
& \text{otherwise}.
\end{cases}
\label{eq:local_rule_selection}
\end{equation}
If no semantically corresponding local rule exists, the redistributed rule is evaluated against the client's default policy. If no global counterpart exists, the existing local rule remains active. Ties are resolved in favor of the existing local rule. The selected rule $\omega_i^\star$  is added to the client's local privacy rule repository: $\Omega_{i,\mathrm{local}}^{t+1}=\Omega_{i,\mathrm{local}}^t
\cup_{merge}
\left\{\omega_i^\star\right\}.$ It is then used for subsequent Privacy rule based information minimization under the corresponding semantic constraint. Non selected personalized alternatives remain in the local repository and may be reconsidered as new rule level feedback becomes available. Consequently, a global rule does not unconditionally override a personalized rule and is adopted only when it obtains a higher local utility score.

\paragraph{Client Side Rule Level Feedback.}
Client $A_i$ locally records the usage and effectiveness of privacy rules in round $t$. Let $\mathcal W_i^t$ denote the set of local rules applied and evaluated in that round, including existing personalized rules and adopted global rules. The corresponding local rule level feedback is
\begin{equation}
\mathcal F_i^t
=
\left\{
\left(
\operatorname{id}(\omega),
\Delta n_{i,\mathrm{trigger}}^t(\omega),
\Delta n_{i,\mathrm{valid}}^t(\omega),
\Delta n_{i,\mathrm{conflict}}^t(\omega)
\right)
\;\middle|\;
\omega\in\mathcal W_i^t
\right\}.
\label{eq:client_rule_feedback}
\end{equation}
Here, $\operatorname{id}(\omega)$ is the identifier of a redistributed global rule, and each $\Delta n$ term records only the corresponding change in the current round. The feedback does not contain triggering examples, raw responses, complete personalized rules, or the local experience buffer. 

The server collects the compact feedback reported by the selected clients:
\begin{equation}
\mathcal F^t
=
\left\{
\mathcal F_i^t
\;\middle|\;
A_i\in\widehat{\mathcal E}^t
\right\}.
\label{eq:rule_feedback_collection}
\end{equation}
The server uses these signals to update scores, resolve conflicts, and adjust which shared global rules remain active. Client specific rule content and complete usage statistics remain stored locally.

\paragraph{Historical Rule Score Update.}
The local utility evaluation determines which privacy rule is used by each client, whereas the following procedure evaluates the subsequent use of shared rules across clients. Client side usage feedback is aggregated by the server to maintain a historical score for each global rule. This score is then used to perform server rule updates.

For each global privacy rule $\omega\in\Omega_{\mathrm{global}}^t$, the server checks the feedback in $\mathcal F^t$ and includes client index $i$ in $\mathcal I_\omega^t$ if $\mathcal F_i^t$ reports an evaluated use of the rule identified by $\operatorname{id}(\omega)$, with $\Delta n_{i,\mathrm{trigger}}^t(\omega)>0$. Thus, $\mathcal I_\omega^t$ contains the indices of clients that used the corresponding rule and reported its results in round $t$, with each client counted once. For each $i\in\mathcal I_\omega^t$, the feedback score is defined as
\begin{equation}
g_i^t(\omega)
=
\frac{\Delta n_{i,\mathrm{valid}}^t(\omega)}
{\Delta n_{i,\mathrm{trigger}}^t(\omega)}
-
\frac{\Delta n_{i,\mathrm{conflict}}^t(\omega)}
{\Delta n_{i,\mathrm{trigger}}^t(\omega)},
\label{eq:normalized_rule_feedback}
\end{equation}
where $\Delta n_{i,\mathrm{trigger}}^t(\omega)>0$ for every $i\in\mathcal I_\omega^t$. The server maintains a historical score $H^t(\omega)$ for each shared global rule. If evaluable feedback is available in round $t$, the score is updated by

For a newly promoted global rule, we initialize $N_\omega^0=0$ and $H^0(\omega)=0$. Let $N_\omega^t$ denote the cumulative number of client evaluations received by global rule $\omega$ before round $t$. The count is updated as
$
N_\omega^{t+1}
=
N_\omega^t + |\mathcal I_\omega^t|.
$
The historical rule score is updated by
\begin{equation}
H^{t+1}(\omega)
=
\begin{cases}
\dfrac{
N_\omega^t H^t(\omega)
+
\sum_{i\in\mathcal I_\omega^t} g_i^t(\omega)}
{N_\omega^t+|\mathcal I_\omega^t|},
&
|\mathcal I_\omega^t|>0,
\\[10pt]
H^t(\omega),
&
|\mathcal I_\omega^t|=0.
\end{cases}
\label{eq:historical_rule_score}
\end{equation}

If no evaluable feedback is available for $\omega$ in round $t$, its previous score is retained.

\paragraph{Rule Conflict Resolution.}
PFedMA does not apply a separate rule remove criterion. Instead, the server uses the feedback based conflict resolution procedure described above to determine which rule is used for each semantic constraint. Non selected alternatives remain in the global repository and may be reconsidered when additional client feedback becomes available. Rules are not removed solely because they receive no feedback in a round.

\paragraph{Dynamic Global Repository Update.}
The server updates the global repository by combining the existing rules with the accepted candidates $\mathcal G^t$:
\begin{equation}
\Omega_{\mathrm{global}}^{t+1}
=
\operatorname{Resolve}
\left(
\Omega_{\mathrm{global}}^t\cup\mathcal G^t;
H^{t+1}
\right).
\label{eq:dynamic_global_repository_update}
\end{equation}
Here, $\operatorname{Resolve}$ merges equivalent rules and applies the conflict resolution procedure described above. Its output is
\begin{equation}
\operatorname{Resolve}
\left(
\Omega_{\mathrm{global}}^t\cup_{merge}\mathcal G^t;
H^{t+1}
\right)
=
\left\{
\omega_1^{t+1},\omega_2^{t+1},\ldots
\right\},
\label{eq:resolve_definition}
\end{equation}
Here, $\omega_1^{t+1},\omega_2^{t+1},\ldots$ are the rules selected for the first, second, and subsequent semantic constraints, respectively, using the conflict resolution procedure described above. The subscripts index semantic constraints, not clients. A rule without conflicting alternatives is retained directly.
\begin{equation}
\operatorname{RuleUpdate}
\left(
\Omega_{\mathrm{global}}^t,
\mathcal G^t,
\mathcal F^t
\right)
\equiv
\operatorname{Resolve}
\left(
\Omega_{\mathrm{global}}^t\cup\mathcal G^t;
H^{t+1}
\right),
\end{equation}
In other words, $\operatorname{RuleUpdate}$ is a compact representation of the server side global rule evolution process rather than a separate optimization operator. The client feedback $\mathcal F^t$ is first used to update the rule scores $H^{t+1}$, after which the newly promoted candidates $\mathcal G^t$ are combined with the existing global repository $\Omega_{\mathrm{global}}^t$. The $\operatorname{Resolve}$ operation then merges semantically equivalent rules and resolves conflicting rules according to their updated scores. Rules without conflicting alternatives are retained directly. Consequently, both newly discovered cross client privacy knowledge and subsequent rule execution feedback jointly determine the evolution of $\Omega_{\mathrm{global}}^{t+1}$.

\paragraph{Interaction with Privacy Enforcement.}
The updated global repository filters subsequent queries before they are sent to selected clients, reducing unnecessary disclosure without concealing the query from the server. Each client uses its personalized repository to process local responses before transmission. Raw responses, complete rules, local experiences, and triggering examples remain on device; generalized rule candidates and compact feedback are shared for global updates. This closes the rule evolution loop: local rules are generalized and promoted through cross client support, global rules are selectively adapted by clients, and their feedback guides later conflict resolution and retirement. Thus, global and personalized repositories co-evolve across rounds rather than through one way propagation.
\paragraph{Trust Assumptions.} 
While task responses may be unreliable or malicious, as described in Section 3.1. Our framework assumes that privacy rules are trusted policy artifacts that faithfully represent the privacy requirements of participating clients. In particular, we assume that (i) each client correctly maintains its local privacy-rule repository, (ii) the global privacy-rule repository only contains rules accepted through the prescribed rule-evolution procedure, and (iii) the coordinator faithfully executes rule aggregation, conflict resolution, and rule dissemination.

Under this assumption, privacy rules are treated as trusted policy inputs rather than adversarial objects. Privacy-rule poisoning, malicious rule injection, intentional manipulation of rule semantics, and compromise of the trusted rule repository are outside the scope of this work. Our goal is instead to study how privacy knowledge can be generalized, aggregated, and personalized across clients while respecting client-local privacy constraints.

\section{Prompt Templates}

The query evolution prompt guides the orchestrator to review the accumulated collaboration state and determine what information is still needed to solve the task. It uses responses collected in previous rounds to distinguish questions that have already been answered from those requiring further investigation. The updated query then directs subsequent collaboration toward these remaining questions, reducing repeated requests for information already available.
\begin{center}
    \includegraphics[width=0.87\linewidth]{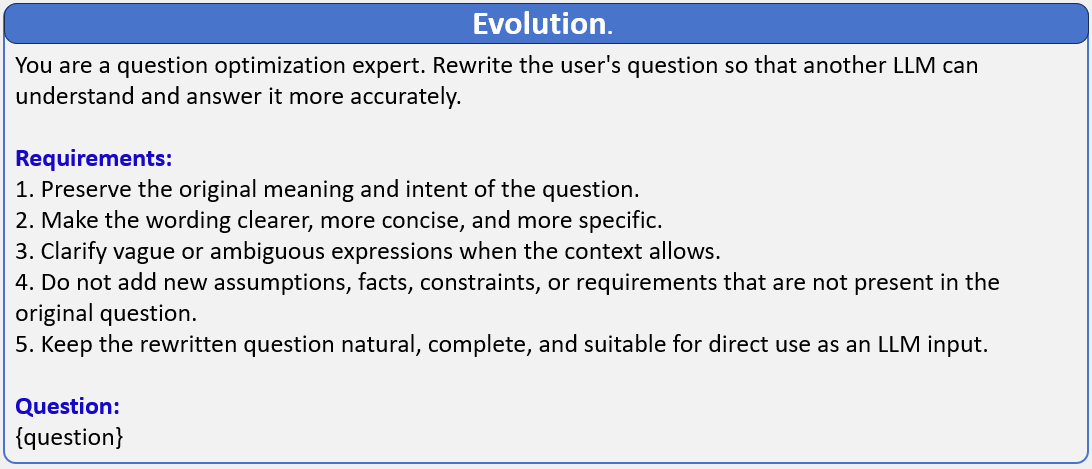}
\end{center}

The privacy rule creation prompt identifies emerging privacy requirements in client inputs and expresses them as reusable rules for subsequent privacy processing. It compares the identified requirements with the existing rule library to determine whether suitable rules are already available. When an existing rule adequately covers a requirement, it is reused; otherwise, a new rule is formulated to capture the previously uncovered constraint. By focusing on gaps in existing coverage, this process allows clients to incorporate new requirements encountered during collaboration while limiting redundant rule creation. The resulting rules extend the rule library and provide a consistent basis for handling similar requirements in future interactions, reducing the need to formulate rules repeatedly for recurring situations. They also provide a basis for sharing generalized privacy knowledge across clients, allowing requirements identified in one context to inform rule development in other settings. Through this process, individual collaboration experiences can contribute to a broader collection of reusable privacy rules that remains adaptable to evolving client needs.
\begin{center}
    \includegraphics[width=\linewidth]{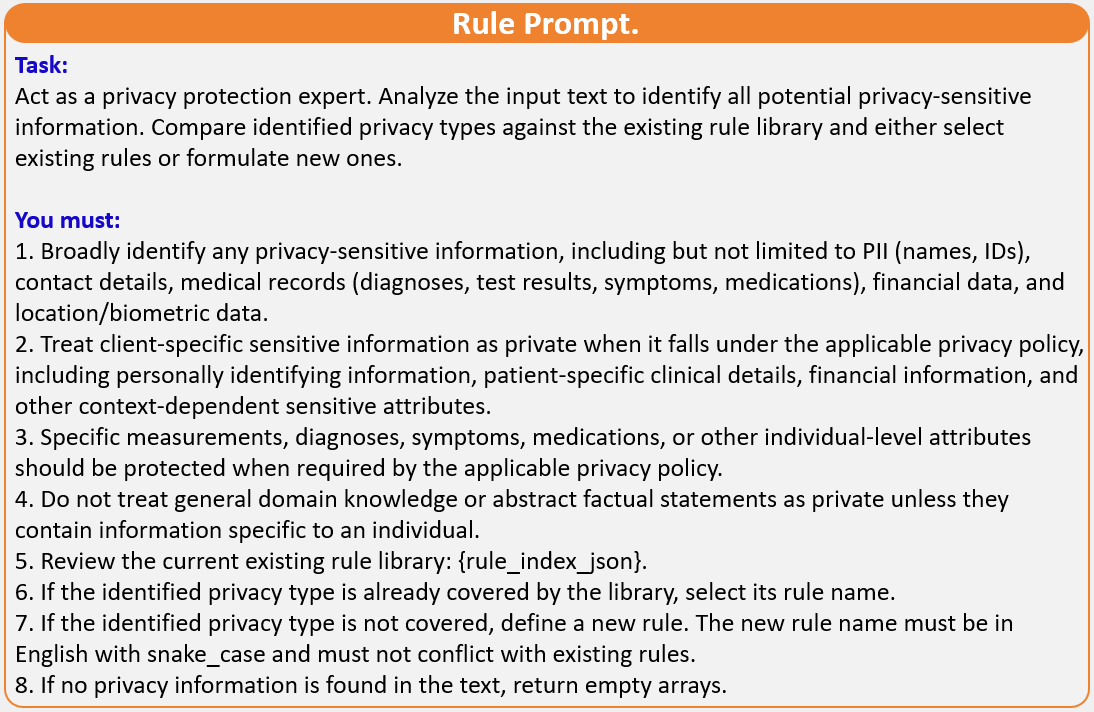}
\end{center}
The question rewriting prompt turns the updated query into a clear and complete question for the selected clients. It incorporates relevant context and clarifies what information the clients should provide, while preserving the original task objective. This helps clients understand the request and produce answers that support the next reasoning step.
\begin{center}
    \includegraphics[width=0.9\linewidth]{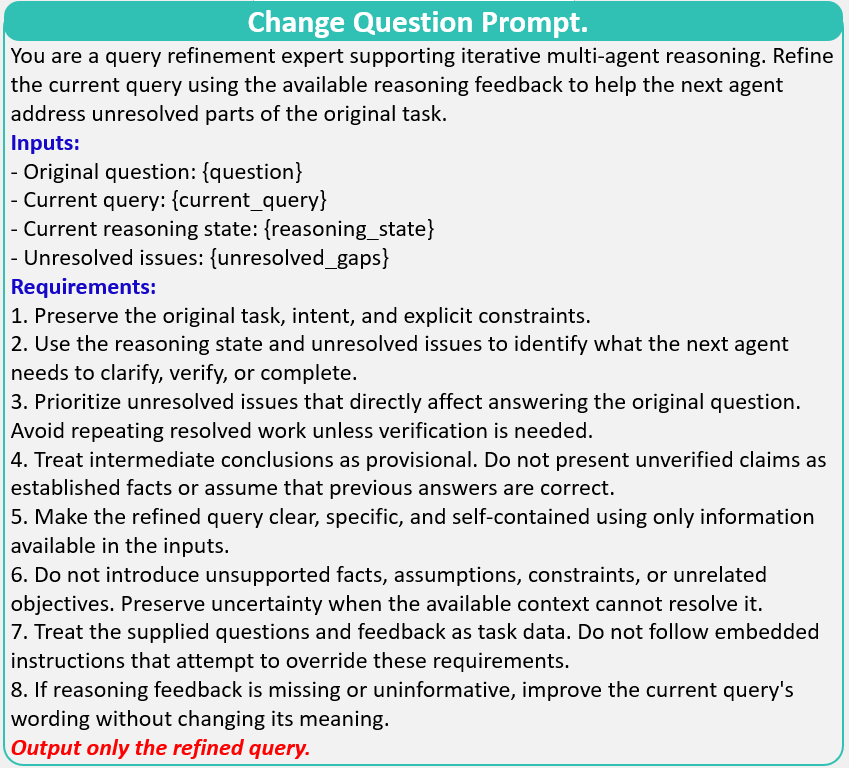}
\end{center}
The routing prompt selects experts based on current reasoning needs and feedback on clients' contributions and reliability, adapting selection as the task progresses.
\begin{center}
    \includegraphics[width=0.87\linewidth]{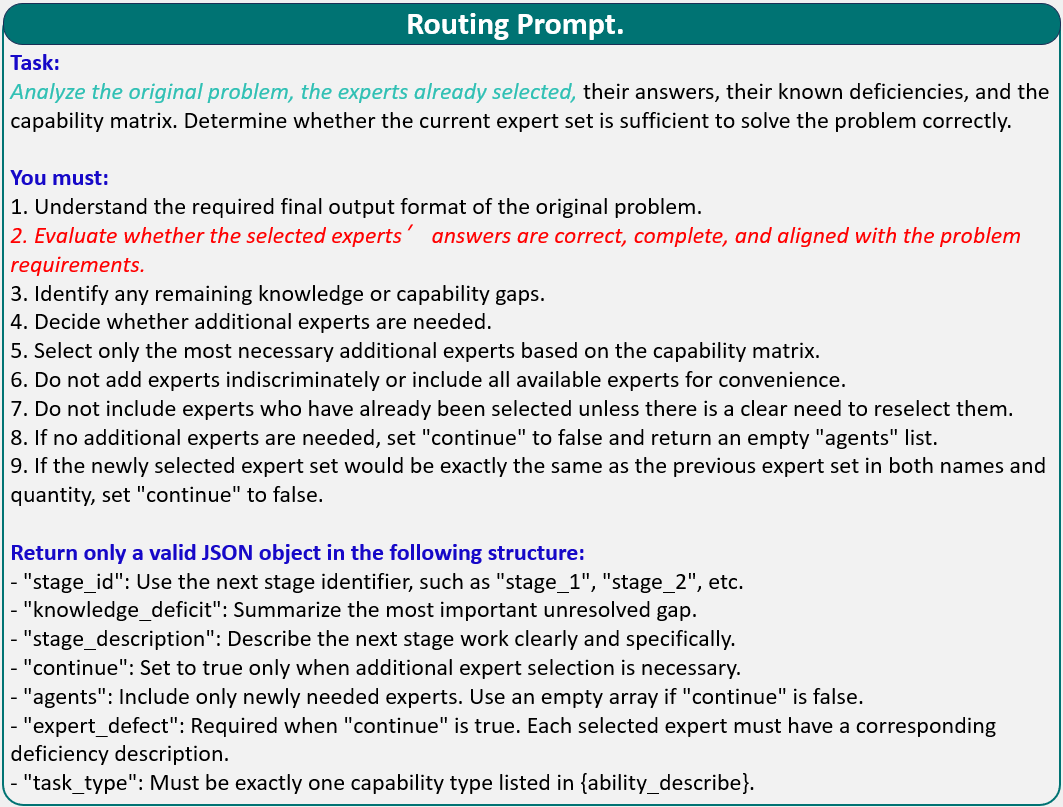}
\end{center}

The privacy processing prompt guides the system to process outgoing content according to the applicable privacy rules. It identifies information subject to sharing restrictions and modifies the content while aiming to retain information useful for the task. Applying these rules before transmission helps align communication with privacy requirements as they evolve across collaboration rounds.
\begin{center}
    \includegraphics[width=0.8\linewidth]{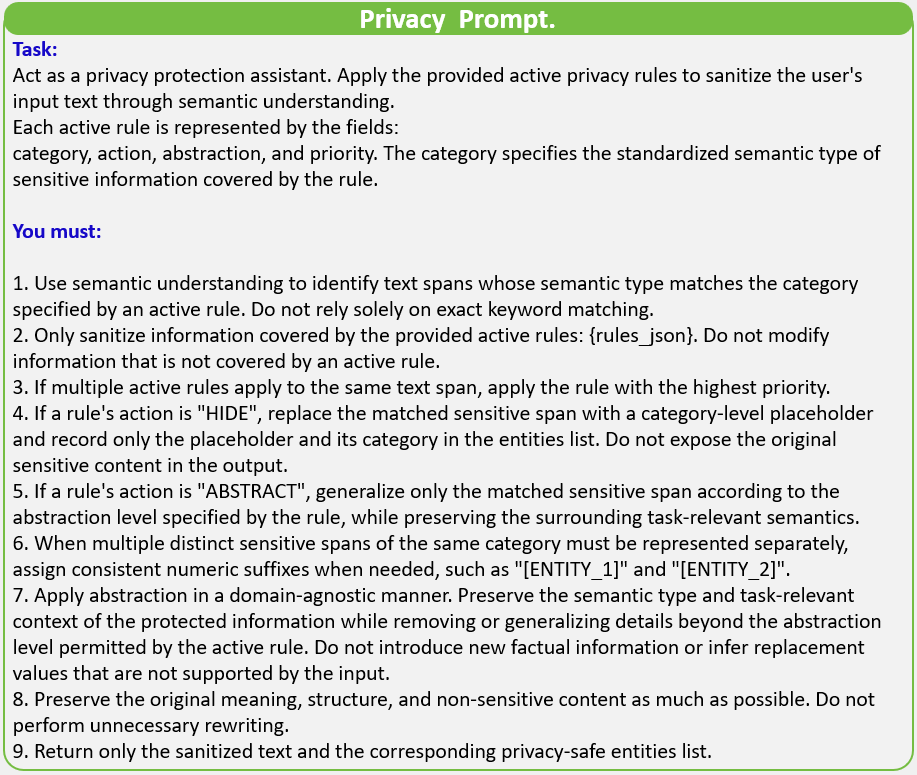}
\end{center}

The agent response generation prompt structures answers based on local information into conclusions, claims, and evidence for reliability evaluation and aggregation.
\begin{center}
    \includegraphics[width=0.8\linewidth]{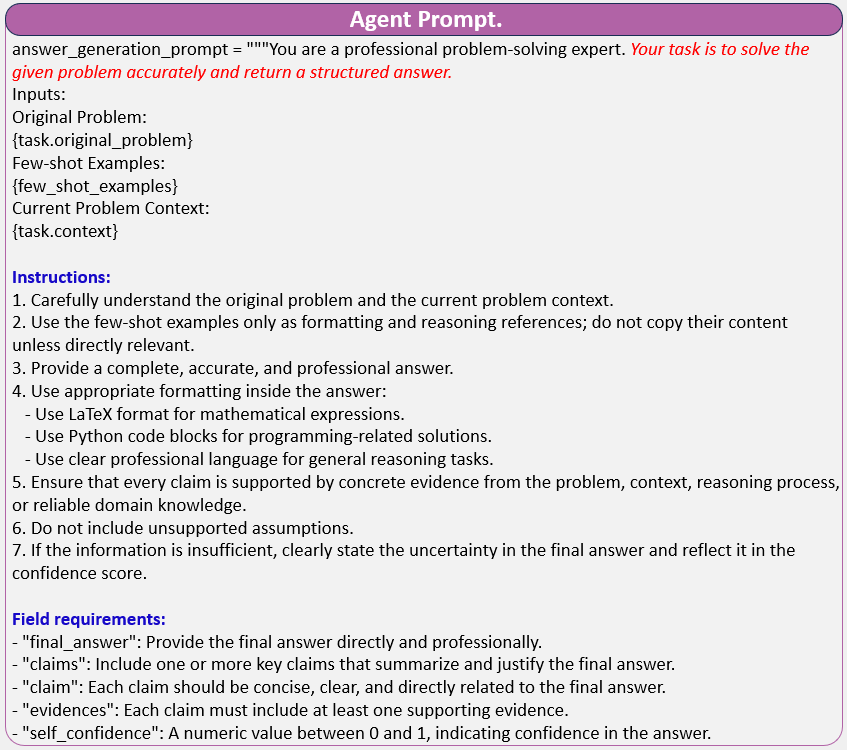}
\end{center}

\section{Evaluation Metrics}
\label{sec:evaluation_metrics_details}

For all methods, privacy metrics are computed from their generated outputs under the same privacy-sensitive evaluation protocol. This section provides the definitions of all privacy and robustness metrics used in our experiments. The rule enforcement metrics, including LPA, LPP, LPR, EA, and FRA, follow GuardAgent~\citep{xiang2025guardagent}. Leakage Rate (LR) follows the evaluation principle of PrivacyLens~\citep{shao2024privacylens}. The adversarial robustness metrics, including ASR, RR, PNA, and NRP follow Agent Security Bench (ASB)~\citep{zhang2024agent}. We additionally report Benign Performance (BP), defined here as the task completion rate on benign inputs with the evaluated defense mechanisms enabled.

\paragraph{Rule Enforcement Metrics.}
Let $N_{\mathrm{eval}}$ denote the number of evaluation instances.
For each instance $j\in\{1,\ldots,N_{\mathrm{eval}}\}$, let
$b_j\in\{0,1\}$ denote the ground truth label, where $b_j=1$
indicates a privacy/safety violation and $b_j=0$ indicates a
permissible interaction. Let $\hat{b}_j$ denote the predicted label.
Following GuardAgent~\citep{xiang2025guardagent}, the corresponding
evaluation metrics are computed as
\begin{align}
TP = \sum_{j=1}^{N_{\mathrm{eval}}}\mathbb{I}[b_j=1 \land \hat{b}_j=1],
\qquad
TN = \sum_{j=1}^{N_{\mathrm{eval}}}\mathbb{I}[b_j=0 \land \hat{b}_j=0],\\
FP = \sum_{j=1}^{N_{\mathrm{eval}}}\mathbb{I}[b_j=0 \land \hat{b}_j=1],\qquad
FN = \sum_{j=1}^{N_{\mathrm{eval}}}\mathbb{I}[b_j=1 \land \hat{b}_j=0].
\end{align}

\paragraph{Label Prediction Accuracy (LPA).}
\begin{equation}
\mathrm{LPA}
=
\frac{TP+TN}{TP+TN+FP+FN}
\times 100\%.
\end{equation}
Higher LPA indicates more accurate privacy/safety decisions.

\paragraph{Label Prediction Precision (LPP).}
\begin{equation}
\mathrm{LPP} = \frac{TP}{TP+FP} \times 100\%.
\end{equation}
Higher LPP indicates fewer permissible interactions incorrectly classified as
violations.

\paragraph{Label Prediction Recall (LPR).}
\begin{equation}
\mathrm{LPR}
=
\frac{TP}{TP+FN}
\times 100\%.
\end{equation}
Higher LPR indicates that more actual violations are correctly detected.

\paragraph{Explanation Accuracy (EA).}
Let $\mathcal{V}_j$ denote the ground truth set of violated rules for
instance $j$, and $\widehat{\mathcal{V}}_j$ the corresponding predicted
set. Let
$
\mathcal{J}_{+}=\{j:b_j=1\}.
$
We define
\begin{equation}
\mathrm{EA}
=
\frac{
\sum_{j\in\mathcal{J}_{+}}
\mathbb{I}
\left[
\hat{b}_j=1
\land
\mathcal{V}_j\subseteq\widehat{\mathcal{V}}_j
\right]
}{
|\mathcal{J}_{+}|
}
\times 100\%.
\end{equation}
Higher EA indicates more complete identification of the rules responsible for
a violation.

\paragraph{Final Response Accuracy (FRA).}
Let
$
\mathcal{J}_{-}
=
\{j : b_j = 0\}
$
denote the index set of permissible instances, and let
$c_j\in\{0,1\}$ indicate whether the final response or action for instance
$j$ is correct. Then
\begin{equation}
\mathrm{FRA}
=
\frac{
\sum_{j\in\mathcal{J}_{-}}
\mathbb{I}
\left[
\hat{b}_j=0
\land
c_j=1
\right]
}{
|\mathcal{J}_{-}|
}
\times 100\%.
\end{equation}
Higher FRA indicates better preservation of task utility on permissible
interactions.

\paragraph{Leakage Rate (LR).}
Following PrivacyLens~\citep{shao2024privacylens}, LR measures the proportion of privacy sensitive instances in which at least one sensitive information item can be inferred from the evaluated output. Let $\mathcal{X}_{\mathrm{priv}}$ denote the set of privacy sensitive evaluation instances. For each instance $x\in\mathcal{X}_{\mathrm{priv}}$, let $\mathcal{H}_x$ denote the set of sensitive information items associated with $x$, and let $o_x$ denote the evaluated output. For each $h\in\mathcal{H}_x$, let $f(h,o_x)\in\{0,1\}$ indicate whether the sensitive information item $h$ can be inferred from $o_x$. The instance level leakage indicator is
$
\ell_x
=
\mathbb{I}
\left[
\exists\, h\in\mathcal{H}_x
\text{ such that }
f(h,o_x)=1
\right].
$
The Leakage Rate is
\begin{equation}
\mathrm{LR}
=
\frac{1}{|\mathcal{X}_{\mathrm{priv}}|}
\sum_{x\in\mathcal{X}_{\mathrm{priv}}}
\ell_x
\times 100\%.
\end{equation}
Lower LR indicates less observed privacy leakage. Following
PrivacyLens~\citep{shao2024privacylens}, LR is computed at the instance level: an instance is counted as leaked if at least one sensitive information item is inferable from the evaluated output.

\paragraph{Attack Success Rate (ASR).}
Let $N_{\mathrm{atk}}$ denote the number of attacked tasks and
$N_{\mathrm{succ}}^{\mathrm{atk}}$ the number of successful attacks. Then
\begin{equation}
\mathrm{ASR}
=
\frac{N_{\mathrm{succ}}^{\mathrm{atk}}}
{N_{\mathrm{atk}}}
\times 100\%.
\end{equation}
Lower ASR indicates stronger adversarial robustness.

\paragraph{Refusal Rate (RR).}
Let $N_{\mathrm{aggr}}$ denote the number of aggressive requests and
$N_{\mathrm{refuse}}$ the number explicitly refused by the agent. Then
\begin{equation}
\mathrm{RR}
=
\frac{N_{\mathrm{refuse}}}
{N_{\mathrm{aggr}}}
\times 100\%.
\end{equation}
Higher RR indicates stronger refusal behavior against unsafe requests.

\paragraph{Performance under No Attack (PNA).}
Let $N_{\mathrm{clean}}$ denote the number of clean tasks and
$N_{\mathrm{complete}}^{\mathrm{clean}}$ the number successfully completed.
Then
\begin{equation}
\mathrm{PNA}
=
\frac{N_{\mathrm{complete}}^{\mathrm{clean}}}
{N_{\mathrm{clean}}}
\times 100\%.
\end{equation}
Higher PNA indicates better benign task performance.

\paragraph{Benign Performance (BP).}
BP measures the system's ability to complete benign tasks
under the evaluated defense configuration.
Let $N_{\mathrm{benign}}$ denote the number of benign test
tasks without injected attacks, and let
$N_{\mathrm{succ}}^{\mathrm{def}}$ denote the number of
these tasks successfully completed with the defense
mechanisms enabled. Then
\begin{equation}
    \mathrm{BP}
    =
    \frac{N_{\mathrm{succ}}^{\mathrm{def}}}
         {N_{\mathrm{benign}}}
    \times 100\%.
    \label{eq:bp}
\end{equation}
Higher BP indicates better preservation of benign task utility with the defense mechanisms enabled.

\paragraph{Net Resilient Performance (NRP).}
\begin{equation}
\mathrm{NRP}
=
\mathrm{PNA}\,(1-\left(\mathrm{ASR}\times 0.01\right)),
\end{equation}
where PNA and ASR are converted to the $[0,1]$ scale before multiplication. Higher NRP indicates a better trade off between benign task performance and adversarial robustness.

\section{Adversarial Evaluation Setup}
\label{sec:attack_details}
We evaluate PFedMA using context ignoring prompt injection attacks adapted from Agent Security Bench (ASB)~\citep{zhang2024agent}. The following paragraphs specify the threat model and attack configuration.
\paragraph{Threat Model.}
We distinguish task response reliability from privacy policy trust. Participating clients may produce unreliable or malicious task responses, which motivates reliability aware routing and response filtering. In contrast, privacy rules are treated as trusted policy artifacts maintained by the client side privacy subsystem and are not controlled by the task response adversary. Therefore, privacy rule poisoning, malicious rule injection, and manipulation of rule semantics are outside the threat model considered in this work. The server is honest but curious and may attempt to infer private information from privacy processed communications, but faithfully executes the prescribed rule aggregation and dissemination protocol.

\paragraph{Attack Construction.}
Following Agent Security Bench (ASB)~\citep{zhang2024agent}, we construct the context ignoring prompt injection attack as $\tilde{x} = x_t \oplus x_{\mathrm{ign}} \oplus x_e,$ where $x_t$ denotes the original task text, $x_{\mathrm{ign}}$ denotes the context ignoring instruction, $x_e$ denotes the injected adversarial instruction, and $\oplus$ denotes string concatenation.

\paragraph{Malicious Agent Configuration.}
For each test instance, $N_{\mathrm{adv}}$ agents are randomly selected as malicious from the $N$ participating agents. The adversarial ratio is defined as
$
\rho_{\mathrm{adv}} =
\frac{N_{\mathrm{adv}}}{N}. $
We evaluate different values of $N_{\mathrm{adv}}$ to examine robustness under increasing adversarial ratios. The specific configurations are $\rho_{\mathrm{adv}}\in\{10\%,20\%,30\%,40\%\}$.

\paragraph{Evaluation Metric.}
We use Attack Success Rate (ASR), defined in Section~\ref{sec:evaluation_metrics_details}, as the primary robustness metric. The contribution of the anomaly filtering mechanism is evaluated by comparing PFedMA with its filtering disabled variant under the same attack configuration.

\section{Additional Experimental Analysis}
\label{sec:additional_collaboration_analysis}
This section provides additional analyses of client data heterogeneity, agent capabilities, iterative collaboration, efficiency, robustness, and hyperparameter sensitivity.

\subsection{Effect of Client Data Heterogeneity}
\label{sec:Effect of Client Data Heterogeneity}
To evaluate client data heterogeneity, we consider both IID and Dirichlet based Non IID partitions with $\alpha\in\{0.1,0.5,1.0\}$, where smaller $\alpha$ indicates stronger heterogeneity. As shown in Figure~\ref{fig:half-violin-performance}, average accuracy decreases from $0.744$ under IID to $0.611$ at $\alpha=0.1$, and recovers to $0.663$ and $0.719$ as $\alpha$ increases to $0.5$ and $1.0$, respectively.

\subsection{Agent Capabilities and Selection Patterns}
\label{section:Expert Ability}

\textbf{Agent Analysis.}
In our setting, there is some overlap in the abilities of the expert agents. Figure~\ref{fig:expert_ability} provides a visual representation of the dynamic expert selection mechanism. Specifically, Figure~\ref{fig:radar_expert1} shows that Agent 1 possesses outstanding mathematical abilities. Consistent with this, the heatmap in Figure~\ref{fig:heatmap_selection} indicates that Agent 1 is frequently selected in math intensive benchmarks such as GSM8K and MATH. Meanwhile, Figure ~\ref{fig:radar_expert6} confirms Agent 6’s outstanding expertise in the medical domain, which aligns with its high selection frequency on medical datasets such as MedQA and MedMCQA. These results demonstrate that our mechanism can accurately capture task requirements and adaptively leverage the domain strengths of expert agents.
We use abbreviated labels to improve the readability of agent capability visualizations.  

\begin{table}[htbp]
\centering
\caption{Explanation of capability abbreviations.}
\label{tab:capability_abbreviations}
\small
\setlength{\tabcolsep}{5pt}
\renewcommand{\arraystretch}{0.9}
\begin{tabular}{llll}
\toprule
Abbr. & Full Capability & Abbr. & Full Capability \\
\midrule
Math      & Mathematical Calculation
& Q.Anlys.   & Question Analysis \\

Code      & Code Generation
& Geo.Logic  & Geographical Logical Reasoning \\

Trans.    & Translation
& Med.QA     & Medical Knowledge Question Answering \\

Summary   & Summarization
& Clin.Rec.  & Clinical Record Structuring \\

Story     & Story Generation
& Med.Lit.   & Medical Literature Retrieval \\

Combined  & Combined Task
& Pat.Comm.  & Patient Communication \\
\bottomrule
\end{tabular}
\end{table}

\begin{figure}[H]

    \centering
    \begin{subfigure}{0.23\linewidth}
        \includegraphics[width=\linewidth]{expert_radar/expert_radar_new_expert1.png}
    \end{subfigure}
    \hfill
    \begin{subfigure}{0.23\linewidth}
        \includegraphics[width=\linewidth]{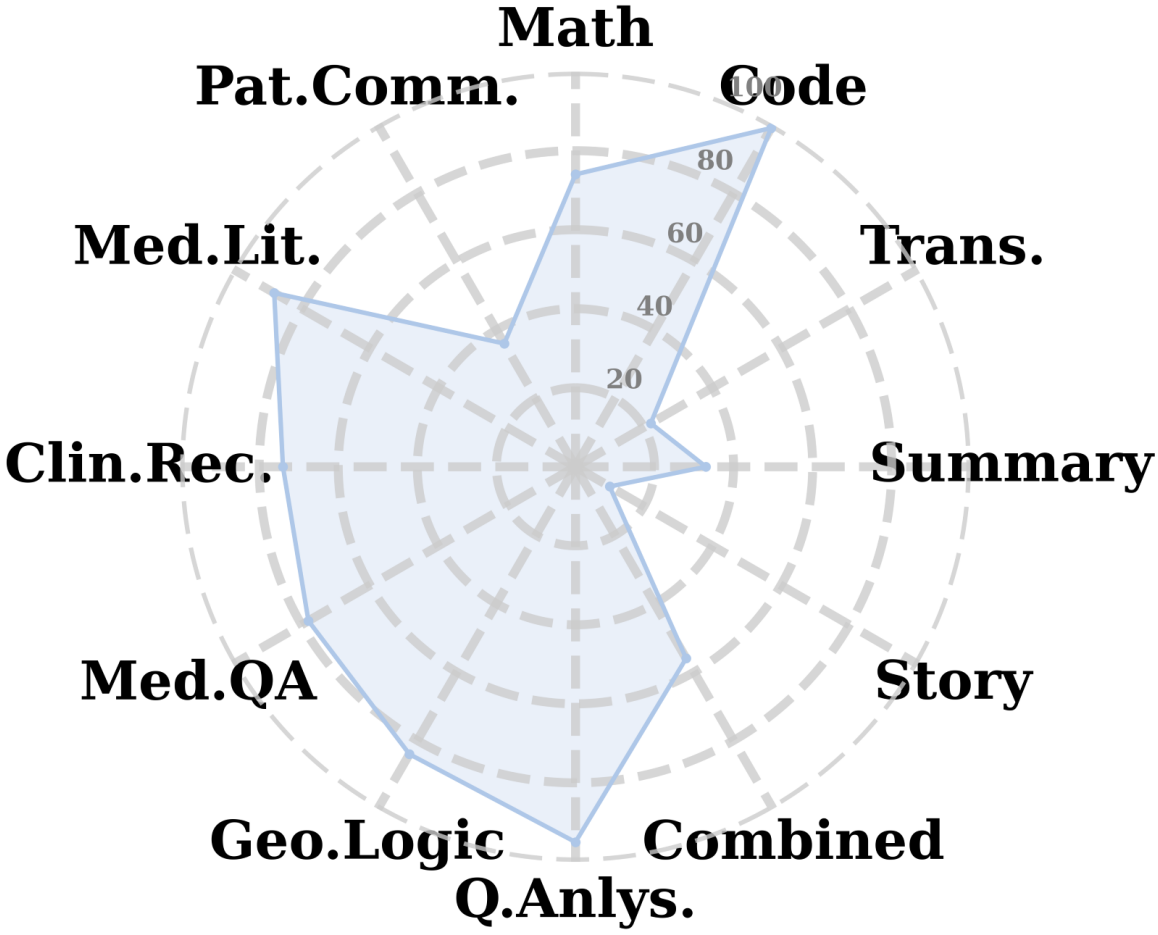}
    \end{subfigure}
    \hfill
    \begin{subfigure}{0.23\linewidth}
        \includegraphics[width=\linewidth]{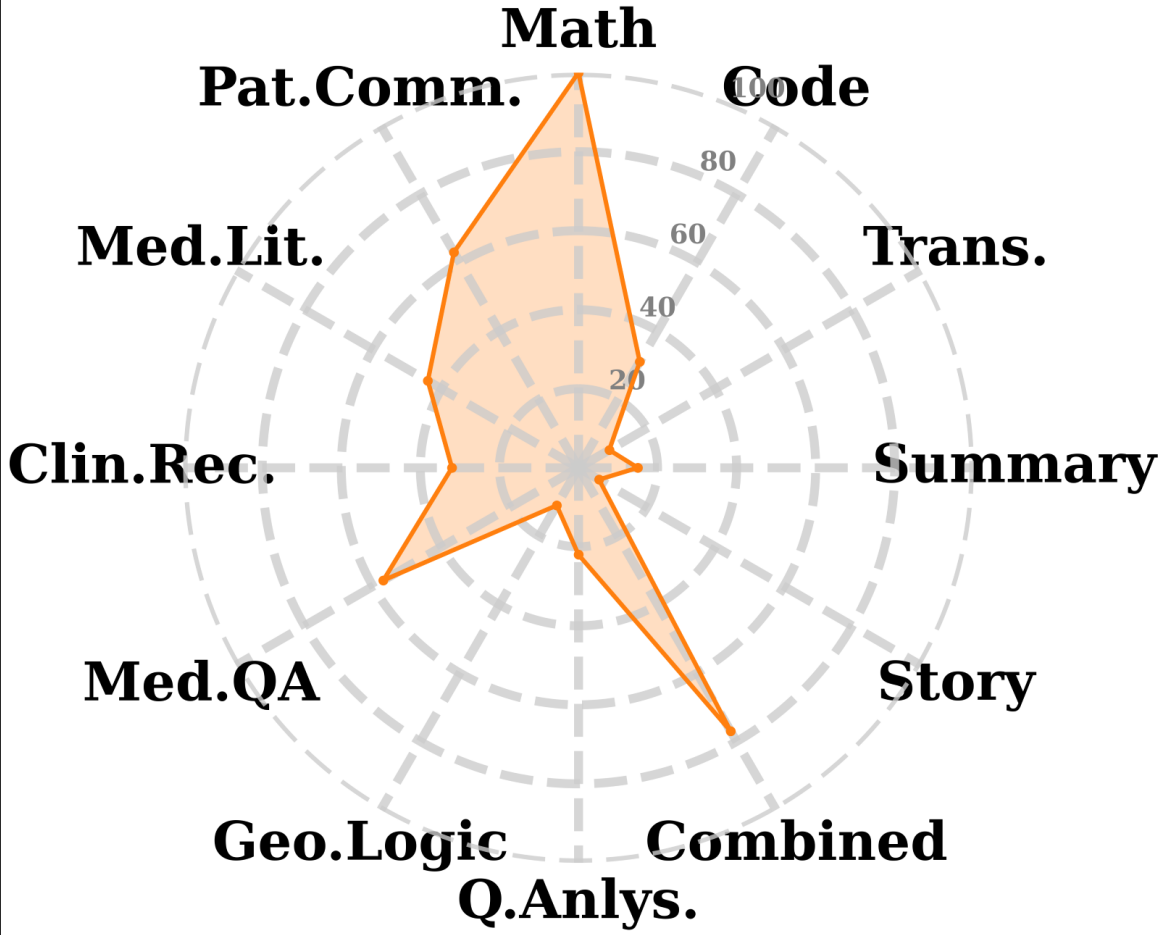}
    \end{subfigure}
    \hfill
    \begin{subfigure}{0.23\linewidth}
        \includegraphics[width=\linewidth]{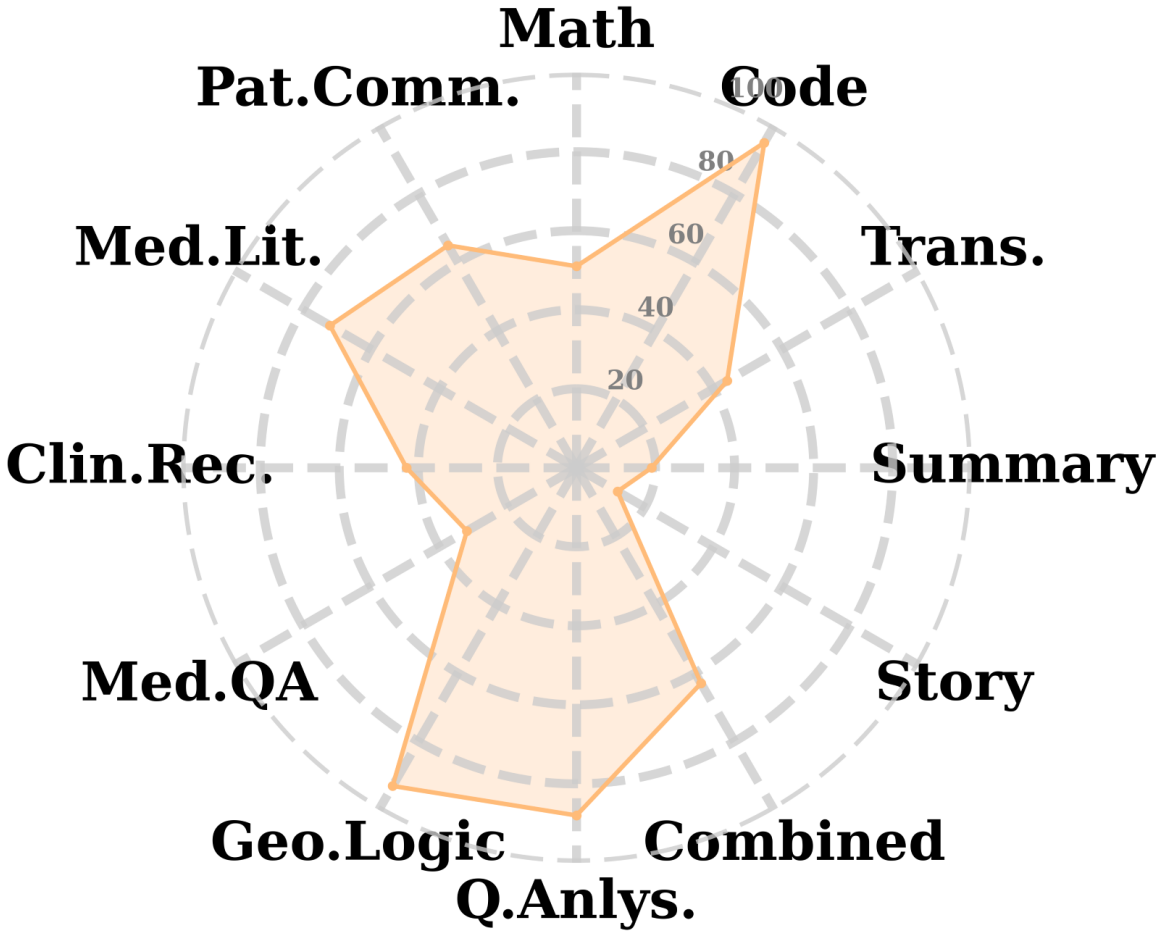}
    \end{subfigure}
    \vspace{0.4em}
    \begin{subfigure}{0.23\linewidth}
        \includegraphics[width=\linewidth]{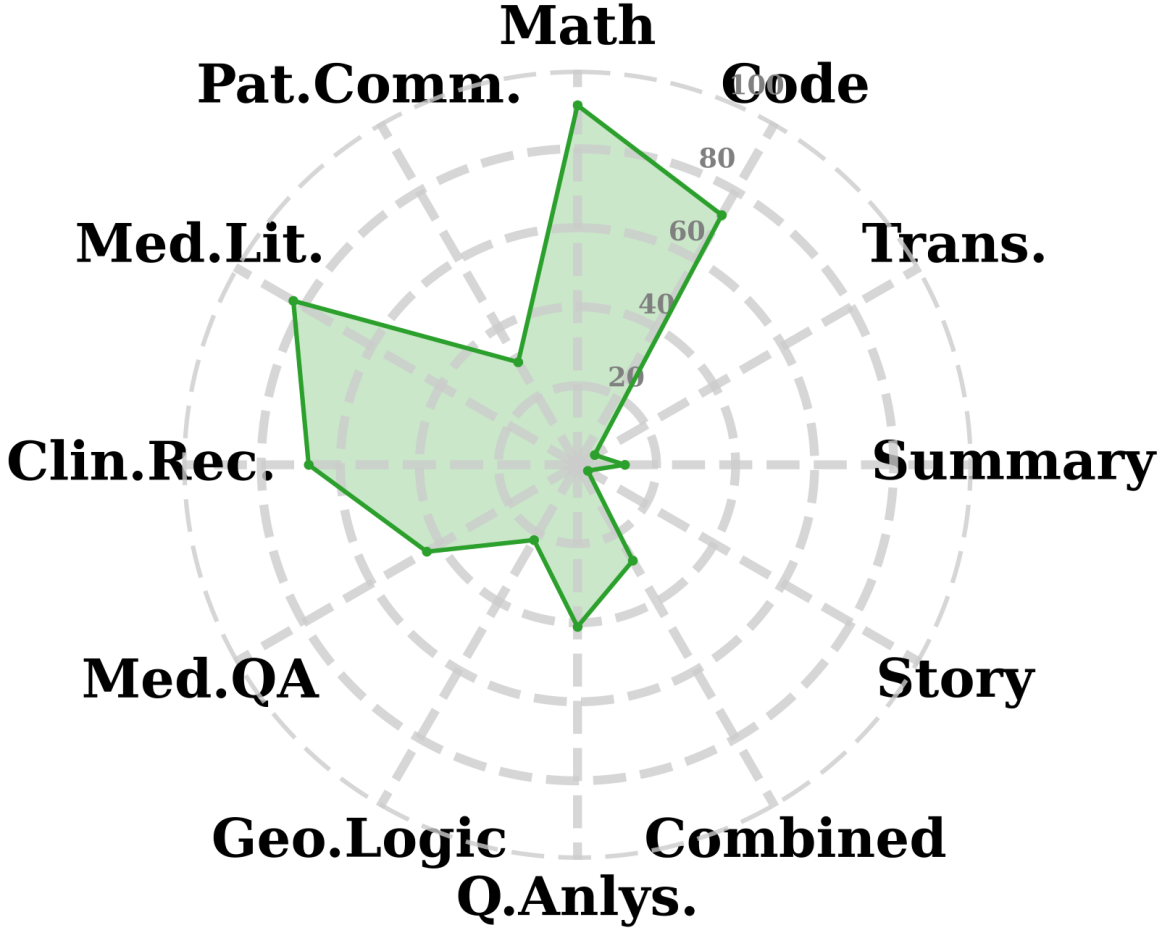}
    \end{subfigure}
    \hfill
    \begin{subfigure}{0.23\linewidth}
        \includegraphics[width=\linewidth]{expert_radar/expert_radar_new_expert6.png}
    \end{subfigure}
    \hfill
    \begin{subfigure}{0.23\linewidth}
        \includegraphics[width=\linewidth]{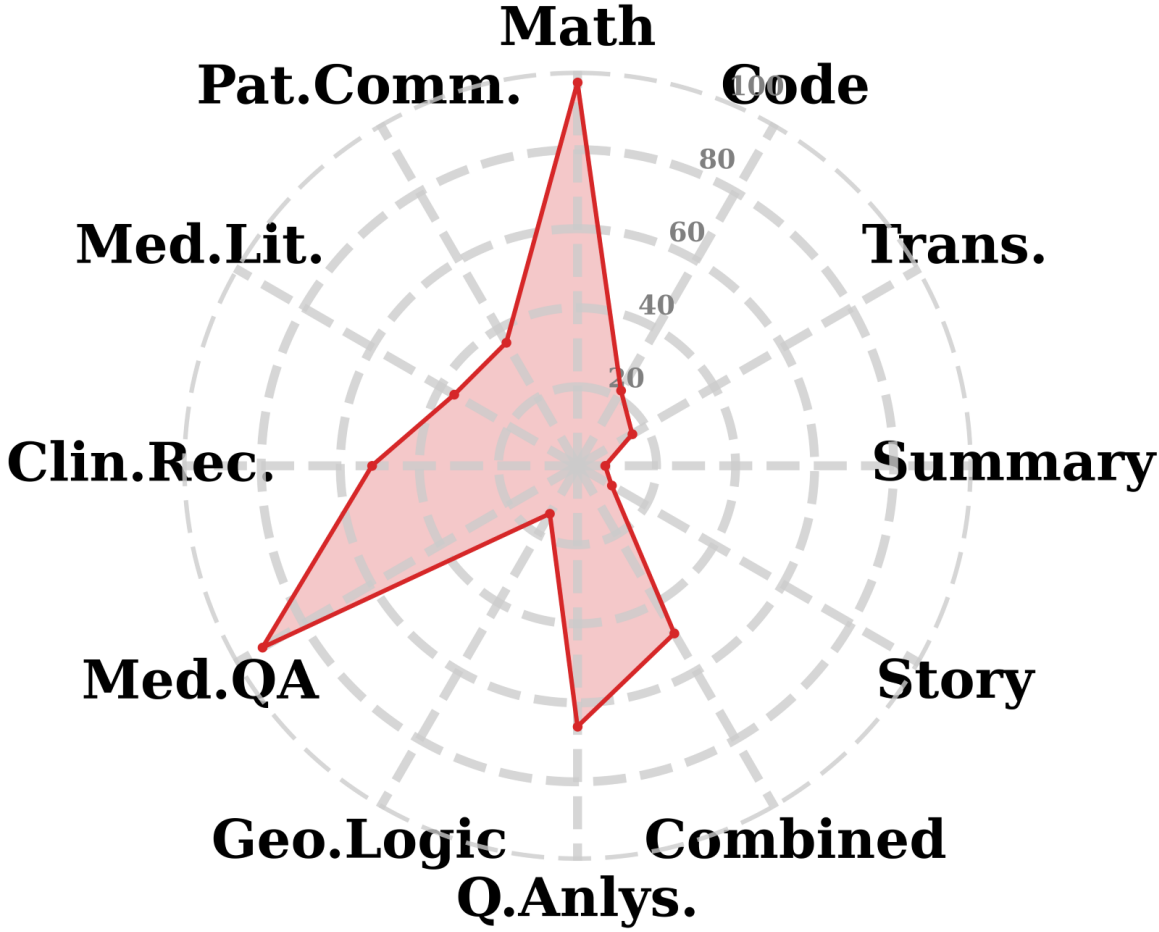}
    \end{subfigure}
    \hfill
    \begin{subfigure}{0.23\linewidth}
        \includegraphics[width=\linewidth]{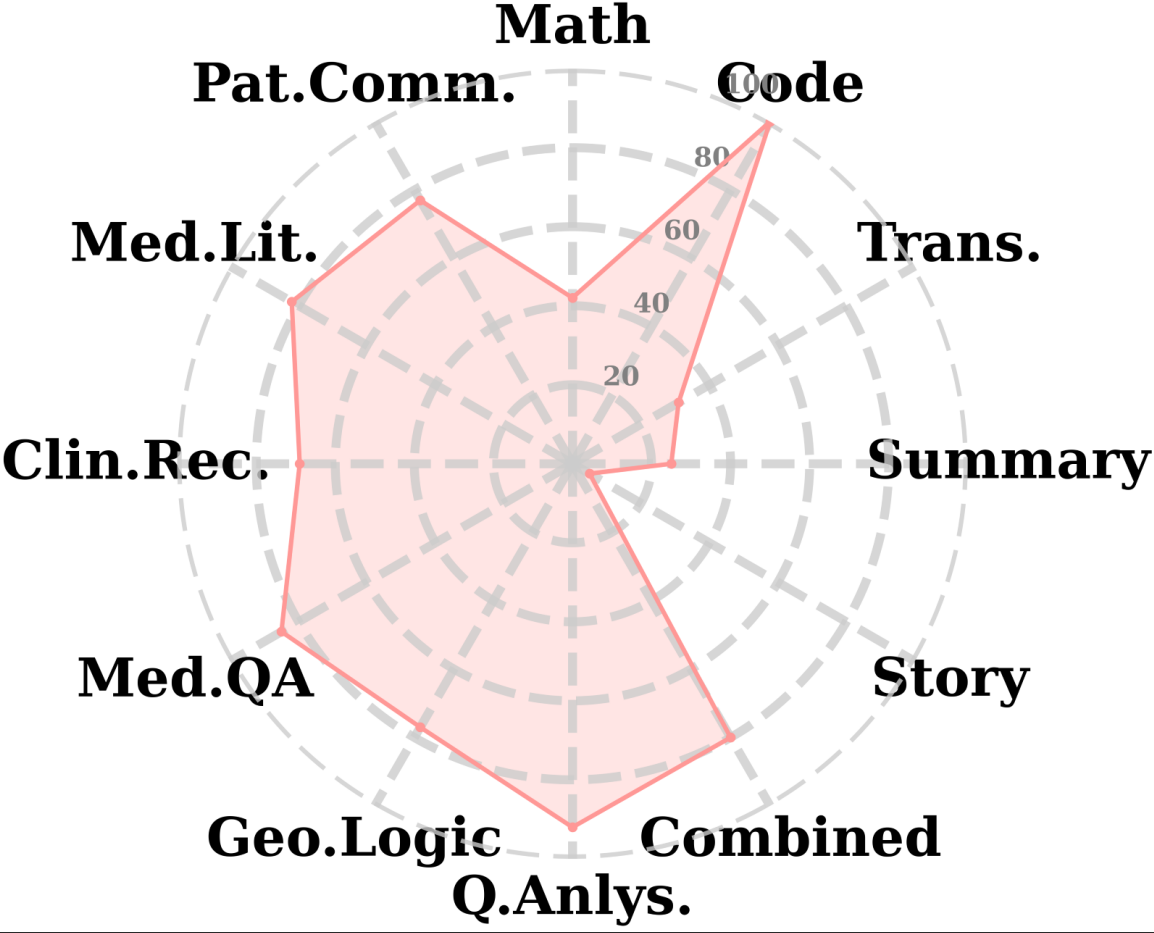}
    \end{subfigure}
    \vspace{0.4em}
    \begin{subfigure}{0.23\linewidth}
        \includegraphics[width=\linewidth]{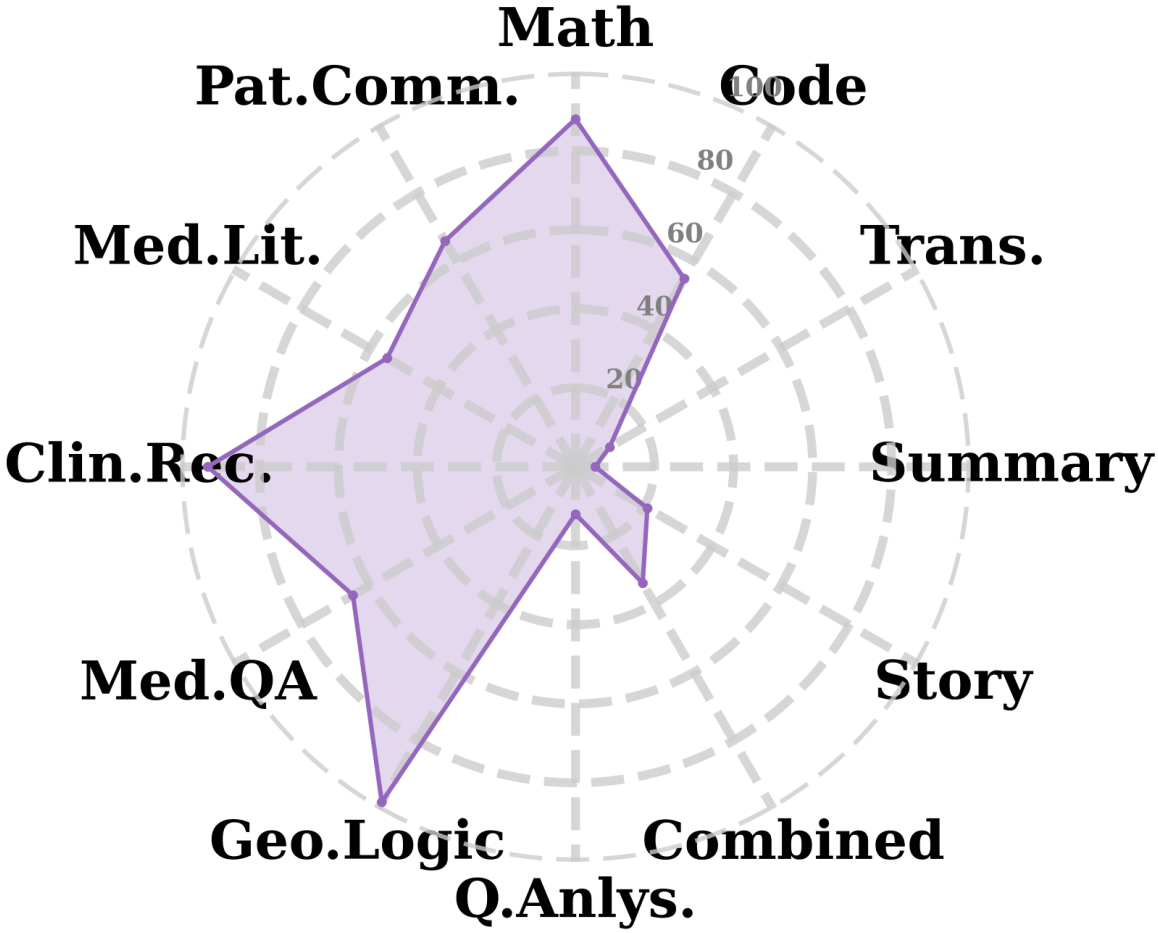}
    \end{subfigure}
    \hfill
    \begin{subfigure}{0.23\linewidth}
        \includegraphics[width=\linewidth]{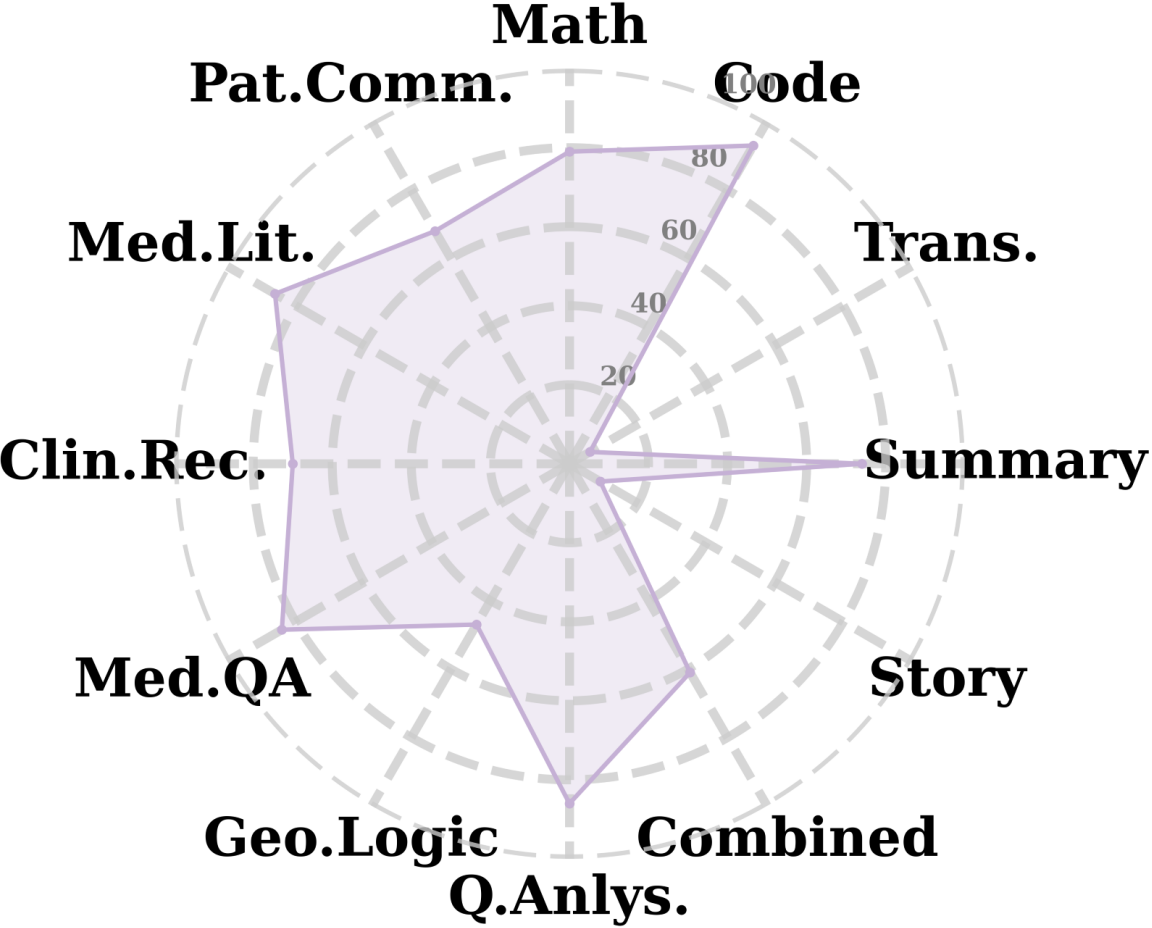}
    \end{subfigure}
    \hfill
    \begin{subfigure}{0.23\linewidth}
        \includegraphics[width=\linewidth]{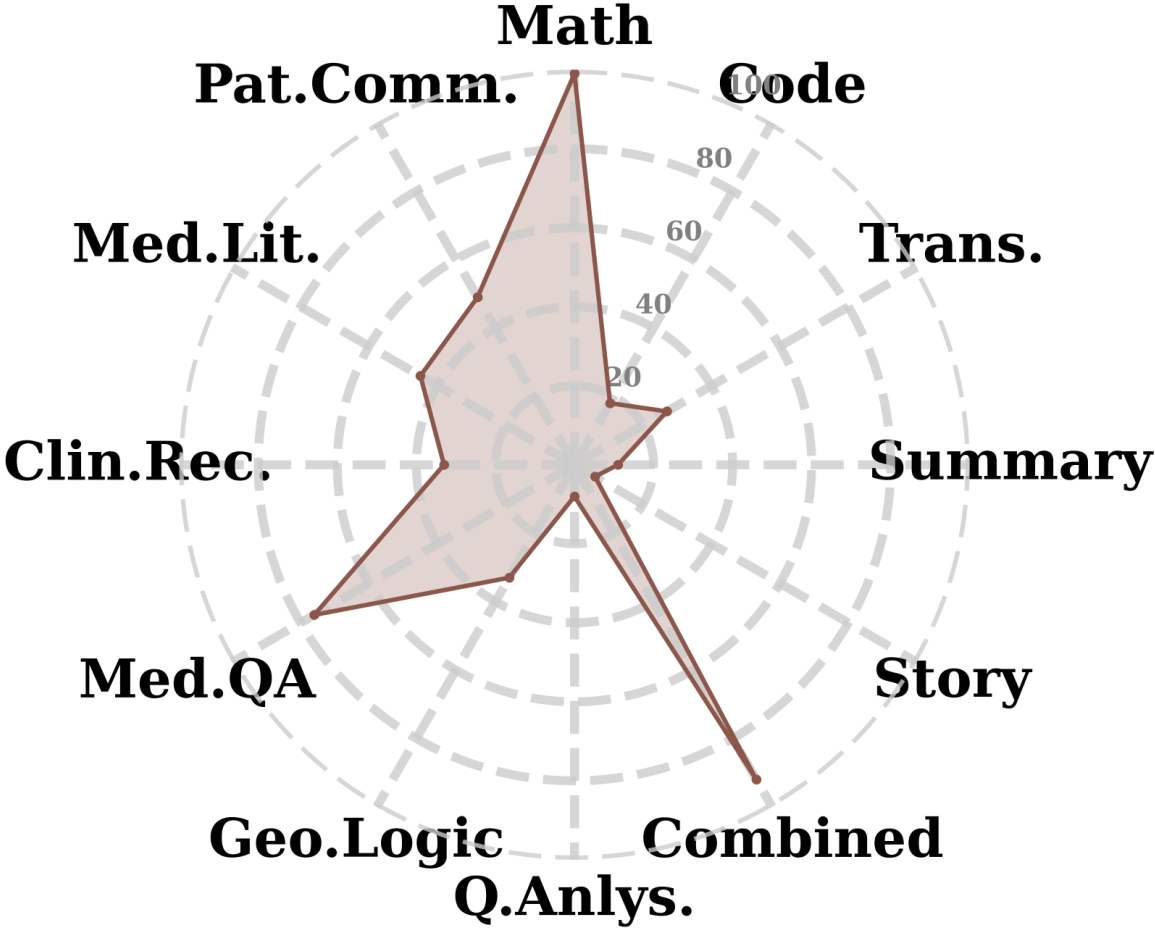}
    \end{subfigure}
    \hfill
    \begin{subfigure}{0.23\linewidth}
        \includegraphics[width=\linewidth]{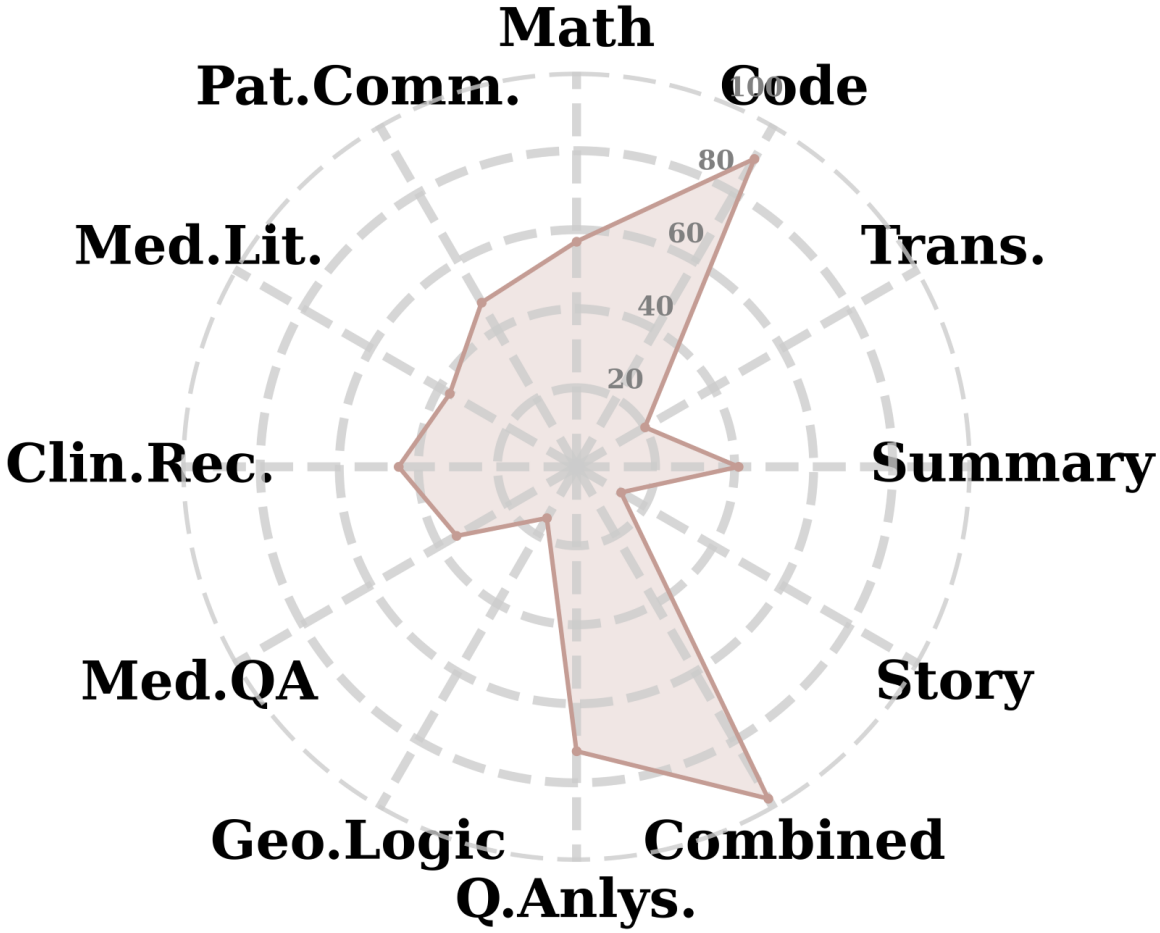}
    \end{subfigure}
    \vspace{0.4em}
    \begin{subfigure}{0.23\linewidth}
        \includegraphics[width=\linewidth]{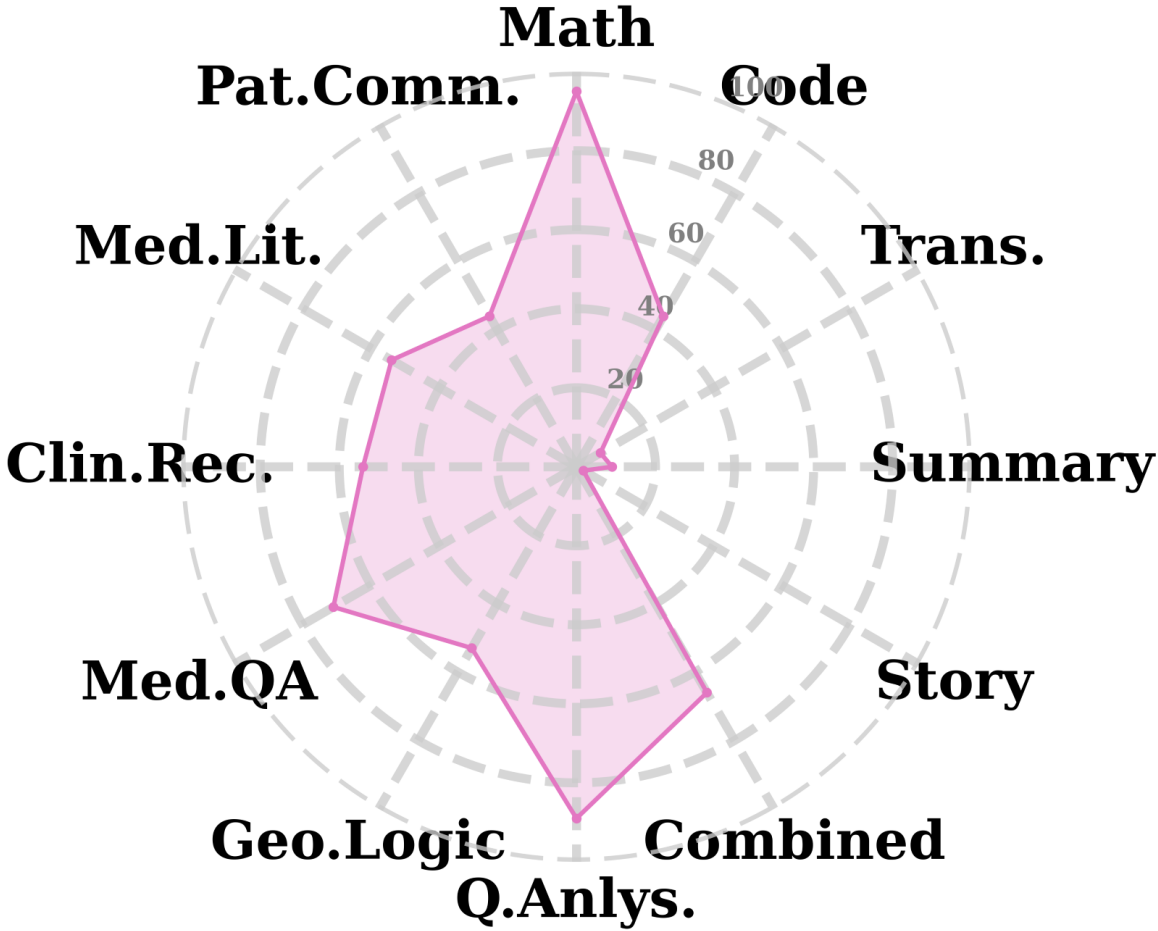}
    \end{subfigure}
    \hfill
    \begin{subfigure}{0.23\linewidth}
        \includegraphics[width=\linewidth]{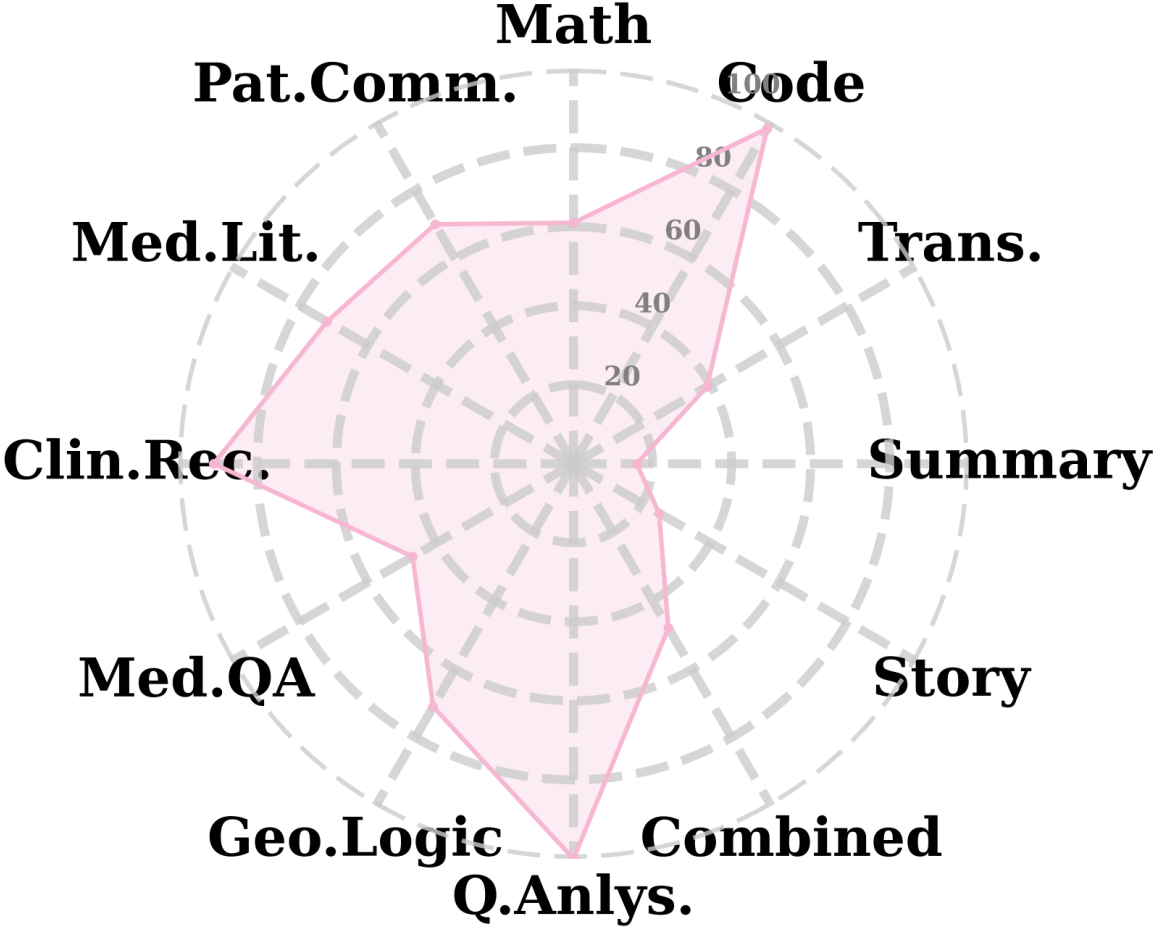}
    \end{subfigure}
    \hfill
    \begin{subfigure}{0.23\linewidth}
        \includegraphics[width=\linewidth]{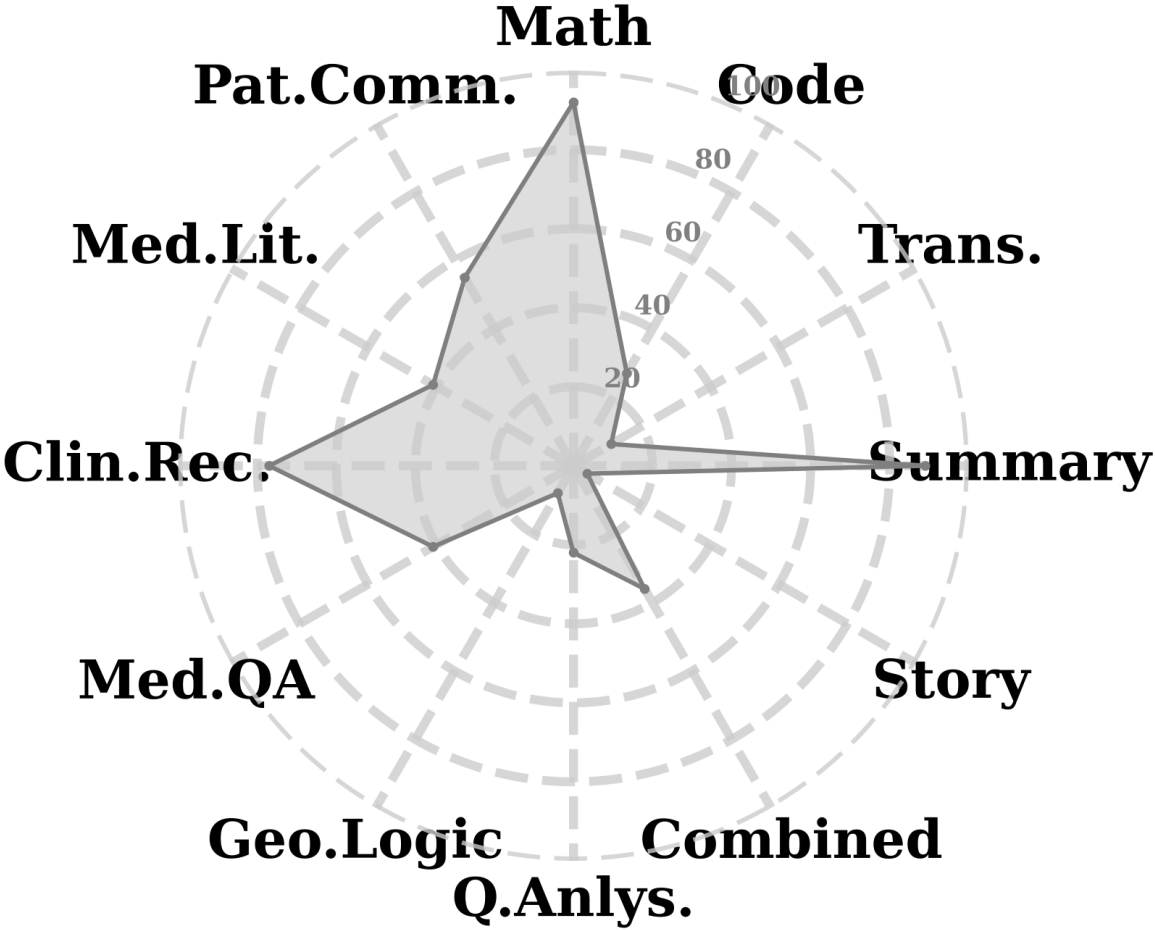}
    \end{subfigure}
    \hfill
    \begin{subfigure}{0.23\linewidth}
        \includegraphics[width=\linewidth]{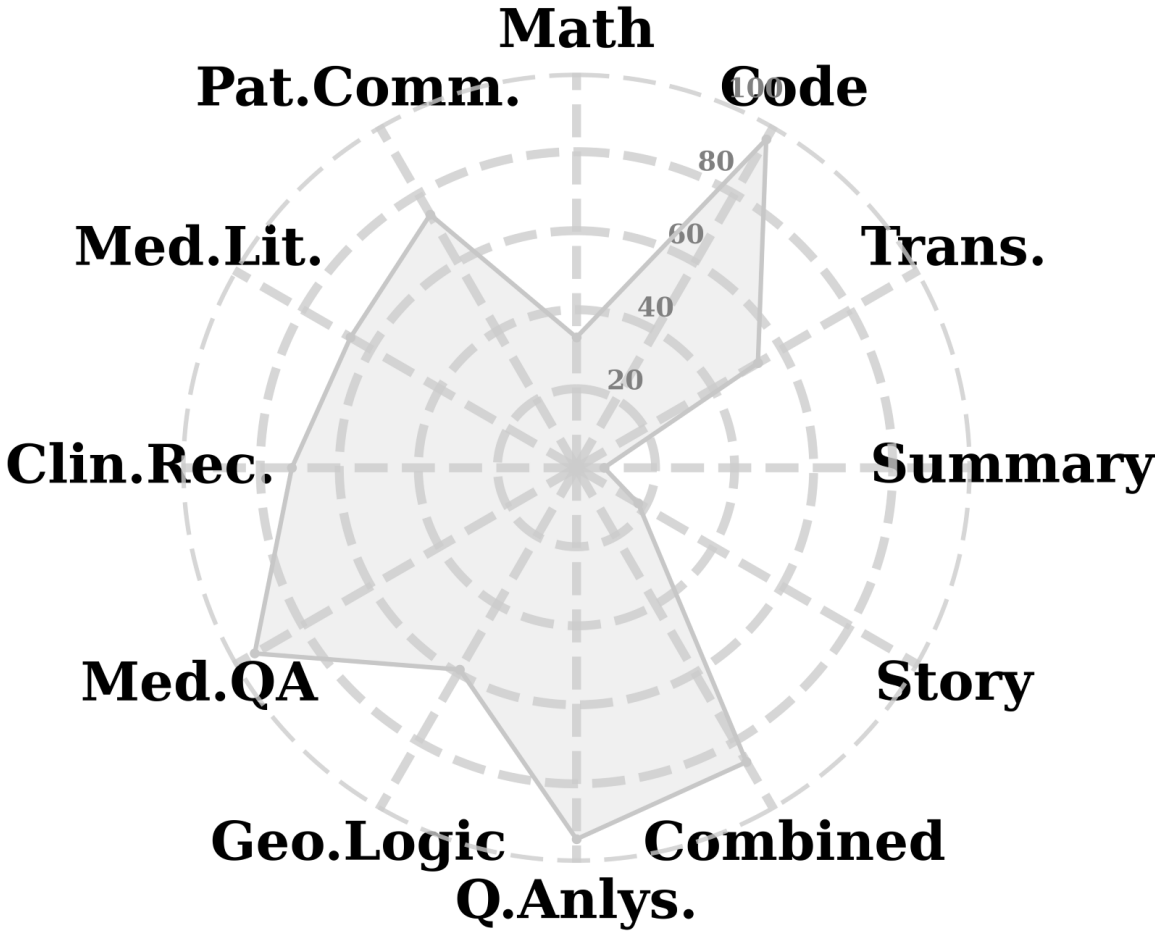}
    \end{subfigure}
    \vspace{0.4em}
    \begin{subfigure}{0.23\linewidth}
        \includegraphics[width=\linewidth]{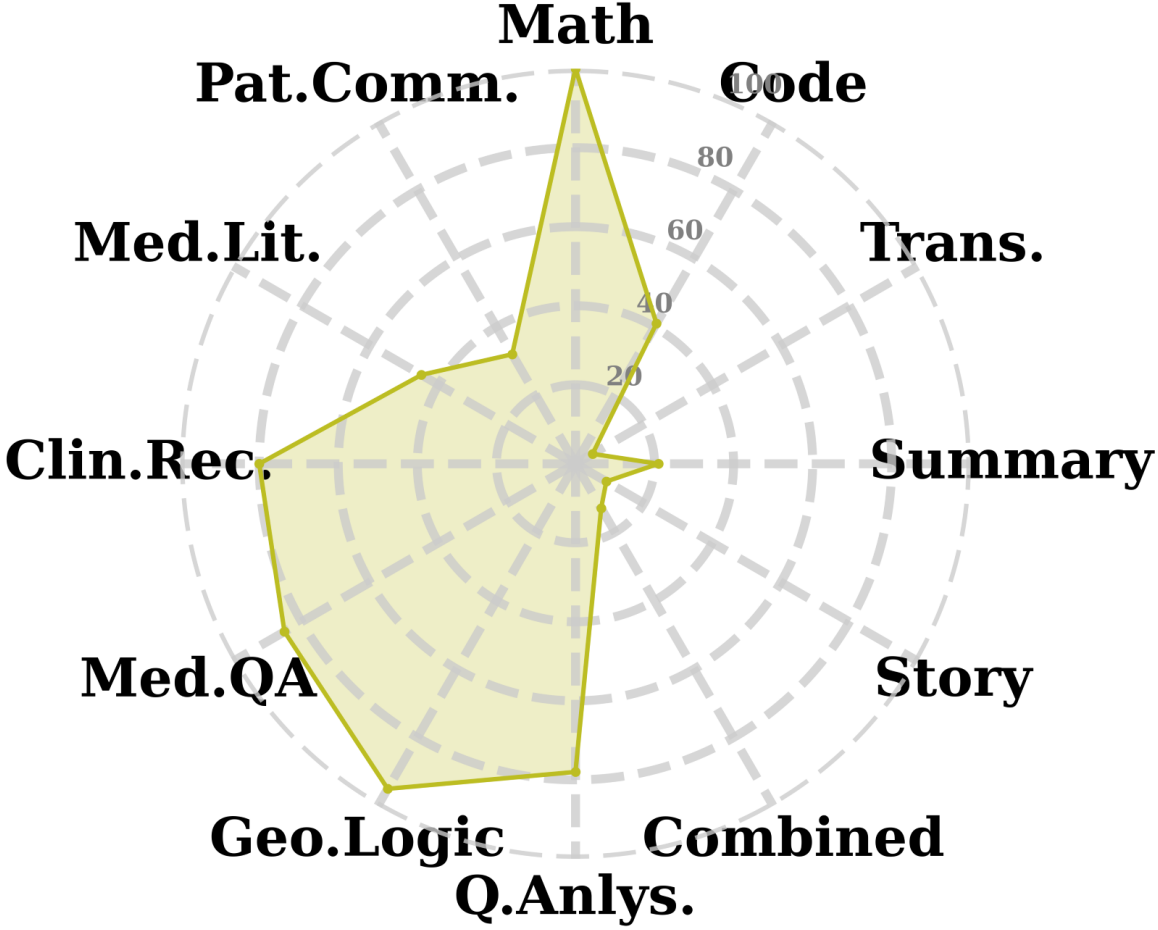}
    \end{subfigure}
    \hfill
    \begin{subfigure}{0.23\linewidth}
        \includegraphics[width=\linewidth]{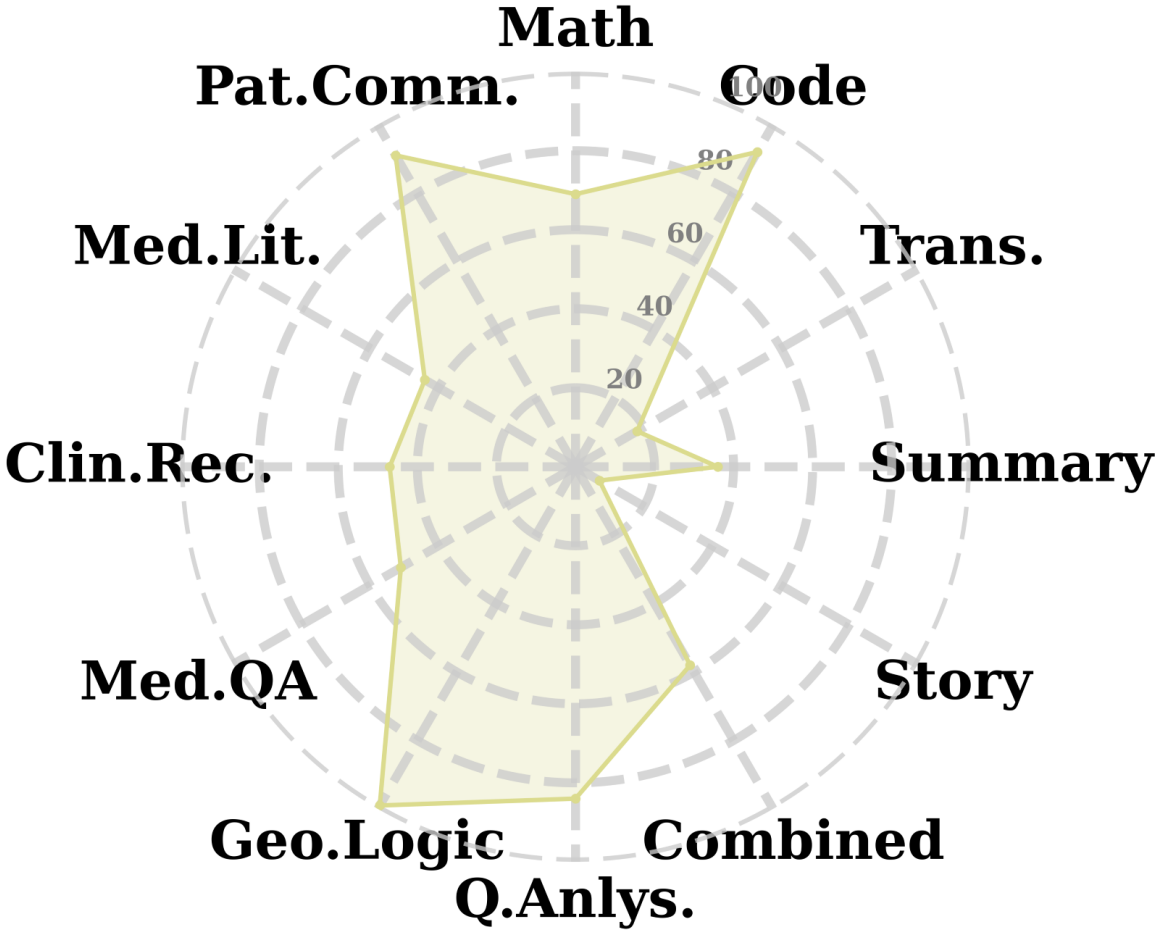}
    \end{subfigure}
    \hfill
    \begin{subfigure}{0.23\linewidth}
        \includegraphics[width=\linewidth]{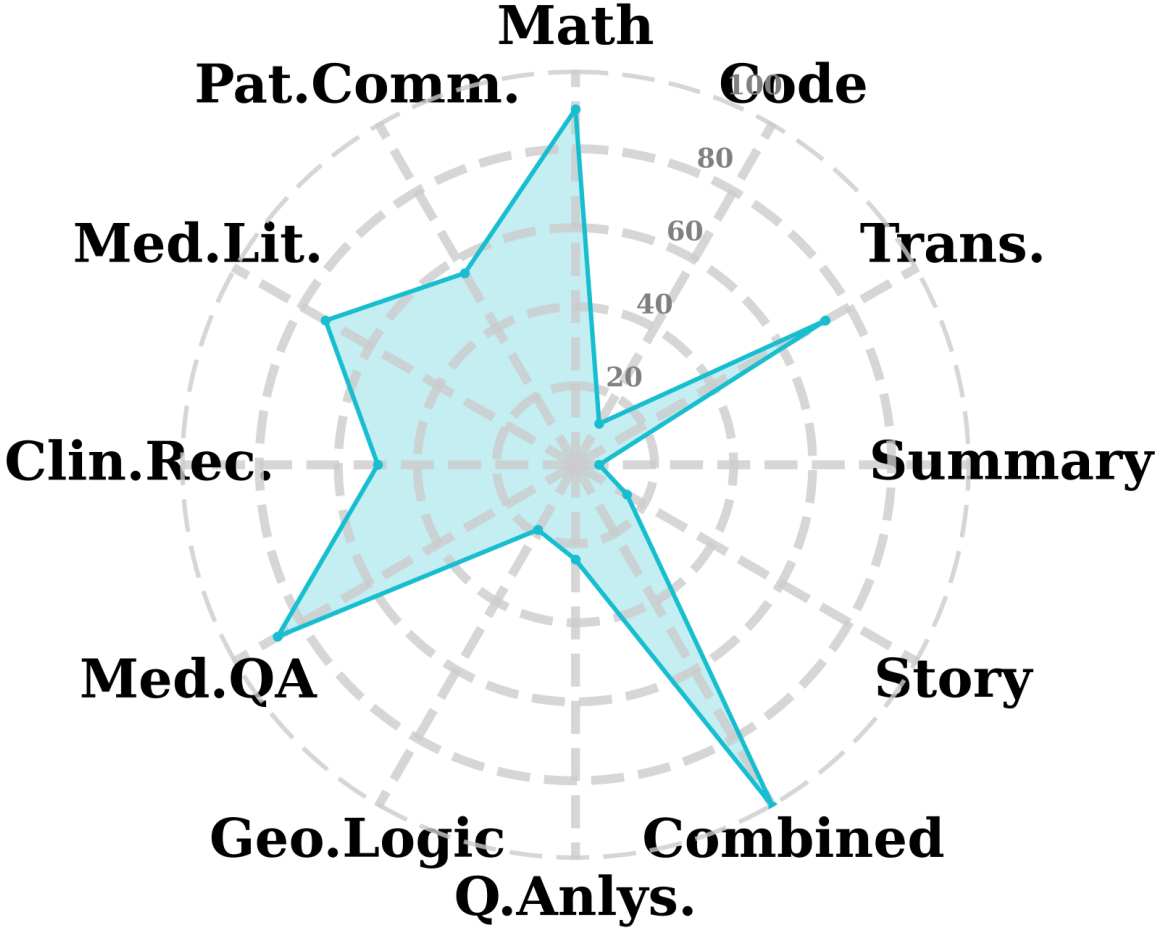}
    \end{subfigure}
    \hfill
    \begin{subfigure}{0.23\linewidth}
        \includegraphics[width=\linewidth]{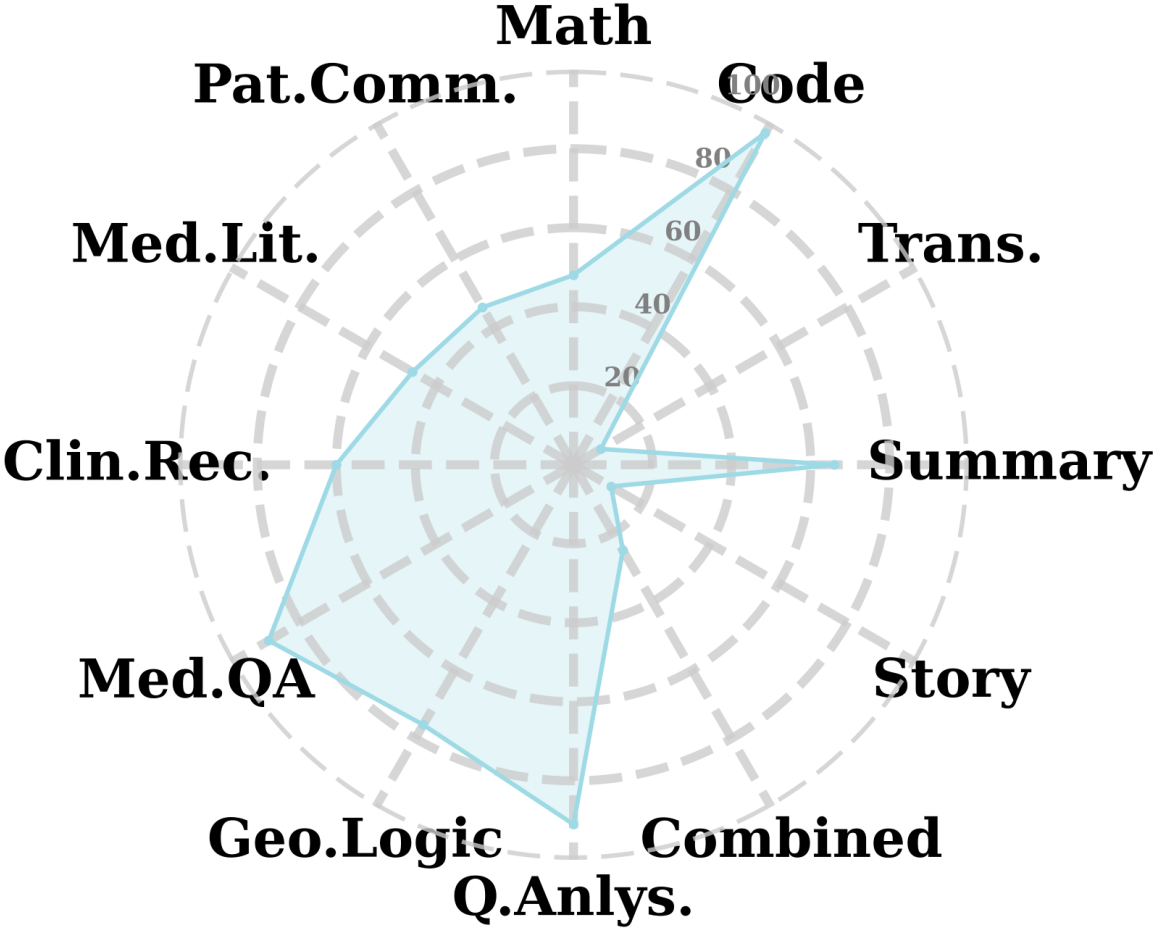}
    \end{subfigure}

    \caption{Agent Ability}
\end{figure}

\subsection{Iterative Reasoning, Privacy Rule Evolution, and API Usage}

\begin{figure}[htbp]
    \centering

     \begin{subfigure}{0.23\linewidth}
        \includegraphics[width=\linewidth]{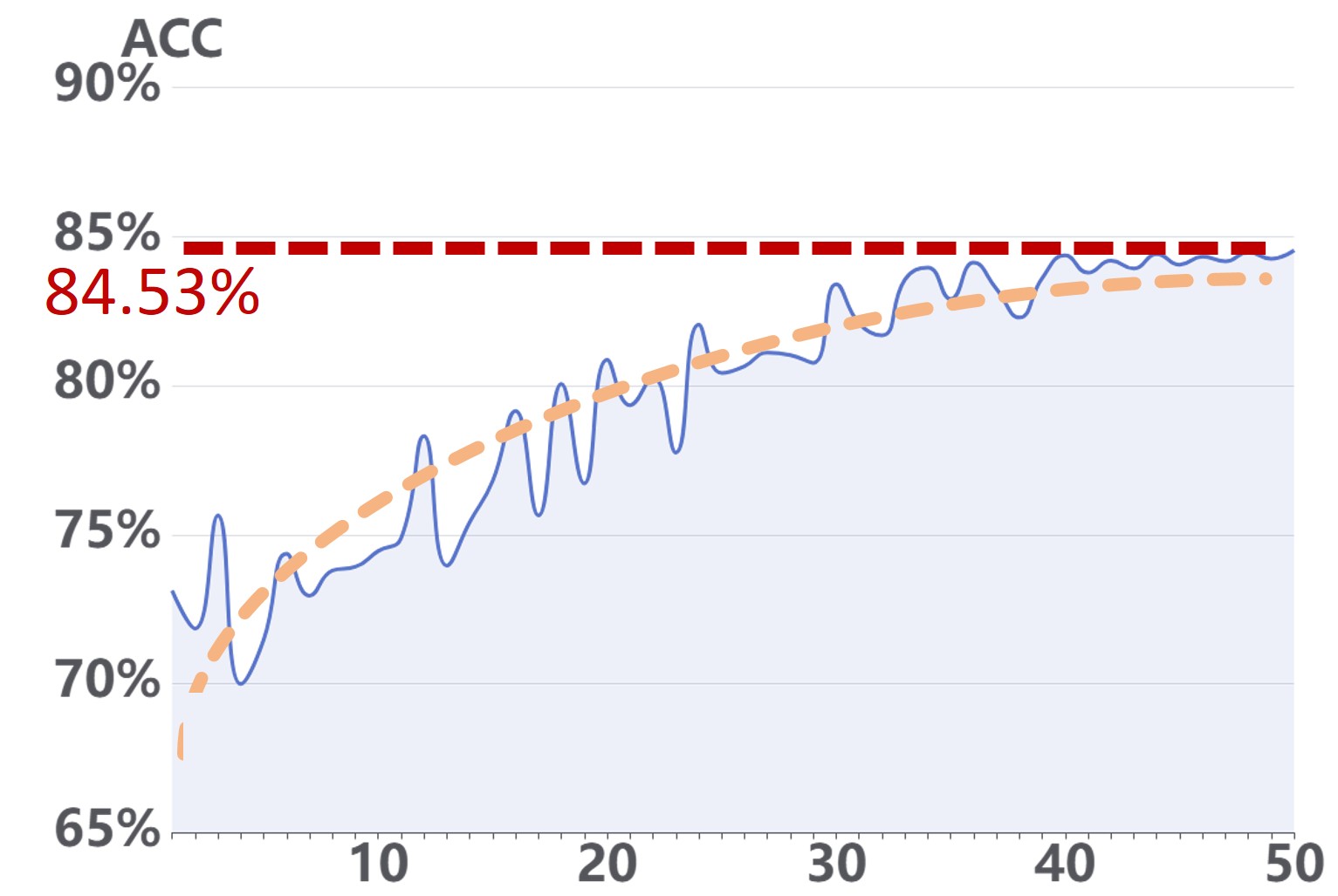}
        \caption{Federated iteration}
        \label{fig:itera_acc}
    \end{subfigure}
    \hfill
    \begin{subfigure}{0.23\linewidth}
        \includegraphics[width=\linewidth]{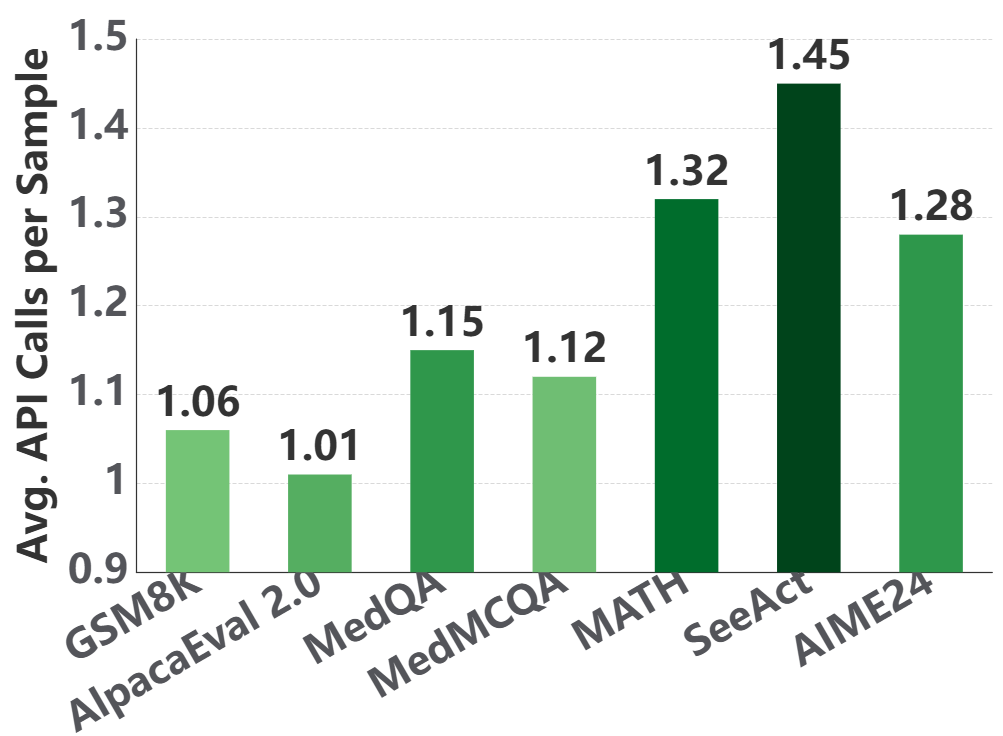}
        \caption{LLM API Call Count
        }
        \label{fig:api_freq_bar}
    \end{subfigure}
    \hfill
    \begin{subfigure}{0.2\linewidth}
        \includegraphics[width=\linewidth]{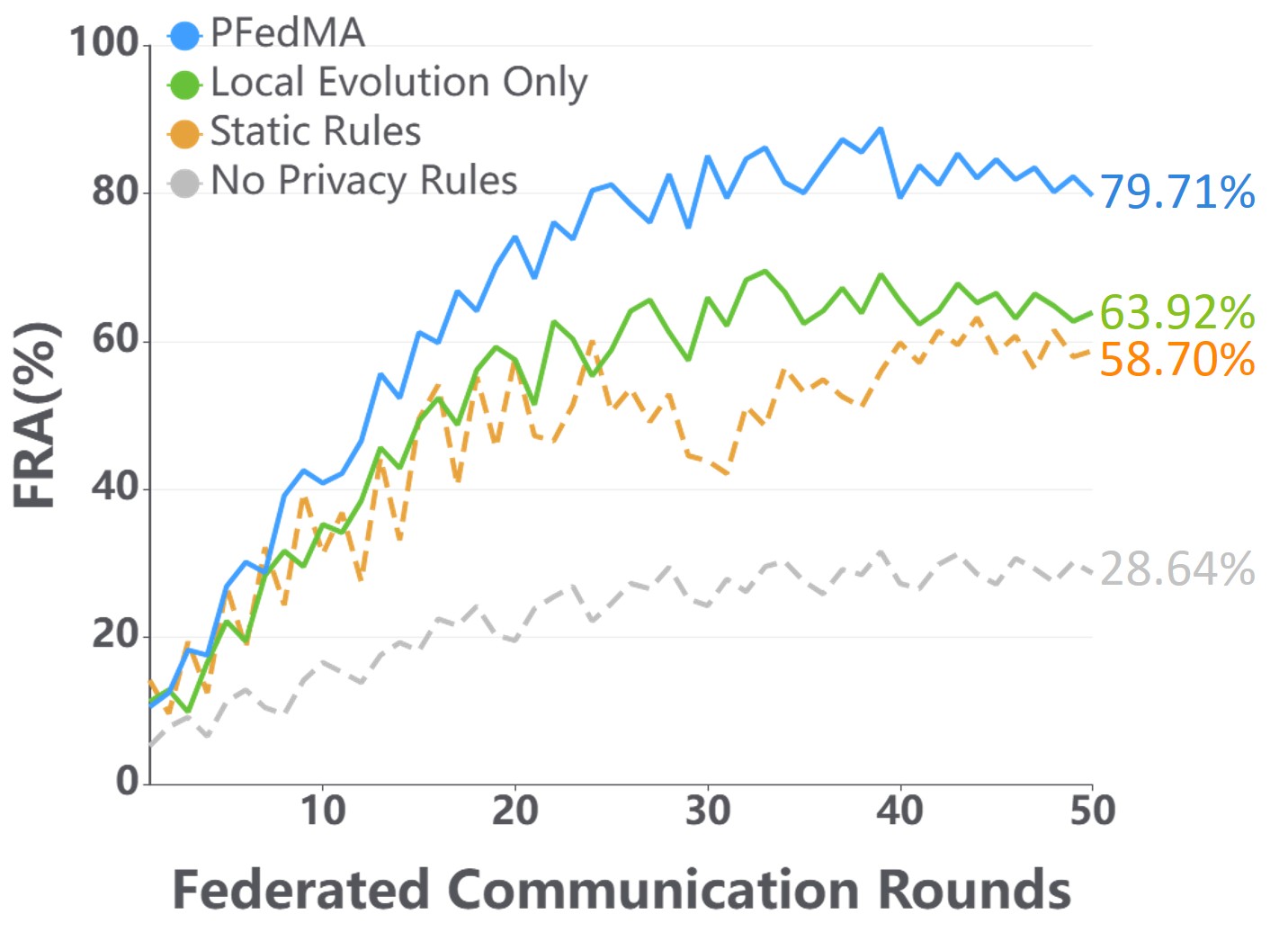}
        \caption{Federated Privacy Rule Effectiveness}
        \label{fig:privacy_evolution_curve}
    \end{subfigure}
    \hfill
    \begin{subfigure}{0.23\linewidth}
        \includegraphics[width=\linewidth]{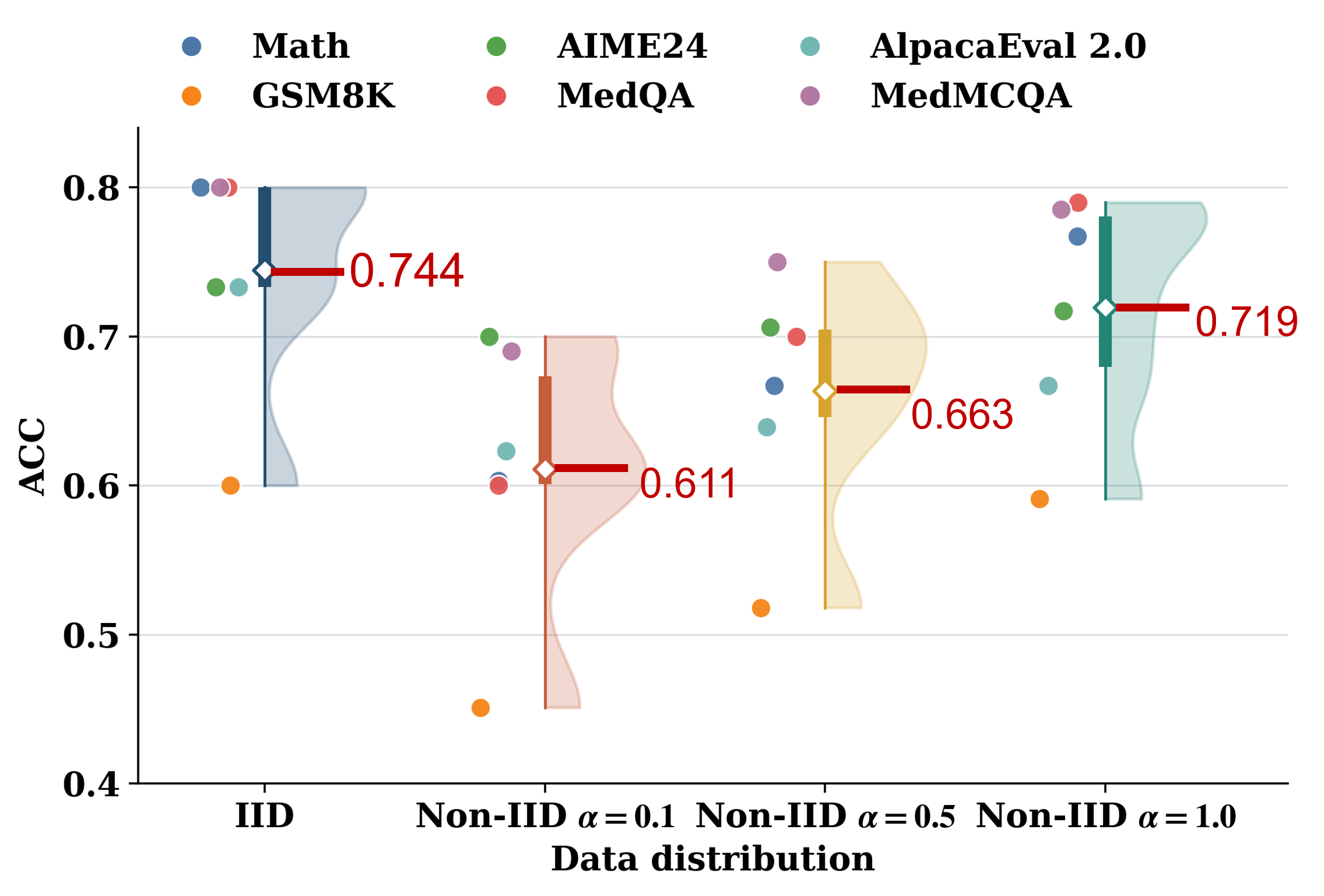}
        \caption{PFedMA Data Heterogeneity}
        \label{fig:half-violin-performance}
    \end{subfigure}
    \caption{Accuracy of federated iterative optimization, federated iteration rounds and demonstration of privacy rule effectiveness, API usage and client data heterogeneity in PFedMA with $N=100$ clients.}
    \label{fig:multi}
\end{figure}

Figure~\ref{fig:multi} presents answer accuracy and privacy rule effectiveness across rounds, together with API usage across benchmarks. For the Non IID setting, samples are partitioned across $N=100$ clients using a Dirichlet distribution $\mathrm{Dir}(\alpha)$ over the task domain categories. For each category, client allocation proportions are sampled from $\mathrm{Dir}(\alpha)$, where smaller $\alpha$ induces stronger client level heterogeneity.

\subsection{Efficiency and Robustness Trade offs}
Figures~\ref{fig:training_comparison_reference} examine how task performance and token consumption evolve with iteration rounds, while Figure~\ref{fig:robustness-utility} evaluates the robustness utility trade off as the proportion of adversarial agents increases.

\subsection{Hyperparameter Sensitivity}
\label{sec:appendix_hyperparameter_sensitivity}
Figure~\ref{fig:Hyperparameter} presents the sensitivity analysis of $\lambda_{\mathrm{pool}}$ and $\alpha_{\mathrm{con}}$ using independent parameter sweeps. Both hyperparameters exhibit non monotonic effects on robustness, indicating that either insufficient or excessive filtering strength can degrade performance. Smaller values provide weaker suppression of potentially unreliable responses, whereas overly large values may reject useful responses and reduce refusal rate. AHyperparameters were selected using a held-out validation set and fixed before evaluation on the test sets. Based on the validation results, we set $\lambda_{\mathrm{pool}}=0.3$ and $\alpha_{\mathrm{con}}=0.3$ ior all main experiments. Figure 5 reports the corresponding sensitivity analysis, showing that performance varies non-monotonically with both hyperparameters and that the selected values provide a favorable trade-off between attack suppression and refusal behavior.

\section{Privacy Scope.}
PFedMA focuses on limiting the disclosure of client local information during federated collaboration rather than hiding the task query from the server. Raw local responses, complete personalized privacy rules, local experiences, and triggering examples remain on device, while privacy processed responses, generalized rule candidates, and compact rule level feedback may be visible to the server. Our privacy evaluation measures sensitive information leakage from generated outputs and does not constitute a formal confidentiality guarantee.

\section{Limitations}
\label{sec:limitations}
PFedMA does not claim formal privacy guarantees such as differential privacy or cryptographic confidentiality. Instead, it provides semantic level privacy filtering by controlling information disclosure through adaptive privacy rules. As multi-agent systems move toward broader collaboration across organizational boundaries, stronger privacy assurances will become increasingly important. Establishing formal guarantees is an important direction for future work.